# Research Article

# European Space Agency Benchmark for Anomaly Detection in Satellite Telemetry

**Krzysztof Kotowski[a,*], Christoph Haskamp[b,*], Jacek Andrzejewski[a], Bogdan Ruszczak[a,c], Jakub Nalepa[d,a], Daniel Lakey[e], Peter Collins[f], Aybike Kolmas[g], Mauro Bartesaghi[h], Jose Martínez-Heras[i], Gabriele De Canio[f,*]**

[a] KP Labs, Bojkowska 37J, 44-100 Gliwice, Poland. E-mail: {kkotowski, jandrzejewski, bruszczak, jnalepa}@kplabs.pl

[b] Airbus Defence and Space GmbH, Airbus-Allee 1, 28199 Bremen, Germany. E-mail: christoph.haskamp@airbus.com

[c] Opole University of Technology, Proszkowska 76, 45-7558 Opole, Poland. E-mail: b.ruszczak@po.edu.pl

[d] Silesian University of Technology, Akademicka 16, 44-100 Gliwice, Poland. E-mail: jnalepa@ieee.org

[e] CGI Deutschland B.V. & Co. KG, Mornewegstr. 30, 64295 Darmstadt, Germany. E-mail: daniel.lakey@cgi.com

[f] European Space Operations Centre, European Space Agency, Robert-Bosch-Str. 5, 64293 Darmstadt, Germany. E-mail: {Peter.Collins, Gabriele.DeCanio}@esa.int

[g] Telespazio Germany GmbH, Europaplatz 5, 64293, Darmstadt. Email: aybike.kolmas@telespazio.de

[h] LSE Space GmbH, Robert-Bosch-Str. 16A, 64293 Darmstadt, Germany. E-mail: mauro.bartesaghi@lsespace.com

[i] Solenix GmbH, Spreestr. 3, 64295 Darmstadt, Germany. E-mail: jose.martinez@solenix.ch

[*] Corresponding authors

## ABSTRACT

Machine learning has vast potential to improve anomaly detection in satellite telemetry which is a crucial task for spacecraft operations. This potential is currently hampered by a lack of comprehensible benchmarks for multivariate time series anomaly detection, especially for the challenging case of satellite telemetry. The European Space Agency Benchmark for Anomaly Detection in Satellite Telemetry (ESA-ADB) aims to address this challenge and establish a new standard in the domain. It is a result of close cooperation between spacecraft operations engineers from the European Space Agency (ESA) and machine learning experts. The newly introduced ESA Anomalies Dataset contains annotated real-life telemetry from three different ESA missions, out of which two are included in ESA-ADB. Results of typical anomaly detection algorithms assessed in our novel hierarchical evaluation pipeline show that new approaches are necessary to address operators' needs. All elements of ESA-ADB are publicly available to ensure its full reproducibility.

**Frequently used abbreviations**: CNN – convolutional neural network, DL – deep learning, ESA – European Space Agency, ESA-AD – ESA Anomalies Dataset, ESA-ADB – ESA Anomaly Detection Benchmark, ESOC – European Space Operations Centre, LSTM – long short-term memory, ML – machine learning, SOE – spacecraft operations engineer, RNN – recurrent neural network, TC – telecommand, TSAD – time series anomaly detection, VAE – variational autoencoder

## 1. Main

Monitoring satellite telemetry time series data for anomalies is a daily practice of thousands of spacecraft operations engineers (SOEs) in mission control centres around the world. It ensures the safe and uninterrupted operation of multiple scientific, communication, observation, and navigation satellites. SOEs are typically supported with simple automatic anomaly detection systems that alarm when a measurement goes outside its predefined nominal limits or when a measurement correlates with a known anomalous pattern[1,2], but more sophisticated anomalies are usually detected manually which is a very expensive and error-prone task[3]. For this reason, all major space-related entities have been actively researching, developing and testing advanced automatic anomaly detection systems in the past years, including European Space Agency (ESA)[4,5], National Aeronautics and Space Administration (NASA)[2], Centre National d'Études Spatiales (CNES)[6], German Aerospace Center (DLR)[7], Japan Aerospace Exploration Agency (JAXA)[8], and Airbus, among others[9–12]. It is also one of the prioritised domains of Artificial Intelligence for Automation (A2I) Roadmap[13] of ESA and there is a growing trend in applying such systems directly on board satellites toward faster alarming and autonomous satellite health monitoring[14].

There are hundreds of algorithms for time series anomaly detection (TSAD) proposed in the literature (158 according to Schmidl et al.[15]) that could be viable solutions for the space sector, but currently, the main challenge everyone is facing regards the evaluation of different approaches. This happens because there are relatively few anomalies in flying missions[3] and no comprehensive data collection from multiple sources, thus it is hard to objectively conclude that one approach works better than the other. Moreover, multiple recent papers show that the majority of publicly available datasets, benchmarks, metrics, and protocols for TSAD are flawed and cannot be used for an unbiased evaluation of emerging machine learning (ML) techniques[16–19]. In addition, real-life satellite telemetry is an especially challenging example of a multi-variate time series with many specific problems and complexities related to its:

- high dimensionality and volume (years of recordings from up to thousands of channels per satellite[20]),
- complex network of dependencies between channels,
- complex characteristics (i.e. varying sampling frequencies across time and channels; data gaps caused by idle states and communication problems; trends connected with the degradation of spacecraft components; concept drifts related to different operational modes and mission phases),
- diverse types of channels (i.e. large variety and ranges of physical measures, categorical status flags, counters, and binary telecommands),
- noise and measurement errors due to the influence of the space environment.

The European Space Agency Benchmark for Anomaly Detection in Satellite Telemetry (ESA Anomaly Detection Benchmark or ESA-ADB, in short) aims not only to address all the mentioned challenges and flaws reported in the literature but also to establish a new standard for ML-based satellite telemetry analysis and general TSAD. It consists of three main components (visualised in Supplementary Fig. 1 for easier comprehension):

1. <u>Large-scale, curated, structured, ML-ready ESA Anomalies Dataset (ESA-AD, in short)</u> of real-life satellite telemetry collected from three ESA missions (out of which two are selected for benchmarking in ESA-ADB), manually annotated by SOEs and ML experts, and cross-verified using state-of-the-art algorithms.
2. <u>Evaluation pipeline</u> designed by ML experts for the practical needs of SOEs from the ESA's European Space Operations Centre (ESOC). It introduces new metrics designed for satellite telemetry according to the latest advancements in TSAD[16,18,21–23] and simulates real operational scenarios, e.g. different mission phases and real-time monitoring.

3. Benchmarking results of TSAD algorithms selected and improved to comply with the space operations requirements.

The main goal of ESA-ADB is to allow researchers and practitioners to design and thoroughly validate methods that could be directly applied in real space operations environments, taking into account all real-life challenges of satellite telemetry.

## 2. Results

The detailed nomenclature used in ESA-ADB is explained in Supplementary Material 1. Datasets and results are anonymised to avoid disclosing sensitive mission-specific information, such as channel names, timelines, or types of measured values, among others. The anonymisation does not affect the data integrity and it was verified that algorithms produce the same results as before anonymisation, see Supplementary Material 2.3. It does prevent from using physics-informed approaches[24] or domain-specific knowledge to design algorithms (for example, to match telecommands and channels by names or to expect anomalies in specific times, e.g. during increased solar activity). However, it also enforces the usage of universal data-driven approaches, instead of focusing on mission-specific intricacies.

### ESA Anomalies Dataset

The summary statistics of two missions from ESA-AD are presented in Table 1. The third mission from ESA-AD (Mission3) is not a part of ESA-ADB, because of a small number and triviality of anomalies (according to Definition 1 of Wu & Keogh[18]) and a large number of communication gaps and invalid segments – see Supplementary Material 2.1. Hence, it is omitted in this section for clarity. ESA-AD is publicly available under the link https://doi.org/10.5281/zenodo.12528696 .

The dataset includes 76 channels from Mission1 and 100 channels from Mission2, but only 58 and 47 channels, respectively, are monitored for anomalies (target channels) while the rest are meant to support the detection process (non-target channels; see the detailed definition in Supplementary Material 1.4). Channels are grouped into 6 subsystems – 4 in Mission1 and 5 in Mission2, with 3 matching subsystems between missions. Additionally, related channels with similar characteristics are organised into 18 (Mission1) and 29 (Mission2) numbered groups, so it is easier to manage the dataset for ML purposes. For each mission, there are hundreds of different telecommands with millions of executions, but only a small fraction directly relates to annotated anomalies and rare nominal events. Although telecommands were initially prioritised from 0 (least important) to 3 (most important), it is a part of the challenge to discover their true importance for TSAD. The number of data points exceeds 700 million for each mission which gives more than 7 gigabytes (GB) of compressed data in total. It is orders of magnitude more than for any other public satellite telemetry dataset, especially NASA SMAP and MSL datasets[2] (see Supplementary Table 6). The number of points is proposed as the main measure of the dataset volume because the duration of 17.5 years is not objective due to varying sampling rates and anonymisation.

Table 1. **Summary statistics of missions included in ESA-ADB**.

| | **Mission1** | **Mission2** | **Both missions** |
|---|---:|---:|---:|
| **Channels** | **76** | **100** | **176** |
| Target / Non-target | 58 / 18 | 47 / 53 | 105 / 71 |
| Channel groups | 18 | 29 | 47 |
| Subsystems | 4 | 5 | 6* |
| **Telecommands** | **698** | **123** | **821** |
| Priority 0/1/2/3 | 345 / 323 / 19 / 11 | 0 / 0 / 119 / 4 | 345 / 323 / 138 / 15 |
| Total executions | 1,594,722 | 1,918,002 | 3,512,724 |
| **Data points** | **774,856,895** | **776,734,364** | **1,551,591,259** |
| Duration (anonymised) | 14 years | 3.5 years | 17.5 years |
| Compressed size [GB] | 3.51 | 3.81 | 7.32 |
| Annotated points [%] | 1.80 | 0.58 | 1.19 |
| **Annotated events** | **200** | **644** | **844** |
| Anomalies | 118 | 31 | 148 |
| Rare nominal events | 78 | 613 | 690 |
| Communication gaps | 4 | 0 | 4 |
| Univariate / Multivariate | 32 / 164 | 9 / 635 | 41 / 799 |
| Global / Local | 113 / 83 | 585 / 59 | 698 / 142 |
| Point / Subsequence | 12 / 184 | 0 / 644 | 12 / 828 |
| **Distinct event classes** | **22** | **32****| **54** |

* there are 3 matching subsystems between missions.
** including unknown anomalies as a single class.

The anomaly density, in terms of annotated data points, is between 0.57% (Mission2) and 1.80% (Mission1) which addresses the flaw of unrealistic anomaly density reported for many popular TSAD datasets[18]. There are 844 annotated events (anomalies, rare nominal events, and communication gaps) in total. The majority of annotations for Mission2 are *rare nominal events* – atypical but expected or planned changes in the telemetry that are not anomalies from the operators' point of view (e.g. commanded manoeuvres, resets, or calibrations), but are likely to be detected as anomalies at their first occurrence by standard TSAD algorithms (see definitions in Supplementary Table 1). It would be of high practical importance to design algorithms that can recognise or memorise rare nominal events, so they are not alarmed as anomalies. There are just 4 short communication gaps (missing data) reported for Mission1.

Each anomaly and rare nominal event is described by three attributes corresponding to its dimensionality (uni-/multivariate), locality (local/global), and length (point/subsequence) according to the adjusted nomenclature of anomaly types by Blázquez-García et al.[25]. Most annotations are categorised as multivariate global subsequence, but there is also a diverse set of other types of anomalies (see Supplementary Table 7 for detailed statistics), including some especially challenging ones (Supplementary Material 2.2). Additionally, similar events are grouped into classes according to SOEs, so it is easier to analyse results and design detectors targeted at a specific class. The distributions of classes of events across missions' timelines are presented in Fig. 1. Note that events of the same class can have different categories, e.g. resets caused by telecommands are rare nominal events, but unexpected and non-commanded resets are anomalies. This difference is also reflected by the subclasses of events.

Our dataset has several features distinguishing it from the majority of related datasets. It is intended to reflect the raw telemetry data accessible for SOEs, with all its pros and cons, including irregular timestamps, varying sampling rates, anomalies in training data, communication gaps, and an overabundance of telecommands. Each channel has a separate set of annotations (like in the latest SMD[26], CATS[27], and TELCO[28] datasets), because the same anomaly may affect different channels in different ways and it is crucial to assess whether the algorithms can properly indicate affected channels to operators. Additionally, a single annotated event may be composed of multiple non-overlapping segments separated by nominal data, e.g. a series of short attitude disturbances caused by the same anomaly. An example of such an annotation is presented in Supplementary Fig. 9. This is to avoid assessing each segment as a separate anomaly in the evaluation metrics (the unrealistic anomaly density flaw). Missions vary significantly in terms of signal characteristics and specific challenges posed for TSAD algorithms. They are summarised in Supplementary Table 2.

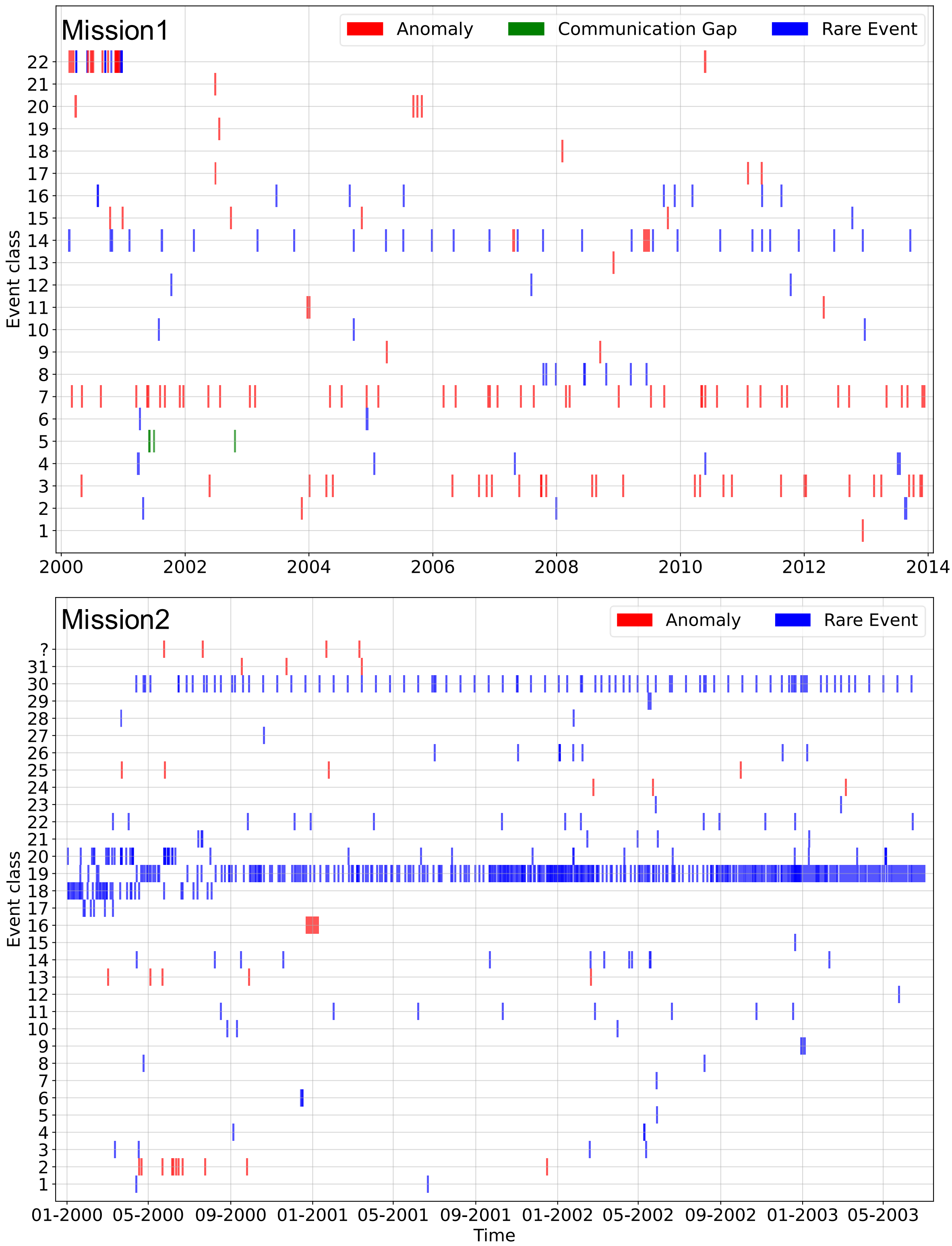

Fig. 1. **Distributions of events from different classes across timelines of Mission1 (top panel) and Mission2 (bottom panel)**. The bar width corresponds to the event length, but for better visualisation, the minimum width was limited to 10 and 2.5 days for Mission1 and Mission2, respectively. The question mark represents anomalies of unknown class for Mission2.

## Benchmarking results

The objective of the benchmarking experiments is to validate the performance of selected TSAD algorithms on ESA-AD using the proposed evaluation pipeline and the TimeEval framework[29]. The benchmarking code is publicly available under the link https://github.com/kplabs-pl/ESA-ADB to ensure its full reproducibility. The results of this study are intended to become a baseline and a starting point for future research. Hence, experiments do not aim to extensively tune hyperparameters or to find the best algorithm for satellite telemetry. They use default settings or parameters recommended by algorithms' authors, sometimes adjusted to the specific features of our datasets (see Supplementary Material 3.4). This is done intentionally, to present the results of typical TSAD approaches and to encourage the community to propose their own improvements and ideas. There are no algorithms in ESA-ADB that can explicitly distinguish between anomalies and rare nominal events, so the results in Table 2 are presented for all events (excluding only communication gaps). However, separate results considering only anomalies are available in Supplementary Table 9 for future comparisons. The metrics in the tables are ordered according to their priority in our hierarchical evaluation pipeline. For F-scores, there are also corresponding precisions and recalls for more detailed analysis. Unsupervised algorithms do not provide lists of affected channels, so channel-aware and subsystem-aware scores are not reported. Scores are rounded to 3 significant digits to account for the inherent uncertainty of manual annotations.

According to our hierarchical evaluation of the results in Table 2, Telemanom-ESA-Pruned is the best algorithm for Mission1. It achieves much higher corrected event-wise F0.5-scores than any other algorithm, in all mission phases (Supplementary Table 13) and with a very high value of 0.968 for anomalies only (Supplementary Table 9). It also achieves the highest alarming precision thanks to its dynamic thresholding scheme (NDT) which merges adjacent detections together. This highlights the importance of proper thresholding and postprocessing methods as a part of an algorithm. On the other hand, pruning significantly decreases channel-aware, subsystem-aware, and affiliation-based scores. Telemanom has the lowest ADTQC because 1) being a forecasting-based algorithm, it tends to detect anomalies too early (low ADTQC *After ratio*), and 2) the smoothing of forecasting errors applied in NDT strongly magnifies this effect. Unsupervised algorithms perform very poorly for Mission1 in terms of event-wise scores. DC-VAE-ESA and GlobalSTD are just slightly better which is especially disappointing for the former deep learning method. The main problem of these algorithms is a massive number of false detections caused by the noise and varying sampling rates in the data, as visible in the examples in Supplementary Material 4. However, DC-VAE-ESA has the highest ADTQC and affiliation-based scores, sometimes higher than for Telemanom-ESA. This suggests that more advanced thresholding or postprocessing may significantly improve the event-wise scores.

For Mission2, simple Windowed iForest[30] and Telemanom-ESA-Pruned algorithms turned out to be the best algorithms for the lightweight and full sets, respectively. Overall, unsupervised algorithms perform relatively well for Mission2, sometimes better than deep learning-based ones. It supports the need to always consider simple algorithms as a baseline[18,31]. Windowed iForest achieved very high corrected event-wise F0.5-score (0.949), ADTQC (0.985), and affiliation-based F0.5-score (0.959). The main reason is the relative triviality of the lightweight subset of Mission2 which contains mainly rare nominal events characterised by significant sudden changes in the signal (see Supplementary Fig. 5). However, the full set is much more challenging and contains many less obvious events (see Supplementary Fig. 6). It is reflected by much lower corrected event-wise F0.5-scores. Moreover, metrics for anomalies alone (Supplementary Table 9) show that no algorithm was able to accurately identify 9 actual anomalies in this overabundance of rare nominal events. This is the main challenge of this mission. Mission2 is particularly problematic for Telemanom because of a lack of clear periodicity and many commanded events that are impossible to forecast.

In most cases, the results in Table 2 for full sets of channels are much worse than for lightweight subsets, but they are not directly comparable due to the lower number of annotated events in the lightweight test sets (see tables in Supplementary Material 4.3). To allow for direct comparison, Supplementary Table 10 presents the results of the DC-VAE-ESA and Telemanom-ESA algorithms trained on full sets and tested on lightweight subsets. It confirms the initial observation – event-wise precisions and F-scores for Telemanom-ESA are much worse when trained on full sets of channels. This is one of the main challenges of high dimensional telemetry data – the more target channels there are, the higher chance of false detections is. Additionally, due to the strong interconnections between channels, false detections frequently seep into many irrelevant channels. The similar comparison for DC-VAE-ESA is inconclusive, but the overall results for this algorithm are relatively low.

Overall, the benchmarking results confirm that ESA-ADB poses a significant challenge for typical TSAD algorithms and none of them offer a perfect solution for both missions, especially for complete sets of channels. Some challenging events listed in Supplementary Material 2.2 are not detected by any algorithm.

Table 2. **Benchmarking results for detection of all events (excluding communication gaps) in lightweight subsets of channels and all channels for missions in ESA-ADB**. Boldfaced results indicate the best values among all algorithms (excluding *After ratio* of ADTQC which is just a helper value).

**Mission1 – trained and tested on the lightweight subset of channels 41-46**

| Metric | | PCC[32] | HBOS[33] | iForest[30] | Window iForest[30] | KNN[34] | Global STD3 | Global STD5 | DC-VAE-ESA STD3 | DC-VAE-ESA STD5 | Teleman-ESA | Teleman-ESA-Pruned |
|---|---|---|---|---|---|---|---|---|---|---|---|---|
| Event-wise | Precision | < 0.001 | < 0.001 | < 0.001 | < 0.001 | < 0.001 | 0.001 | 0.288 | 0.002 | 0.063 | 0.148 | **0.999** |
| | Recall | 0.554 | 0.585 | 0.585 | 0.738 | 0.754 | 0.431 | 0.169 | 0.554 | 0.338 | **0.894** | 0.424 |
| | F0.5 | < 0.001 | < 0.001 | < 0.001 | < 0.001 | < 0.001 | 0.001 | 0.253 | 0.003 | 0.075 | 0.178 | **0.786** |
| Channel-aware | Precision | Not available | | | | | 0.431 | 0.169 | 0.550 | 0.338 | **0.894** | 0.424 |
| | Recall | | | | | | 0.285 | 0.159 | 0.463 | 0.221 | **0.738** | 0.275 |
| | F0.5 | | | | | | 0.351 | 0.167 | 0.514 | 0.283 | **0.837** | 0.362 |
| Alarming precision | | 0.033 | 0.047 | 0.017 | 0.015 | 0.017 | 0.057 | 0.035 | 0.070 | 0.028 | 0.868 | **0.875** |
| ADTQC | After ratio | 0.833 | 0.763 | 0.711 | 0.375 | 0.612 | 0.929 | 0.909 | 0.972 | 0.955 | 0.136 | 0.143 |
| | Score | 0.840 | 0.781 | 0.784 | 0.563 | 0.803 | 0.770 | 0.688 | **0.901** | 0.803 | 0.428 | 0.197 |
| Affiliation-based | Precision | 0.535 | 0.543 | 0.543 | 0.599 | 0.522 | 0.559 | 0.699 | 0.584 | **0.780** | 0.727 | 0.711 |
| | Recall | 0.334 | 0.352 | 0.357 | 0.424 | 0.322 | 0.375 | 0.422 | 0.377 | 0.593 | **0.662** | 0.423 |
| | F0.5 | 0.477 | 0.490 | 0.492 | 0.553 | 0.464 | 0.509 | 0.618 | 0.526 | **0.734** | 0.713 | 0.626 |

**Mission1 – trained and tested on the full set of channels**

| Metric | | PCC[32] | HBOS[33] | iForest[30] | Window iForest[30] | KNN[34] | Global STD3 | Global STD5 | DC-VAE-ESA STD3 | DC-VAE-ESA STD5 | Teleman-ESA | Teleman-ESA-Pruned |
|---|---|---|---|---|---|---|---|---|---|---|---|---|
| Event-wise | Precision | < 0.001 | < 0.001 | < 0.001 | Out-of-memory | Out-of-memory | < 0.001 | 0.002 | < 0.001 | 0.005 | 0.007 | **0.050** |
| | Recall | 0.870 | **0.957** | 0.901 | | | 0.848 | 0.761 | 0.924 | 0.804 | 0.946 | 0.870 |
| | F0.5 | < 0.001 | < 0.001 | < 0.001 | | | < 0.001 | 0.003 | < 0.001 | 0.007 | 0.008 | **0.061** |
| Subsystem-aware | Precision | Not available | | | | | 0.520 | **0.728** | 0.526 | 0.640 | 0.676 | 0.395 |
| | Recall | | | | | | 0.694 | 0.538 | 0.764 | 0.670 | 0.859 | **0.861** |
| | F0.5 | | | | | | 0.528 | 0.664 | 0.538 | 0.623 | **0.689** | 0.436 |
| Channel-aware | Precision | Not available | | | | | 0.380 | 0.276 | 0.398 | 0.359 | **0.514** | 0.267 |
| | Recall | | | | | | 0.292 | 0.208 | 0.414 | 0.266 | 0.569 | **0.725** |
| | F0.5 | | | | | | 0.325 | 0.241 | 0.350 | 0.282 | **0.477** | 0.291 |
| Alarming precision | | 0.003 | 0.002 | 0.001 | | | 0.004 | 0.049 | 0.002 | 0.017 | 0.074 | **0.206** |
| ADTQC | After ratio | 0.613 | 0.443 | 0.378 | | | 0.718 | 0.743 | 0.647 | 0.716 | 0.322 | 0.463 |
| | Score | 0.642 | 0.603 | 0.643 | | | 0.723 | 0.691 | **0.752** | 0.692 | 0.673 | 0.684 |
| Affiliation-based | Precision | 0.563 | 0.539 | 0.559 | | | 0.560 | 0.575 | 0.559 | 0.578 | 0.545 | **0.649** |
| | Recall | 0.522 | **0.578** | 0.503 | | | 0.492 | 0.462 | 0.476 | 0.511 | 0.368 | 0.484 |
| | F0.5 | 0.554 | 0.547 | 0.547 | | | 0.545 | 0.548 | 0.540 | 0.563 | 0.497 | **0.607** |

**Mission2 – trained and tested on the lightweight subset of channels 18-28**

| Metric | | PCC[32] | HBOS[33] | iForest[30] | Window iForest[30] | KNN[34] | Global STD3 | Global STD5 | DC-VAE-ESA STD3 | DC-VAE-ESA STD5 | Teleman-ESA | Teleman-ESA-Pruned |
|---|---|---|---|---|---|---|---|---|---|---|---|---|
| Event-wise | Precision | 0.029 | 0.055 | 0.557 | 0.951 | < 0.001 | 0.006 | 0.061 | 0.003 | 0.064 | 0.188 | **0.978** |
| | Recall | **1.000** | 0.911 | 0.974 | 0.940 | **1.000** | **1.000** | **1.000** | **1.000** | **1.000** | 0.986 | 0.540 |
| | F0.5 | 0.036 | 0.068 | 0.609 | **0.949** | 0.001 | 0.007 | 0.075 | 0.003 | 0.079 | 0.224 | 0.842 |
| Channel-aware | Precision | Not available | | | | | 0.951 | 0.992 | 0.904 | **0.995** | 0.831 | 0.465 |
| | Recall | | | | | | 0.462 | 0.372 | 0.554 | 0.451 | **0.870** | 0.384 |
| | F0.5 | | | | | | 0.767 | 0.723 | 0.787 | 0.783 | **0.822** | 0.442 |
| Alarming precision | | 0.061 | 0.105 | 0.075 | 0.217 | 0.060 | 0.054 | 0.061 | 0.052 | 0.068 | **0.912** | 0.862 |
| ADTQC | After ratio | 0.983 | 0.994 | 1.000 | 0.948 | 0.391 | 0.946 | 0.989 | 0.908 | 0.991 | 0.087 | 0.351 |
| | Score | **0.999** | 0.990 | 0.991 | 0.985 | 0.724 | 0.997 | 0.997 | 0.996 | 0.997 | 0.507 | 0.757 |
| Affiliation-based | Precision | 0.890 | 0.936 | **0.982** | 0.968 | 0.561 | 0.740 | 0.935 | 0.680 | 0.939 | 0.688 | 0.759 |
| | Recall | 0.580 | 0.867 | **0.952** | 0.925 | 0.243 | 0.296 | 0.717 | 0.293 | 0.788 | 0.544 | 0.530 |
| | F0.5 | 0.804 | 0.921 | **0.976** | 0.959 | 0.445 | 0.569 | 0.881 | 0.538 | 0.904 | 0.654 | 0.699 |

**Mission2 – trained and tested on the full set of channels**

| Metric | | PCC[32] | HBOS[33] | iForest[30] | Window iForest[30] | KNN[34] | Global STD3 | Global STD5 | DC-VAE-ESA STD3 | DC-VAE-ESA STD5 | Teleman-ESA | Teleman-ESA-Pruned |
|---|---|---|---|---|---|---|---|---|---|---|---|---|
| Event-wise | Precision | 0.064 | 0.016 | 0.022 | 0.034 | Out-of-memory | 0.006 | 0.052 | 0.002 | 0.008 | 0.052 | **0.058** |
| | Recall | **0.997** | 0.820 | 0.903 | 0.746 | | **0.997** | 0.992 | **0.997** | 0.994 | 0.992 | 0.964 |
| | F0.5 | 0.079 | 0.020 | 0.027 | 0.042 | | 0.008 | 0.064 | 0.002 | 0.011 | 0.064 | **0.071** |
| Subsystem-aware | Precision | Not available | | | | | 0.910 | **0.983** | 0.672 | 0.911 | 0.409 | 0.258 |
| | Recall | | | | | | 0.965 | 0.951 | 0.967 | 0.952 | **0.984** | 0.896 |
| | F0.5 | | | | | | 0.910 | **0.969** | 0.699 | 0.907 | 0.451 | 0.298 |
| Channel-aware | Precision | Not available | | | | | 0.899 | **0.941** | 0.774 | 0.931 | 0.584 | 0.326 |
| | Recall | | | | | | 0.550 | 0.490 | 0.592 | 0.507 | 0.783 | **0.823** |
| | F0.5 | | | | | | **0.791** | 0.787 | 0.713 | 0.783 | 0.592 | 0.368 |
| Alarming precision | | 0.094 | 0.148 | 0.112 | 0.179 | | 0.060 | 0.077 | 0.066 | 0.083 | 0.771 | **0.790** |
| ADTQC | After ratio | 0.975 | 0.906 | 0.939 | 0.852 | | 0.925 | 0.989 | 0.663 | 0.930 | 0.104 | 0.274 |
| | Score | 0.990 | 0.939 | 0.967 | 0.928 | | 0.983 | **0.993** | 0.825 | 0.985 | 0.513 | 0.648 |
| Affiliation-based | Precision | 0.814 | 0.570 | 0.621 | 0.608 | | 0.680 | **0.904** | 0.603 | 0.859 | 0.586 | 0.591 |
| | Recall | **0.631** | 0.455 | 0.499 | 0.474 | | 0.280 | 0.628 | 0.324 | 0.625 | 0.348 | 0.347 |
| | F0.5 | 0.770 | 0.543 | 0.592 | 0.575 | | 0.529 | **0.831** | 0.515 | 0.799 | 0.516 | 0.518 |

The processing times of the algorithms are given in Supplementary Material 4.6. They are all possible to run in real-time using our computational resources given in Supplementary Material 4.5.

## 3. Discussion

ESA-ADB is a starting point for the development of better algorithms for satellite telemetry anomaly detection. It was designed in close collaboration between ML experts and SOEs to fulfil the need for a reliable benchmark for both communities. The results show that ESA-ADB poses a significant challenge for popular TSAD algorithms, and many changes had to be applied in the TimeEval framework[29], training procedures, and algorithms (i.e. Telemanom[2] and DC-VAE[28]) to make them applicable to our use case (i.e. to handle large datasets, tens of channels, varying sampling rates, streaming evaluation, and anomalies in the training data). Although the results of Telemanom-ESA-Pruned may seem promising, it is a highly parametrised approach and the selected thresholds may not be optimal for other missions. It is an invitation for the community to build upon those algorithms, try other ones (out of hundreds in the literature), or propose new approaches to address the challenges and requirements of anomaly detection in satellite telemetry.

The ESA Anomalies Dataset contains tens of target channels and millions of data points which makes it a challenging data volume for most algorithms, while still being manageable using a standard PC and relatively comprehensible for manual analysis. However, one needs to remember that ESA-AD contains only a small subset of channels from selected missions. There may be thousands of channels in actual telemetry and proposing a perfect solution for ESA-ADB would be still just the first milestone on the way to reliable anomaly detection systems for space operations. Moreover, potential solutions must be not only accurate but also fast enough to be run in real-time on computational resources accessible to mission control and, as the ultimate goal, on board satellites.

Our evaluation pipeline considers all recent recommendations for multivariate time series anomaly detection[16,18,22,23]. It proposes new quantitative metrics, dataset splits simulating real operational scenarios, and the idea of hierarchical evaluation. Some of the proposed metrics may seem too strict when looking at the results, but they represent practical aspects of space operations and encourage to look for better methods. Priorities or weights of aspects may differ between use cases and only a part of the evaluation pipeline may be relevant in domains different from satellite telemetry. Also, not every mission is appropriate for objective testing of anomaly detection as shown in the example of the rejected Mission3. Our goal was to ensure that improving the results of ESA-ADB does not create an illusion of progress but solves real challenges in space operations and TSAD domains. To support this statement, Table 3 includes a summary of how ESA-ADB addresses flaws reported by Wu & Keogh[18].

Table 3. **A list of flaws reported by Wu & Keogh[18] and how they are addressed by ESA-ADB**.

| Flaw | How does ESA-ADB address it? |
|---|---|
| Triviality | • ESA-AD is large and contains a diverse set of anomaly types and concept drifts which hamper the usage of simple algorithms<br>• ESA-AD offers a selection of non-trivial anomalies, so they can be evaluated separately (Supplementary Material 2.2)<br>• ESA-ADB includes a set of simple algorithms to verify the potential triviality of anomalies |
| Unrealistic anomaly density | • ESA-AD is large and the anomaly density in the dataset is below 2% of data points<br>• There are only dozens of anomalous events per year<br>• Series of separate annotated segments within a short region are usually assigned to the same event and are treated as such when computing metrics |
| Mislabelled ground truth | • While this flaw cannot be fully resolved in real-life datasets there were several iterations of the annotation refinement process aided by unsupervised and semi-supervised algorithms to identify potential mislabelling[35] |
| Run-to-failure bias | • Anomalies are scattered across long, failure-free, operational periods of acquired telemetry data from real satellite missions (see Fig. 1.) |

ESA-ADB is an important departure point for further endeavours. The benchmark is meant to evolve and new contributions by researchers and organisations to further improve and extend it are welcome. Despite our best efforts, some mislabelling is inevitable in such substantial amounts of real-life data, so we are open to requests for corrections, and we plan to release updated versions of ESA-AD. Other potential improvements include adding new missions, proposing new algorithms, and better thresholding schemes to fulfil all posed requirements. To create even larger and better datasets, it would be very desirable to introduce a standardised ML-oriented anomaly reporting system for space operations. Especially interesting algorithmic directions are related to the adaptation of Matrix Profile methods[36,37] and transformers with positional time encoding[38]. ESA-AD allows for testing few-shot and one-shot learning techniques that utilise known anomalies in training sets and can memorise rare nominal events[39]. Due to its size and diversity, it can be useful in the domains of time series forecasting[40], telemetry data compression[41], and foundation models[42]. Distribution shifts caused by some rare nominal events make it an interesting resource in continual learning[43] and change-point detection[44]. The large collection of telecommands may be explored to test methods of decision support for SOEs (one of use cases from the ESA A2I Roadmap[13]).

ESA-ADB should allow researchers from academia and practitioners from space agencies and space industry to cooperate, compete and develop newer approaches to solve a common problem. Our future activities include the organisation of open data science competitions based on our dataset, stimulating the international community to develop better methods for satellite telemetry analysis. Enabling this exchange should help to produce useful solutions for the daily practice of thousands of SOEs in mission control centres around the world and for future autonomous space missions.

# 4. Methods

## Dataset collection and curation

Three missions (satellites) of different types (purposes, orbits, and launch dates) were selected from the ESA portfolio based on the survey conducted among SOEs about the presence of historical anomalies that are problematic to detect using existing out-of-limit approaches (some of them are listed in Supplementary Table 4). The selection focused on collecting a large dataset with a possibly diverse spectrum of signals and anomalies, to avoid common flaws of triviality, unrealistic anomaly density, or run-to-failure bias[18]. The most interesting continuous time windows for anomaly detection were identified based on the occurrence of reported problematic anomalies and other events reported in the Anomaly Report Tracking System (ARTS) (artsops.esa.int/documentation/about) used at ESOC. Raw satellite telemetry was structured according to Supplementary Material 2.4 and manually annotated in cooperation with SOEs using the OXI annotation tool (oxi.kplabs.pl)[45] created specifically for the project needs. While annotating, a special focus was put on the precise identification of anomaly starting points for all channels. On the other hand, anomaly end times may be less accurate, because they are much harder to identify objectively, especially for long anomalies. Importantly, ARTS reports are intended for human use and are not well-suited for ML purposes. They usually include only approximate time ranges and a small fraction of affected channels. Moreover, well-known anomalies and rare nominal events are often not reported. Thus, the whole signal was carefully revisited by the ML team in search of any suspicious events. An initial list of subsystems, channels, and telecommands relevant for anomaly detection was proposed by SOEs, but was gradually extended during several iterations of the annotation refinement process in which overlooked anomalies were discovered by the ML team using different TSAD algorithms. During this process, channels were divided into target and non-target for anomaly detection. Non-target channels should only be used as additional information for the algorithms. They are not annotated and are not assessed in the benchmark. Examples include status flags, counters, and metadata such as location coordinates, where anomalies are not expected or it is not possible to check for anomalies without external data. Related channels measuring the same physical values and showing similar characteristics are organised into numbered groups, so it is easier to manage the dataset for ML purposes, e.g. to train group-specific models or to visualise results. For more details about data collection, annotation, and refinement processes, see our previous related work (Kotowski et al., 2023)[35].

There are hundreds of different telecommands (TCs) in each mission. Some of them are critical foe detecting annotated anomalies (i.e. when there is no reaction to the TC or the reaction is different than usual) or distinguishing anomalies from rare nominal events. However, it may be impractical to use them all in anomaly detection algorithms. Thus, 4 different priority levels for TCs were introduced as a suggestion about their potential usefulness for anomaly detection algorithms. The priorities from the least important to the most important are:

0. TCs not directly related to any subsystem included in the dataset.
1. TCs related to subsystems included in the dataset but not marked as potentially valuable for anomaly detection by SOEs.
2. TCs selected as potentially valuable for anomaly detection by SOEs.
3. A fraction of TCs of priority 2. assessed as valuable for anomaly detection by the ML team. The main rejection criteria were the scarcity of occurrences in the training data (less than 3) or no occurrences in the test data.

TCs of priority 3. are used as input for DC-VAE-ESA and Telemanom-ESA algorithms trained on full sets of channels. These priorities are only suggestions and ESA-ADB users are welcome to experiment with any combination of TCs.

### Dataset division

Each mission is divided into halves of which the first half is taken as a training set and the second half as a test set. This gives 84 months of training data for Mission1 and 21 months for Mission2. In both cases, the last 3 months of the training set are taken as the validation set. As agreed between the ML team and SOEs, 3 months is long enough to reliably monitor the performance of algorithms in the latest environmental conditions. The validation and test sets include only samples later than the training ones to avoid data leakage from future samples. Anomalies appear in all sets, including training and validation ones. Such a division employs all available data and it represents a mature phase of the mission in which a significant amount of data is available for training. However, anomaly detection systems are also desirable for SOEs already in the early phases of missions. Thus, shorter training sets are also proposed and analysed in Supplementary Material 4.3 to assess the robustness of the algorithms to changing mission conditions and to identify the earliest mission phase in which reliable detectors can be trained.

**Lightweight subsets of channels**

In the default setting of ESA-ADB, all channels and telecommands (of priority 3.) are used as input and all target channels are used as output from algorithms. However, anomaly detection in tens or hundreds of channels simultaneously may be a very challenging task and it takes a lot of computing power to process such data, so for initial experiments, familiarisation with ESA-ADB, simpler models, and potential on-board applications, there are also lightweight subsets of channels proposed in ESA-ADB. These are channels 41-46 from subsystem 5 for Mission1 and channels 18-28 from subsystem 1 for Mission2. The selection is subjective, but the main goal was to provide channels that are challenging for algorithms (in terms of the number and difficulty of anomalies), interesting for SOEs (in terms of the satellite health monitoring), relatively easy to visualise and analyse manually, and not strongly dependent on other channels or subsystems (so they are possible to analyse in isolation from the whole system to some extent). The lightweight subsets do not include any telecommands. Selected channels from these subsets are presented in Supplementary Fig. 4 and Supplementary Fig. 9 for Mission1 and in Supplementary Fig. 5 and Supplementary Fig. 8 for Mission2.

## Taxonomy of anomaly types

To the best of our knowledge, the taxonomy by Blázquez-García et al.[25] is the only one in the literature that comprehensively defines multivariate anomaly types, and our definitions are built based on this foundation. It divides anomaly types into point and subsequence ones, where point anomalies are defined as single outlying data points. However, this definition does not take into account varying sampling rates for which the length of "a single data point" may differ in time. Thus, for our purposes, multi-instance point anomalies are allowed if they are relatively short fragments of the signal that resemble points or peaks when inspected using a typical sampling frequency for the channel. Both point and subsequence anomalies may be univariate or multivariate depending on whether they affect one or more channels. Anomalies can additionally be divided into global and local (contextual) ones, similarly as proposed in behaviour-driven taxonomy by Lai et al.[46]. To make the original definitions more specific in our taxonomy, the global subsequence anomaly is defined as a subsequence of anomalous values in which at least one instance can be treated as a global point anomaly.

In the proposed taxonomy, each anomaly type can be described by three attributes: dimensionality (uni-/multi-variate), locality (local/global), and length (point/subsequence), as presented in Supplementary Fig. 10. These attributes can be automatically inferred from per-channel annotations:

1. **Dimensionality** can be inferred by counting the number of channels affected by an anomaly. One affected channel makes it univariate and more affected channels make it multivariate.
2. To infer **locality**, we calculate the minimum and maximum values of all nominal samples in the dataset for each channel. If any sample of an annotated event lays out of <min, max> range for any channel, we mark it as global, otherwise it is local. This approach is a bit simplistic taking into account severe distribution shifts and different nominal levels of the signal in some missions, but it should be enough to identify global anomalies which could be detected with an out-of-distribution approach from more challenging local anomalies.
3. In terms of **length**, considering non-uniform sampling rates and the differences between mission and channels, it is hard to give a strict definition of a point anomaly. One option is to make it dependent on the dominant sampling frequency for each mission (0.033 Hz for Mission1, 0.056 Hz for Mission2 and 0.065 Hz for Mission3). A point anomaly is defined as a sequence of up to 3 samples after resampling to the dominant sampling frequency. Importantly, some anomalies are fragmented into several non-overlapping annotated regions. In this case, we treat each region separately, so even if an anomaly contains several regions it can be a point anomaly if all of these regions are categorised as point anomalies.

Such automatically inferred attributes for every anomaly and rare event are given in anomaly_types.csv for each mission, taking into account annotations for all channels. However, when working with subsets of channels, only the specific subset of channels should be considered to infer anomaly types. For this purpose, the script infer_anomaly_types.py is available in the code repository. The attributes are not inferred for communication gaps and invalid fragments.

## Metrics and hierarchical evaluation

The selection of metrics and evaluation pipeline is a crucial step in establishing a reliable benchmark. Our selection is based on the close cooperation between SOEs and ML engineers and is primarily targeted at practical aspects of mission control in ESA. Five such aspects were identified and prioritised based on their importance for SOEs. They are listed in Table 4 together with the metrics used to assess them. Importantly, each metric was designed to focus solely on a single specific aspect, in the maximum isolation from the other factors. There are several reasons for this, 1) to improve the interpretability of results by avoiding complex metrics assessing multiple aspects at once, 2) to allow researchers from different domains to easily reorder or discard priorities, and 3) to enable the hierarchical evaluation pipeline. In the hierarchical evaluation pipeline, algorithms are compared for one aspect at a time, from the highest to the lowest priority. The process continues to the next aspect only if the algorithms are equal in terms of the previous aspect. This kind of evaluation has three important practical advantages, 1) it puts a strong emphasis on the priorities suggested by SOEs, 2) there is no need to select the weights of specific aspects, and 3) it saves computational time by calculating only the necessary metrics.

The highest priority aspect relates to the proper identification of anomalous events, but with a strong emphasis on avoiding false alarms at the same time (aspects 1a. "No false alarms" and 1b. "Anomaly existence" in Table 4). This is because false positives are costly to resolve and deter operators from using the system. A high false positive rate is reported in the literature as the main obstacle to the wider adoption of anomaly detection algorithms in space operations[2]. This fact additionally supports our idea of hierarchical evaluation, since a high false positive rate disqualifies an algorithm even if it obtains perfect scores in other aspects. Moreover, many other aspects focus only on performance for true positive detections (i.e. channel identification, alarming precision, timing quality), so they indirectly depend on the anomaly existence aspect.

The second highest priority for SOEs is to have the information about subsystems (aspect 2a. “Subsystems identification”) and channels (aspect 2b. “Channels identification“) affected by anomalies. Proper subsystem identification is more important for SOEs as it gives a more concise overview of the situation than a long list of specific affected channels. Again, it is of paramount important to avoid false positives. It is strongly preferable to miss some channels rather than to wrongly identify many irrelevant channels. ESA-AD contains tens of target channels which is already hardly manageable for manual analysis, moreover, it is just a fraction of channels from actual missions. Hence, an algorithm which does not provide affected channels is of low practical utility, or even worse, it may amplify the “black box” nature of advanced algorithms and decrease trust in this kind of system among operators. That is why it was considered as a part of two *primary* aspects of highest priority.

The following 3 *secondary* aspects are not so crucial for SOEs but are certainly useful to differentiate between algorithms having the same *primary* scores. The 3$^{rd}$ priority is to avoid algorithms that frequently repeat alarms for the same continuous anomaly segment (aspect 3. “Exactly one detection per anomaly”). It is strongly connected to the highest priority (1a. “No false alarms”), because even if one considered these repeated alarms as “true positives”, they would be annoying and confusing to operators, nearly as badly as false positives. The last 2 priority levels directly relate to the anomaly detection timing. It is obviously better to detect anomalies earlier than later (aspect 4. “Detection timing”), it is preferable to detect a whole time range of an anomaly instead of just a part of it, and, in case of false detections, it is better to show them close to real anomalies (aspect 5. “Anomaly range and proximity”). These aspects are often highly emphasised in TSAD benchmarks from the literature, i.e. NAB[47] and Exathlon[48]. However, they are relatively less important for on-ground mission control. Additionally, the latter aspect cannot be precisely assessed due to the mentioned problems with the objective identification of some anomaly end times.

Despite many years of research in the domain, there is no consensus on a reliable and unified set of TSAD metrics. Many recent advances criticise popular sample-wise and point-adjust protocols for being overoptimistic and propose better alternatives[16,17,19,21–23,49–55]. Some of these latest recommendations are directly applied in ESA-ADB, i.e. the corrected event-wise F-score[16] and affiliation-based F-score[22]. Besides that, there are several constraints on the selection of metrics arising from SOEs needs. Metrics must operate on binary detections, so threshold-agnostic metrics based on continuous anomaly scores (for example, areas under curves) cannot be used. Due to irregular timestamps and varying sampling rates, metrics must operate in the time domain instead of the samples domain, so the evaluation is independent of the algorithm-specific resampling. Each metric must be adapted to give a single score for multivariate anomalies. The computational complexity of metrics calculation also matters when dealing with large datasets. Based on that aspect, metrics with complexities higher than quadratic such as VUS[53] are rejected, so the evaluation could run in a reasonable time.

The definitions of the proposed metrics are given in the following subsections. All implementations are available in the published code. All metrics are defined in the <0, 1> range where 1 is the perfect score. All metrics give an option to include only specific events in the calculation. In default, only communication gaps are excluded, but in Supplementary Material 4.2 this feature is used to calculate results for anomalies only. Technical details of the implementations are listed in the Supplementary Material 3.2.2.

Table 4. **Priority aspects and proposed metrics for assessing algorithms in ESA-ADB**.

| Group | Aspect with priority level and brief description | Proposed metric |
|---|---|---|
| *Primary* | 1a. **No false alarms** – minimise the number of false detections | Corrected event-wise F0.5-score |
| | 1b. **Anomaly existence** – maximise the number of correctly detected anomalies | |
| | 2a. **Subsystems identification** – find a list of affected subsystems | Subsystem-aware F0.5-score |
| | 2b. **Channels identification** – find a list of affected channels | Channel-aware F0.5-score |
| *Secondary* | 3. **Exactly one detection per anomaly** – avoid multiple detections for the same annotated segment | Event-wise alarming precision |
| | 4. **Detection timing** – determine the anomaly start time as precisely as possible | Anomaly detection timing quality curve (ADTQC) |
| | 5. **Anomaly range and proximity** – find the exact duration of the anomaly and promote detections in close proximity to the ground truth | Modified affiliation-based F0.5-score |

### Corrected event-wise F-score

Event-wise F-score promoted for satellite telemetry by Hundman et al.[2] has two features that make it better suited for practical applications than the classic sample-wise (or time-wise) approach, 1) all anomalies have the same weight independent of their length (in practice, short anomalies may be even more important than long ones which are easier to spot manually), and 2) the metric value does not depend on a level of overlap of detections and ground truth (in practice, it is usually enough to give an approximate location of the anomaly to human operators). However, the classic event-wise precision has one serious flaw – an algorithm that simply detects anomalies in every sample in the dataset would have a perfect score (see Supplementary Fig. 11). To mitigate this, Sehili and Zhang[16] proposed to involve the true negative rate (TNR) at the sample level (at the time level in our case) in the computation of the event-wise precision. Such a corrected event-wise precision $Pr_{e_{corr}}$ is defined by equation (1),

$$Pr_{e_{corr}} = \frac{TP_e}{TP_e + FP_e} \cdot TNR_t, \quad TNR_t = \frac{TN_t}{N_t}, \tag{1}$$

where $TP_e$ is the number of event-wise true positives, $FP_e$ is the number of event-wise false positives, $TN_t$ is the number of nanoseconds with true negatives, and $N_t$ is the number of nominal nanoseconds. Based on that, the corrected event-wise $F_\beta$-score is defined by equation (2):

$$F_{\beta_{e_{corr}}} = (1 + \beta^2) \frac{Pr_{e_{corr}} \cdot Rec_e}{\left(\beta^2 \cdot Pr_{e_{corr}}\right) + Rec_e}, \quad Rec_e = \frac{TP_e}{TP_e + FN_e} \tag{2}$$

The factor $\beta$ gives us the flexibility to control the relative importance of recall. Betas lower than 1 are preferred in our case to weigh precision (fewer false positives) higher than recall (fewer false negatives). It is challenging to objectively select a specific $\beta$, so following Hundman et al.[2] the value of 0.5 was agreed as a good baseline. However, $\beta$ can be adjusted to specific operational needs.

For multivariate anomalies, the metric is calculated between logical sums of annotations and detections across all target channels. In rare cases where multiple events overlap in time, each event is analysed separately, i.e. separate true positives (max. 1) and false negatives (max. 1) are counted for each event.

### Subsystem-aware and channel-aware F-scores

Typical TSAD metrics are applicable only in univariate settings with a single series of ground truth annotations and detections as input. To get a single score for multiple channels, there must be some aggregation performed, either across annotations/detections or across scores for individual channels. Such aggregation loses information about the performance for individual subsystems or channels, so it is impossible to assess their correct identification. In recent articles[56,57], special *anomaly diagnosis* metrics are proposed to address this issue, namely HitRate and Normalised Discounted Cumulative Gain (NDCG). These metrics measure how relevant are the detected channels according to the list of annotated channels. However, they need information about the relative relevance of detections which is not available when using binary outputs. Thus, a new *anomaly diagnosis* approach is proposed based on precisions and recalls of identifying the list of affected subsystems and channels.

SOEs inspect potential anomaly sources at two levels of detail. First, they check which subsystems are affected by the anomaly. Later on, they look at the specific channels affected in those subsystems. The usefulness of algorithms supporting such inspection is proposed to be measured with subsystem-aware (SA) and channels-aware (CA) F-scores. A subsystem is counted as true positive $TP_{SA}$ if it has at least one annotated channel and at least one detected channel (not necessarily the same) overlapping with the full time span of the anomaly (logical sum of annotations across all channels in all subsystems). A subsystem is considered false negative $FN_{SA}$ if it has at least one annotated channel but no such detections. A false positive subsystem $FP_{SA}$ has no annotated channels but has at least one such detection. Thus, the subsystem-aware F-score $F_{\beta_{SA}}$ is given by equation (3):

$$F_{\beta_{SA}} = (1+\beta^2)\frac{Pr_{SA} \cdot Rec_{SA}}{(\beta^2 \cdot Pr_{SA}) + Rec_{SA}},$$
$$Pr_{SA} = \frac{TP_{SA}}{TP_{SA} + FP_{SA}}, \qquad Rec_{SA} = \frac{TP_{SA}}{TP_{SA} + FN_{SA}} \tag{3}$$

The channel-aware F-score $F_{\beta_{CA}}$ is defined analogously, but an annotated channel is counted as $TP_{CA}$ if it has any overlapping detection in the full time span of the anomaly. An annotated channel is counted as $FN_{CA}$ if there is no such detection. A false positive channel $FP_{CA}$ has no annotation but at least one such detection.

Again, 0.5 is used for $\beta$ as a baseline to be consistent with the event-wise F-score. In rare cases where multiple events overlap in time, each event is analysed separately, i.e. separate true positives (max. 1) and false negatives (max. 1) are counted for each event. Moreover, any false positives related to correct detections of other overlapping anomalies are discarded, see Supplementary Material 3.2.1 for a detailed example. For the lightweight subsets of channels selected from a single subsystem, the subsystem-aware F-score is not reported.

### Event-wise alarming precision

The corrected event-wise F-score counts only a single true positive even if there are multiple separated detections for the same fragment in the ground truth (see Supplementary Fig. 12). In practice, such redundant detections may be considered separate alarms which may be annoying for operators. The event-wise alarming precision $Pr_A$ defined by equation (4) measures the ratio of correctly detected events ($TP_e$) to the sum of correctly detected events and redundant alarms ($TP_r$):

$$Pr_A = \frac{TP_e}{TP_e + TP_r} \tag{4}$$

This metric may seem too strict in some cases, i.e. for many short detections very close to each other, but it represents practical aspects of mission operations and encourages for applying better thresholding or postprocessing approaches to avoid redundant alarms.

### Anomaly detection timing quality curve (ADTQC)

The goal of this novel metric is to assess the accuracy of the anomaly start time identification from the SOEs point of view. Some existing metrics of the anomaly detection latency, such as the After-TP[21] or the Early Detection (ED)[58], assume that an anomaly can be detected only within its ground truth interval — *after* it appears in the signal. However, the question arises how to assess algorithms that detect anomalies too early — *before* they start. They cannot be assessed using After-TP or ED metrics but they certainly have some value. The Before-TP metric[21] and the NAB score[47] rank earlier anomaly *predictions* (to distinguish them from *detections*) as better. However, in practice, as suggested by SOEs, too-early detections may be seen as false positives by operators if they cannot confirm the existence of an anomaly within a definable time. Thus, too early detections may decrease operators' trust in an algorithm and, in this context, are much worse than late detections of comparable distance from an anomaly start time. According to SOEs, the quality of anomaly detection timing should decrease exponentially for detections before the actual start time as opposed to much slower degradation of quality for moderately late detections. A survey was conducted and confronted across SOEs from different missions in ESA and KP Labs to define the timing quality in the range from 0 to 1 as a function of detection start time. The resulting consensus reflecting the operators' point of view is presented as the *anomaly detection timing quality curve (ADTQC)* in Supplementary Fig. 14 described by equation (5):

$$ADTQC(x) = \begin{cases} 0 & , & -\infty < x \le -\alpha \\ \left(\frac{x+\alpha}{\alpha}\right)^e & , & -\alpha < x \le 0 \\ \frac{1}{1+\left(\frac{x}{\beta - x}\right)^e}, & & 0 < x < \beta \\ 0 & , & \beta \le x < +\infty \end{cases}, \quad \alpha, \beta > 0 \tag{5}$$

$$ADTQC(x) = \begin{cases} 0, & x \ne 0 \\ 1, & x = 0 \end{cases}, \quad (\alpha = 0 \land x \le 0) \lor (\beta = 0 \land x \ge 0)$$

$$\alpha = \min(anomaly\ length, anomaly\ start\ time - previous\ anomaly\ start\ time)$$

$$\beta = anomaly\ length$$

After agreeing on the shape of ADTQC, the most important issue was to select the operational range of values for which the function should return a quality higher than 0, that is, for which the detection is not useless from the practical point of view. The first straightforward step was to define detections later than the anomaly end time ($\beta$) as useless. Accordingly, detections earlier than the anomaly length from the start time were also considered useless. Hence, the shorter the anomaly the more accurately it must be detected to achieve similar quality value. In the extreme case of point anomalies, ADTQC returns a value of 1 for exact detections and 0 otherwise. It makes sense from the practical point of view for two reasons, 1) detections for short, hardly noticeable anomalies are likely to be considered false alarms if not well-timed, and 2) end times of long anomalies are usually much harder to annotate precisely than for short anomalies. Another unacceptable situation was identified when a detection is earlier than the previous anomaly start time. When anomalies are close to each other, the detection timing must be even more accurate to ensure their better separation.

The ADTQC metric value for the specific anomaly is determined by simply calculating the value of the $ADTQC(x)$ function where $x$ is the difference between the detection start time and the anomaly start time. Similarly to Before/After-TP, the metric is calculated and averaged across all correctly detected events to get a final score in the range from 0 to 1. To support the analysis of the results, the ratio of detections after the anomaly starting points to all detections is calculated (called the *after ratio*).

For multivariate anomalies, the ADTQC metric is calculated between the logical sums of annotations and detections across all target channels. It does not matter if the detections are for correct channels because the metric focuses on the timing alone. The second possible approach in the multivariate setting would be to calculate the ADTQC metric for each affected channel separately. The average across all affected channels would be the final ADTQC score for a specific anomaly. While this alternative approach would allow for more detailed quantification of the anomaly detection timing across channels, it does not reflect the operators' perspective in which the first detection is the most important one, because it already enforces an action. Later detections for any other channel do not matter so much, because operators are already aware of the potential anomaly.

### Modified affiliation-based F-score

The affiliation-based metric by Huet et al.[22] claims to resolve all the major flaws of previous range-based metrics. That is, it is aware of the temporal adjacency of samples and anomalies duration, has no parameters, and is locally and statistically interpretable (specific problematic time ranges can be easily identified and a score of 0.5 means a random prediction). The main idea is to divide the ground truth into local zones affiliated with consecutive anomaly ranges. The borders of such *affiliation zones* lie in the mid points between consecutive anomalies. Precision and recall are calculated separately for each affiliation zone based on the average directed distance between sets of annotated and detected points, either the distance from annotated to detected (precision) or from detected to annotated (recall). This way it is easy to analyse which zones are the most problematic for an algorithm. Affiliation-based F-score with $\beta$ of 0.5 is calculated to underscore the strong practical need to minimise the number of false positives. The final global F-score is calculated as the arithmetic average of all local F-scores (with each affiliation zone having the same weight).

An important modification to the original implementation relates to frequent situations when it is impossible to calculate the precision in an affiliation zone (there is no detection, so there are no true positives or false positives). In the original formulation, such an affiliation zone was simply ignored when calculating an average precision over all affiliation zones. However, this approach makes it hard to robustly compare different algorithms because of the different numbers of affiliation zones taken into account, e.g. it gives a higher score to an algorithm that detects a single anomaly very precisely and misses 4 others than to an algorithm that detects all 5 anomalies relatively well – see Supplementary Fig. 13. Thus, in our formulation, empty detections get a precision of 0.5. An affiliation-based precision of 0.5 can be interpreted as a random detection, so this modification promotes algorithms that would rather give an empty detection than a false detection that is worse than random. There are also some other technical adaptations to handle point anomalies and fragmented annotations, as described in Supplementary Material 3.2.2.

## Preprocessing

Our dataset contains raw non-uniformly sampled timestamps, so only a few algorithms with positional time encoding (such as TACTiS[38]) could handle it without any resampling. The vast majority of algorithms, including all the algorithms selected for ESA-ADB, operate only on uniformly sampled time series. Additionally, there are many different types of channels, including monotonic, categorical, and binary ones, so a consistent preprocessing procedure is needed to run and compare the majority of algorithms.

### Resampling

The vast majority of algorithms, including all the algorithms selected for ESA-ADB, operate only on uniformly sampled time series. The zero-order hold interpolation (propagating the last known value) is recommended for satellite telemetry in the OXI annotation tool[45]. This interpolation method is well suited for processing binary and quantised signals that are common in satellite telemetry (i.e. telecommands and measurements from analog-to-digital converters) because, unlike the linear or Fourier-based interpolation, it does not create any artificial, impermissible values between points. More importantly, it does not use future samples to perform the interpolation which is necessary in real-time applications. This interpolation is presented in Supplementary Fig. 15 and implemented for the resampling as follows:

1. Construct a uniformly sampled list of timestamps in the target sampling frequency. Set the first/last timestamp in the list to the value of the earliest/latest original timestamp across all channels rounded down/up to the target sampling resolution. Fill the list between the first and the last element using uniformly sampled timestamps in the target frequency, e.g. if we resample a list of original timestamps <8:10:12, 8:10:14, 8:10:38> to the target frequency of 1/10 Hz (target resolution of 10 seconds), the resampled list will be <8:10:10, 8:10:20, 8:10:30, 8:10:40>.
2. Propagate the last known value and label from the original samples (zero-order hold) to each timestamp in the constructed list. If there are still any missing values for the initial element of the list (i.e. when some channels start a little earlier than others), backpropagate the first known value from the original samples. This introduces a bit of information from the future, but it usually concerns only a few samples at the beginning of a test set.

3. Apply a correction for missing anomalies to ensure that no point events are removed due to the resampling. Iterate through consecutive pairs of unannotated timestamps in the resampled list and, if there are any annotated original points in between, take the last annotated sample and assign its value and label to the latter timestamp from the pair. The result of such a correction is visible in the rightmost sample of Channel_1 in Supplementary Fig. 15.

Target sampling frequencies differ across missions. The selection was based on the analysis of the most densely sampled target channels, specifically, to prevent losing any annotated anomalies, especially point anomalies:

1. In Mission1, 0.033 Hz was selected based on the dominant sampling frequency of target channels 41-46 with some point anomalies.
2. In Mission2, 0.056 Hz was selected based on the dominant sampling frequency of all target channels. There are no point anomalies in Mission2, so there was no risk of losing point anomalies.
3. In Mission3, 0.065 Hz was selected based on the dominant sampling frequency of all target channels.

**Standardization**

Standardization is a necessary step for some algorithms (such as KNN[34]) and it may boost the performance of neural networks[59]. In our preprocessing, each channel is standardized separately to zero mean and unit standard deviation according to nominal points in a training set after resampling. However, such standardization is not performed for:

- algorithms that do not need it by definition, i.e. Isolation Forest[30] or COPOD[60],
- binary channels (any channel with only two unique values in the training data). Instead of being standardized, they are normalised to the <0, 1> range. These kinds of channels are quite common in satellite telemetry, i.e. telecommands or status flags. There are often just a few state changes, so the standard deviations may be very small and cause numerical errors,

- constant channels (with zero standard deviation). In this case, only the mean is subtracted,
- monotonic channels that are non-decreasing or non-increasing from the definition of the measured process, i.e. counters or cumulative on-times. In this case, the standardization is preceded by calculating the first difference of the resampled signal.

Channels with categorical values and status flags are enumerated according to the order of occurrence of each state in the training set and standardized. This is a very naïve approach, but it does not require laborious manual analysis of all channels and preparation of state mappings for each potential mission. Also, it does not require special handling of categorical anomalies. Moreover, categorical channels are usually non-target.

### Telecommands' encoding

TCs in the original data are represented by lists of timestamps at which specific TCs were executed on board a satellite. For purposes of ESA-ADB, they are encoded as binary impulses of a single sample length according to the target resampling resolution, so they are not removed by the proposed resampling.

## Algorithms

There are several recent comprehensive reviews of approaches for TSAD[15,17,18,25,49,54,61] that list hundreds of TSAD algorithms. They can be divided into several groups according to the type of learning (supervised, unsupervised, semi-supervised, weakly-supervised), the origin of a method (classic machine learning, signal analysis, data mining, stochastic learning, outlier detection, statistics, deep learning)[15], supported dimensionality (univariate and multivariate), and the main mechanism of anomaly detection (forecasting, reconstruction, encoding, distance-based, distribution-based, decision trees, and rule-based systems)[15]. It is technically infeasible to implement and include all algorithms in ESA-ADB, so the selection had to be performed based on substantive arguments.

The most fundamental and widely used approach for anomaly detection in spacecraft systems is based on checking whether sensor values are within a predetermined nominal range. This out-of-limits method has many advantages (i.e. simplicity, explainability, speed, minimal computational requirements) and works well in surprisingly many situations. However, this approach does not perform and scale well with the exponentially growing complexity of spacecrafts (see Supplementary Fig. 2). The first data-driven machine learning approaches tried to resolve this problem based on adaptive limit checking, principal components analysis (PCA), and Bayesian networks[62]. The Novelty Detection algorithm[63] implemented in the Mission Utility and Support Tools (MUST)[1] of ESOC was one of the first proofs of concept that demonstrated the possibility of having an efficient and generic telemetry monitoring system based on machine learning. This initiative was noticed by the space operations community and encouraged many entities to experiment with similar methods, including NASA[2], CNES[6], DLR[7], JAXA[8], and Airbus, among others[9–12]. In recent years, an unprecedented success of deep learning (DL) was observed in virtually all domains of science and industry, also encompassing an array of space-related issues that relate to solving the problem of fuel-optimal landing[64], designing the solar-sail trajectory for near-Earth asteroid exploration[65], predicting risk of satellite collisions[66], and many more[67,68]. The notable DL-based algorithm designed for satellite telemetry anomaly detection is the semi-supervised, forecasting-based Telemanom based on recurrent neural network (RNN) with Long Short-Term Memory (LSTM) units which established a baseline for all later related works, mainly because of the introduction of NASA SMAP and MSL datasets. Many general-purpose TSAD algorithms have been validated on those datasets, but their results are not indicative, because these datasets are widely criticised in recent publications[17,18,69].

Based on our review of existing TSAD frameworks and benchmarks, the TimeEval framework[29] (github.com/TimeEval) was selected as the foundation for ESA-ADB. It offers more than 70 implementations of TSAD algorithms of various types and a complete evaluation pipeline. Its authors thoroughly tested it on real-life and simulated data[15]. For purposes of ESA-ADB, it was extended with several new algorithms (GlobalSTD, Telemanom-ESA, DC-VAE[28], DC-VAE-ESA), metrics, and evaluation mechanisms. Importantly, default evaluation procedures for unsupervised algorithms in TimeEval do not include any separate training step on a training set. The algorithms are both trained and run on a whole dataset. This is a typical setting for outlier detection tasks. However, this is not a realistic approach in online satellite telemetry monitoring, and it would give an unfair advantage to unsupervised algorithms, because of the information leakage from future samples. The framework and internal implementations of some algorithms were modified, so each unsupervised algorithm is first trained on the training set only (including calculating contamination levels, setting thresholds, and standardization parameters) and then utilised to detect anomalies in the test set in an online manner (without using future samples from the test set).

There are nine technical requirements for anomaly detection algorithms in satellite telemetry. The first two are necessary to conform with the primary needs of SOEs ("**shall**", following the wording recommended by the European Cooperation for Space Standardization ecss.nl/standard/ecss-e-st-10-06c-technical-requirements-specification/) and the rest are recommended in practical applications ("**should**"):

R1. Algorithm **shall** provide a binary response (i.e. 0 – nominal, 1 – anomaly). It is not enough to provide continuous anomaly scores to SOEs, so a thresholding mechanism should be a part of the algorithm. A clear boundary is needed to decide if something should be alarmed to operators or not.

R2. Algorithm **shall** allow for real-time, online, streaming detection. Although on-ground mission control usually does not work in actual real-time, because larger packets of data are collected from a satellite only during infrequent communication windows, real-time monitoring is desirable in the future of mission control and is necessary for on-board anomaly detection systems.

R3. Algorithm **should** be able to model dependencies between multiple channels. Satellite telemetry contains hundreds of interconnected channels and there are many examples of anomalies that can be detected only when using information from multiple channels at once.

R4. Algorithm **should** learn from anomalies in training and validation data.

R5. Algorithm **should** provide a list of channels affected by a detected anomaly. It is a crucial aspect of practical applications in mission control.

R6. Algorithm **should** distinguish between target channels, non-target channels, and telecommands, so it learns from all sources, but only anomalies in target channels are reported to SOEs.

R7. Algorithm **should** learn to distinguish rare nominal events, so they are not alarmed after the first occurrence.

R8. Algorithm **should** natively handle irregular timestamps and varying sampling rates, without the need for additional resampling. Typical resampling schemes make algorithms unaware of varying gap lengths between points which may lead to many false anomaly detections.

R9. Algorithm **should** be possible to run in a reasonable time on a single high-end PC. The specific limits are listed in Supplementary Material 4.5. The algorithm is not included in our benchmarking results if they are not met, so it is effectively **shall** in our case.

Based on the initial requirements analysis, 20 algorithms were preselected among those available (or added) in the TimeEval framework that at least partially fulfil all primary requirements. Table 5 summarises the detailed requirements analysis for those algorithms. Some examples of partially fulfilled requirements are for algorithms that R1) do not provide dedicated thresholding mechanisms, R2) technically allow for the online detection but with a large computational overhead, R4) handle anomalies in training data but cannot learn from them, R5) would need additional mechanisms or modifications of external libraries (i.e., PyOD[70]) to provide a list of affected channels, R7) give only a theoretical option to learn rare nominal events, or R9) are only possible to run for the lightweight subsets of channels (i.e. Windowed iForest and KNN). None of the preselected algorithms are able to explicitly learn rare nominal events (R7) or handle varying sampling rates (R8).

Based on the detailed analysis of the requirements, eight algorithms of various types were selected for ESA-ADB, five unsupervised – principal components classifier (PCC)[32], histogram-based outlier score (HBOS)[33], isolation forest (iForest)[30], k-nearest neighbours (KNN)[34], and three semi-supervised ones – global standard deviations from nominal (GlobalSTD), Telemanom[2], and DC-VAE[28]. The selected unsupervised algorithms have several important limitations in terms of TSAD. They may be give suboptimal results because of the assumptions of independence of samples and identical fractions of anomalies in training and test data (they fulfil R4 because they learn contamination levels from the training data). They only give global scores, so it is impossible to calculate subsystem-aware and channel-aware scores for them. They also do not support non-target channels and telecommands on input, so this information was not used. However, they establish a baseline for more advanced algorithms.

Among the rejected ones, Matrix Profile-based methods like DAMP[36] or MADRID[37] seem to be promising candidates due to their outstanding speed, high interpretability, and a theoretical possibility to memorise rare nominal events. However, they would need a special adaptation to support multidimensional data[71], they do not natively handle anomalies in training, and their implementations in Matlab pose several technical and licensing problems when integrated with TimeEval. The COPOD algorithm does not fulfil R9 after adapting it to online detection required by R2. LOF[72], k-Means[73], Torsk[74], and RobustPCA[75] showed very poor results in initial experiments. All semi-supervised algorithms that only partially fulfil R9 were rejected.

Table 5. **Analysis of preselected algorithms according to ESA-ADB requirements**. 0/0.5/1 means that the requirement is not/partially/fully fulfilled. Asterisks mark new methods added to the TimeEval. Bold-faced requirements are "**shall**".

| | Algorithm | **R1** | **R2** | R3 | R4 | R5 | R6 | R7 | R8 | **R9** | Included in ESA-ADB |
|---|---|---|---|---|---|---|---|---|---|---|---|
| UNSUPERVISED | COPOD[60] | 1 | 0.5 | 1 | 1 | 1 | 0 | 0 | 0 | 1 | NO |
| | HBOS[33] | 1 | 1 | 0 | 1 | 0.5 | 0 | 0 | 0 | 1 | YES |
| | iForest[30] | 1 | 1 | 1 | 1 | 0.5 | 0 | 0 | 0 | 1 | YES |
| | Windowed iForest[30] | 1 | 1 | 1 | 1 | 0.5 | 0 | 0 | 0 | 0.5 | SUBSETS |
| | k-Means[73] | 1 | 1 | 1 | 1 | 0.5 | 0 | 0 | 0 | 0.5 | NO |
| | KNN[34] | 1 | 1 | 1 | 1 | 0.5 | 0 | 0.5 | 0 | 0.5 | SUBSETS |
| | LOF[72] | 1 | 1 | 1 | 1 | 0.5 | 0 | 0 | 0 | 0.5 | NO |
| | Matrix Profile[36,37] | 1 | 1 | 0.5 | 0 | 0.5 | 0 | 0.5 | 0 | 1 | NO |
| | PCC[32] | 1 | 0.5 | 1 | 1 | 0.5 | 0 | 0 | 0 | 1 | YES |
| | Torsk[74] | 0.5 | 1 | 1 | 1 | 1 | 0 | 0 | 0 | 0.5 | NO |
| SEMI-SUPERVISED | DAE[76] | 0.5 | 1 | 1 | 0 | 1 | 0 | 0 | 0 | 0.5 | NO |
| | DC-VAE[28]* | 0.5 | 1 | 1 | 0 | 1 | 0 | 0 | 0 | 0.5 | NO |
| | DC-VAE-ESA* | 1 | 1 | 1 | 0.5 | 1 | 1 | 0 | 0 | 1 | YES |
| | GlobalSTD* | 1 | 1 | 0 | 0.5 | 1 | 0 | 0 | 0 | 1 | YES |
| | Hybrid KNN[77] | 1 | 1 | 1 | 0 | 0.5 | 0 | 0.5 | 0 | 0.5 | NO |
| | LSTM-AD[78] | 0.5 | 1 | 1 | 0 | 0 | 0 | 0 | 0 | 0.5 | NO |
| | OmniAnomaly[26] | 0.5 | 1 | 1 | 0 | 0.5 | 0 | 0 | 0 | 0.5 | NO |
| | RobustPCA[75] | 0.5 | 0.5 | 1 | 0.5 | 0.5 | 0 | 0 | 0 | 1 | NO |
| | Telemanom[2] | 1 | 1 | 1 | 0 | 1 | 0 | 0 | 0 | 0.5 | NO |
| | Telemanom-ESA* | 1 | 1 | 1 | 0.5 | 1 | 1 | 0 | 0 | 1 | YES |

The published code contains implementations of all methods listed in Table 5. New and improved algorithms introduced in ESA-ADB are described in the following subsections.

### GlobalSTD

In this classic distribution-based approach, any samples deviating from the mean of the channel by more than N its standard deviations are detected as anomalies. This approach is categorised as semi-supervised because only nominal samples (excluding rare nominal events) from the training set are used to compute means and standard deviations for each channel to avoid the influence of outliers. In practice, the threshold of 3 standard deviations (STD3) is frequently used (following the empirical statistical rule that 99.7% of data occurs within 3 standard deviations from the mean within a normal distribution[79]), but it may not be optimal when the number of false positives should be minimised, so the threshold of 5 standard deviations (STD5) is also tested to provide a versatile baseline for other algorithms. The main disadvantage of this algorithm is that it is unable to detect local anomalies, so it is usually not a good choice in practice. It is also not aware of dependencies between channels and it is very vulnerable to changes in the data distribution during the mission. It also cannot use the information about non-target channels and telecommands.

**Telemanom-ESA**

This semi-supervised algorithm proposed by NASA engineers[2] is an important point of reference in the domain. It can be considered the most popular algorithm for anomaly detection in satellite telemetry. Its core element is an LSTM-based RNN that learns to forecast a small number of time points (10 by default) for a single channel based on the hundreds of preceding samples (250 by default) from multiple input channels. The mean absolute difference between the forecasted samples and the real signal is treated as an anomaly score, which is thresholded using the non-parametric dynamic algorithm (NDT) to find anomalies. However, this "non-parametric" approach (in the sense that it does not use Gaussian distribution parameters to estimate thresholds) has several hyperparameters. In one of our previous works, a genetic algorithm was used to find optimal hyperparameters of thresholding for NASA SMAP and MSL datasets[80]. However, this wrapper approach would be too computationally expensive to run on our large datasets, so the default settings proposed by the authors are used.

Telemanom has several major issues that had to be addressed for the purposes of ESA-ADB.

- **Memory inefficiency** – Telemanom was designed for small and simplified datasets provided by NASA. Hence, the code is not optimised to handle very large datasets and it results in out-of-memory errors, e.g. there are many unnecessary copies of data, all training windows are loaded into memory at once, and binary annotations are loaded to memory as floating-point numbers. **Telemanom-ESA:** The code is optimised for memory consumption by using lazy generators to prepare training batches, in-place operations instead of copying data to new variables, and optimised data types.

- **Magic numbers in thresholding** – there are several conditions in the thresholding code that are not well documented in the original article. Especially impactful is that windows with smoothed errors below 0.05 are never anomalous (github.com/khundman/telemanom/blob/26831a05d47857e194a7725fd982d5dea5402dd4/telemanom/errors.py#L339). This is a very data-specific condition that is not well-suited for channels with certain signal values. **Telemanom-ESA**: This specific condition was removed from the code. **Telemanom-ESA-Pruned**: The threshold of 0.05 is much too high for ESA-ADB, so it was changed to 0.007 based on the manual analysis of smoothed errors in the training data of both missions. This selection is highly subjective and is probably not optimal, but allows to assess the effect of such a pruning on the results.
- **No proper handling of anomalies in training data** – Telemanom assumes that there are no anomalies in the training set which is not true in our real-life setting. **Telemanom-ESA**: only continuous nominal parts longer than 260 samples and without any anomalies in any target channel are used for training and validation.
- **Only a single output from the LSTM model** – a single Telemanom model can take multiple input channels but it always outputs a prediction for a single target channel. This is a significant shortcoming when scaling this approach to hundreds of channels and gigabytes of data. The training of a single model may last hours or days, so training separate models for tens of channels can take months on a single PC. Also, it is impossible to provide different sets of input (non-target channels, telecommands) and output (target) channels. **Telemanom-ESA**: the output of Telemanom is extended, so that is possible to forecast any number of channels at once from a single model, like in DC-VAE[28]. The channels are still analysed separately, but there is no need to train a separate model for each channel.
- **Problems with GPU support** – the original implementation of Telemanom is based on TensorFlow version 2.0 which does not natively support the CUDA compute capability 8.6 of our Nvidia GPUs. Also, the TimeEval framework lacks GPU support. **Telemanom-ESA**: TensorFlow is upgraded to version 2.5 and the GPU support is added to the TimeEval.

**DC-VAE-ESA**

DC-VAE (Dilated Convolutional – Variational Auto Encoder)[28] is one of the latest published multivariate TSAD algorithms. It is a reconstruction-based method that relies on dilated convolutions to capture long and short-term dependencies without using computation- and memory-intensive multi-layer RNNs. Unlike the original Telemanom, it outputs multiple channels from a single model and does not need a complicated thresholding scheme, because it also estimates nominal standard deviations for each sample in each channel, so that thresholding can simply be applied by looking for real samples exceeding reconstructions by more than N standard deviations. In the original implementation, N is selected from integers between 2 and 7 to maximise the range-based F1-score[81] for each channel in the training set. This approach does not scale well with the number of channels and assumes the similarity of anomalies between the training and test sets. Thus, in DC-VAE-ESA, only two values of N are considered, 3 (STD3) and 5 (STD5). The DC-VAE paper introduces the TELCO dataset, which has three rare features also promoted by ESA-ADB, i.e. separate annotations for each channel, anomalies in training sets, and the idea of gradually increasing training set sizes. Hence, the modified DC-VAE-ESA introduces only two small technical improvements to fully cover 7 of the 9 mentioned requirements:

- an option to handle different numbers of input and output channels,
- L2 regularisation of convolutional layers with the 0.001 rate to stabilise the training of VAE in the presence of concept drifts,

## 5. Data availability

The dataset is publicly available at https://doi.org/10.5281/zenodo.12528696 under CC BY 3.0 IGO license.

## 6. Code availability

The code is publicly available at https://github.com/kplabs-pl/ESA-ADB under the MIT license.

## 8. Acknowledgements

The work was financially supported by the European Space Agency as a part of the “ESA Anomalies Dataset for International AI Anomaly Detection Benchmark” project (contract number 4000137682/22/D/SR). We would like to thank all SOEs from the European Space Operations Centre involved in the project, especially Alessandro Donati and Felix Voelpel who supported this project at different stages. We are grateful to Alicja Musiał, Szymon Rogoziński, and Dawid Lazaj from KP Labs for their valuable insights into the evaluation process.

## Supplementary Material

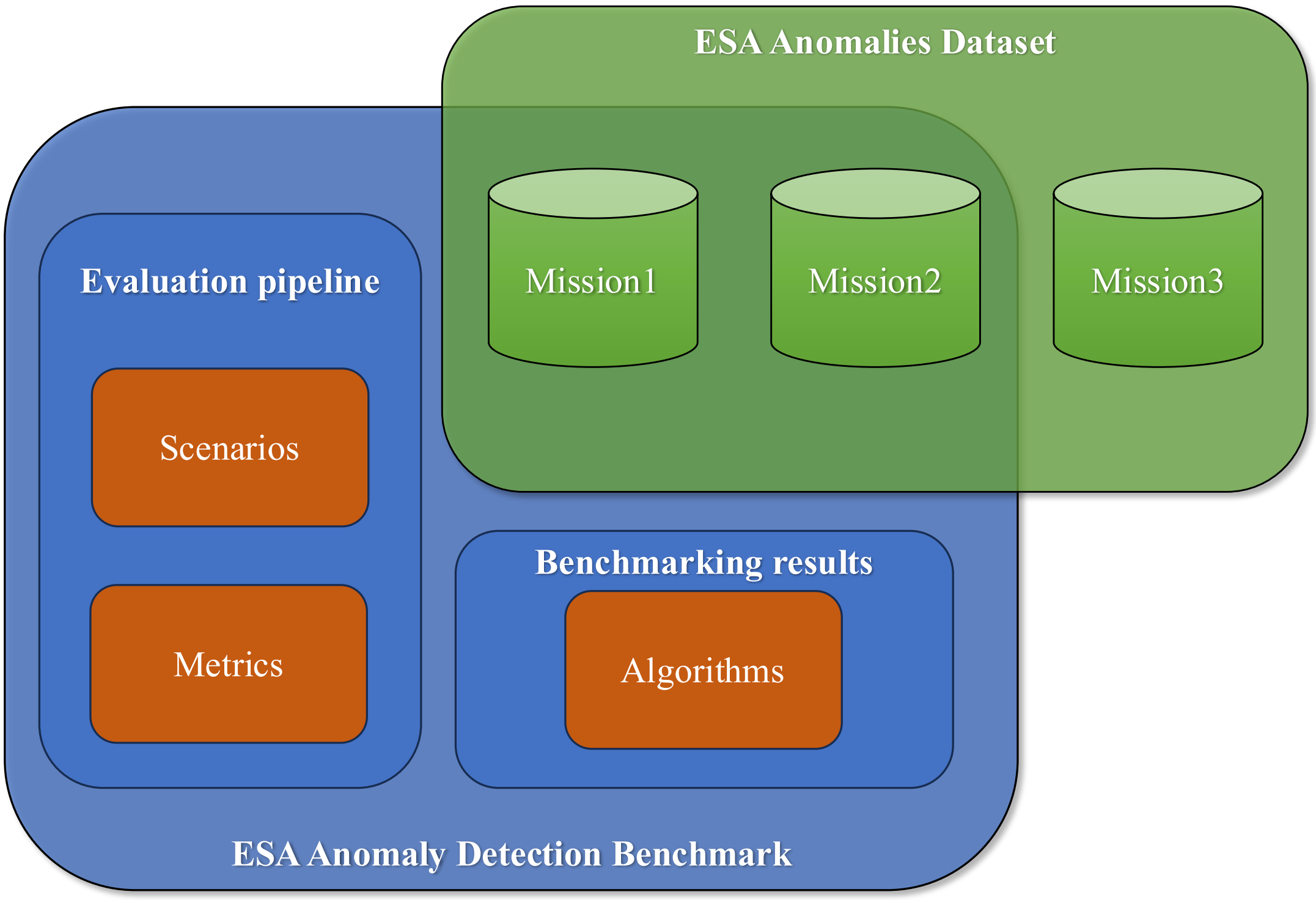

Supplementary Fig. 1. **Main elements of ESA-ADB and relationships between ESA-ADB and ESA-AD**.

### 1. Definitions

#### 1.1. Channel vs parameter

Satellite telemetry consists of multiple time series that are called *parameters* by SOEs. This name is very problematic from the ML point of view because it collides with its fundamental nomenclature in which the *parameter* already has a couple of different meanings:

- a *parameter* of the model that is updated during the training, i.e. a single weight of the neural network
- a *parameter* (or hyperparameter) of the algorithm which controls its behaviour
- a *parameter* of a statistical test (e.g. mean or variance of the estimated Gaussian distribution)

Hence, the *parameter* was replaced with the *channel* for purposes of ESA-ADB to avoid potential nomenclature collisions. *Channels* represent measurements from different sensors, status flags, and payload-related information. Each channel contains a list of samples defined by pairs of timestamps and signal values.

### 1.2. Subsystems

Satellites are typically composed of multiple specialized parts (subsystems) including propulsion, electrical power, thermal control, attitude and orbit control, communication, and data handling subsystems. There are also unique satellite-specific subsystems in some missions. A subsystem gathers all components (and channels) responsible for a specific function.

### 1.3. Telecommands

Telecommands (TCs) are sent from the Earth to the satellite in order to control different aspects of its operation. There are hundreds or thousands of different TCs for each mission with millions of total executions, affecting different subsystems and specific components. Many different TCs are frequently executed simultaneously or in series to perform specific instructions. They may affect the observed telemetry in various ways, from no visible changes to strong disruptions. In our dataset, each TC is a binary signal with values of 1 in the exact timestamps of TC's executions on-board the satellite. TCs are not expected to contain any anomalies and even if they were, anomalies (e.g. missing TCs) would be impossible to identify automatically without additional expert knowledge and information about mission plans. Thus, they are not monitored nor annotated for anomalies.

### 1.4. Target and non-target channels

Not every channel can be a target for anomaly detection benchmarking. Like telecommands, some channels are not expected to contain any anomalies, and it would be impossible to annotate them without additional external data anyway. Examples include status flags, counters, and metadata, such as location coordinates. They often contain important information in the context of anomaly detection but are not monitored nor annotated for anomalies. They may contain outliers that are, however, irrelevant (or nominal) for SOEs. They are called *non-target* channels in ESA-ADB. This aspect is usually not considered in existing multivariate anomaly detection datasets and benchmarks. The selection of *target* and *non-target* channels is somewhat subjective and it may turn out that some algorithms would be able to properly handle some *non-target* channels by discovering some unknown relationships in the data. However, the metrics in ESA-ADB are calculated only for *target* channels. *Non-target* channels may and should be used as input features for algorithms.

## 1.5. Event class vs category vs type

Each annotated event can be assigned to a different class, category, and type:

- Event classes relate to main causes of events and their specific variations (subclasses) as identified by SOEs. For example, attitude disturbances (with subclasses depending on the specific cause), resets, power drops, latch-ups, solar flares, etc.
- Event categories relate to the categorisation of events from the operational point of view, i.e. anomalies, rare nominal events, communication gaps, and invalid segments, as described in the next section.
- Event types relate to the taxonomy of anomaly types introduced in Methods.

Note that each feature is independent of others, that is, events of the same class can have different categories and types, e.g. resets caused by telecommands are categorised as rare nominal events, but unexpected non-commanded resets are categorised as anomalies.

## 1.6. Event categories

For the purposes of our project, 4 categories of events are introduced: anomalies, rare nominal events, communication gaps, and invalid segments. They are defined in Supplementary Table 1. The main reason was to distinguish atypical changes in the telemetry that should not be alarmed to operators (rare nominal events, communication gaps, and invalid segments) from unexpected ones that should be alarmed (anomalies). Rare nominal events are not anomalies from the operators' point of view and they are usually not reported in anomaly tracking systems. Eventually, they are recorded in the mission log as special operations. For some missions, i.e. Mission2, there is a significant number of such operations causing (not so) rare events. Hence, the ideal algorithm should not alarm for rare nominal events, but it is usually impossible to distinguish between novel rare nominal events and anomalies without additional a priori expert knowledge. As agreed with SOEs, it would be acceptable if an anomaly detection system shows a false alarm for the first occurrence of the specific rare event, but it should not alarm for any subsequent occurrences of similar rare events. In machine learning, we can define that problem as active one-shot learning. To enable evaluation in such a scenario using ESA-AD, it is necessary to distinguish rare events from anomalies in ground truth annotations. Besides, such a division allows us to calculate separate performance metrics for rare events and "real" anomalies. It also helps to interpret the results in case of false negative or false positive detections for rare events.

Supplementary Table 1. **Definitions of event categories**.

| **Event category** | **Definition** | **Typical examples** | **Alarming** |
|---|---|---|---|
| **Anomaly** | Atypical, rare, unplanned, and unwanted change in the telemetry. | Micrometeorite impacts, solar flares, hardware or software failures, latch-ups, decontaminations, unexpected attitude disturbances, unexpected responses to telecommands | Every occurrence should be alarmed. |
| **Rare nominal event** | Atypical and rare but expected or planned change in the telemetry. It can be triggered by known telecommands (commanded rare event) or by any other non-commanded special event in the mission timeline. | Commanded: manoeuvres, resets, calibrations, switching devices on/off<br><br>Non-commanded: planned autonomous operations, eclipses, lunar transitions | Only the first occurrence of a rare nominal event from each class may be alarmed. Subsequent occurrences should not be alarmed. |
| **Communication gap** | Unusually long gap in the telemetry (missing data in some or all channels) not directly related to known anomalies. | Problems with the ground infrastructure, effects of resets | It should not be alarmed unless explicitly stated to do so. |
| **Invalid segment** | Fragment of telemetry data containing invalid or forbidden values not directly related to known anomalies. It is neither nominal nor anomalous. | Telemetry does not meet clearly defined validity rules of the mission. | It should not be alarmed unless explicitly stated to do so. |

## 2. ESA Anomalies Dataset

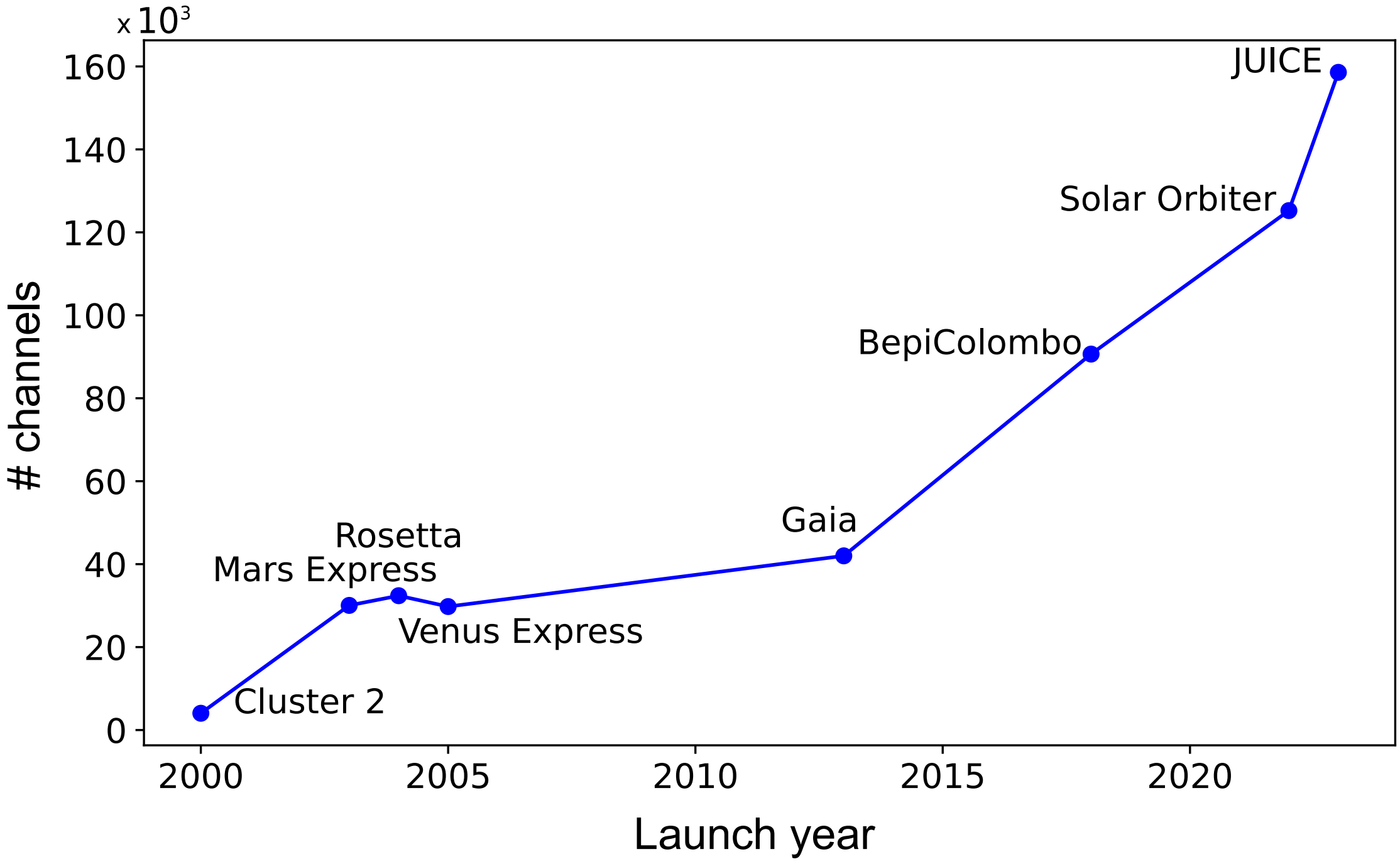

Supplementary Fig. 2. **Increasing complexity of selected ESA spacecrafts over time**[20].

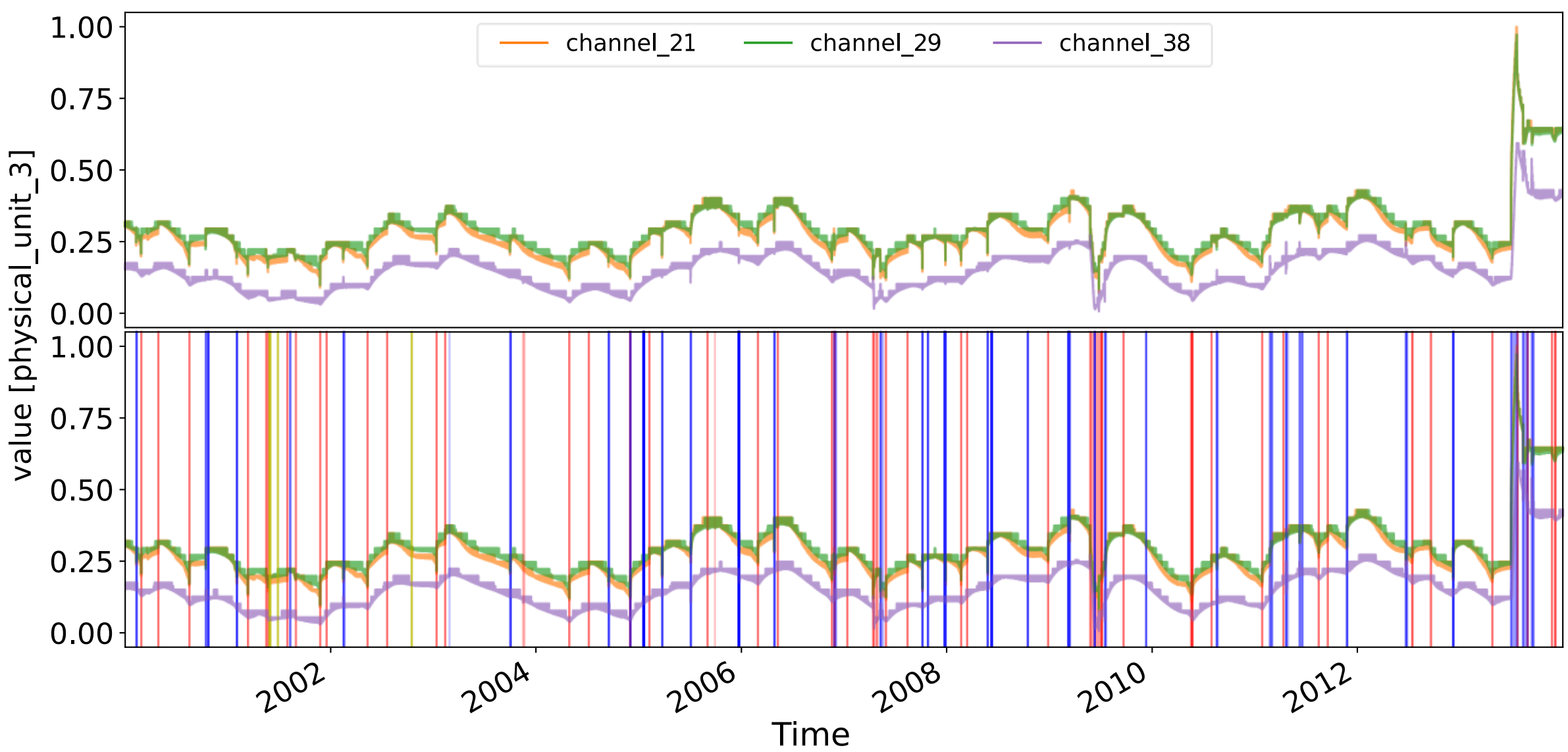

Supplementary Fig. 3. **Overview of 3 channels from group 4 in Mission1 without annotations (top panel) and annotated (bottom panel).** Blue, yellow, and red vertical bars are rare nominal events, communication gaps, and anomalies, respectively.

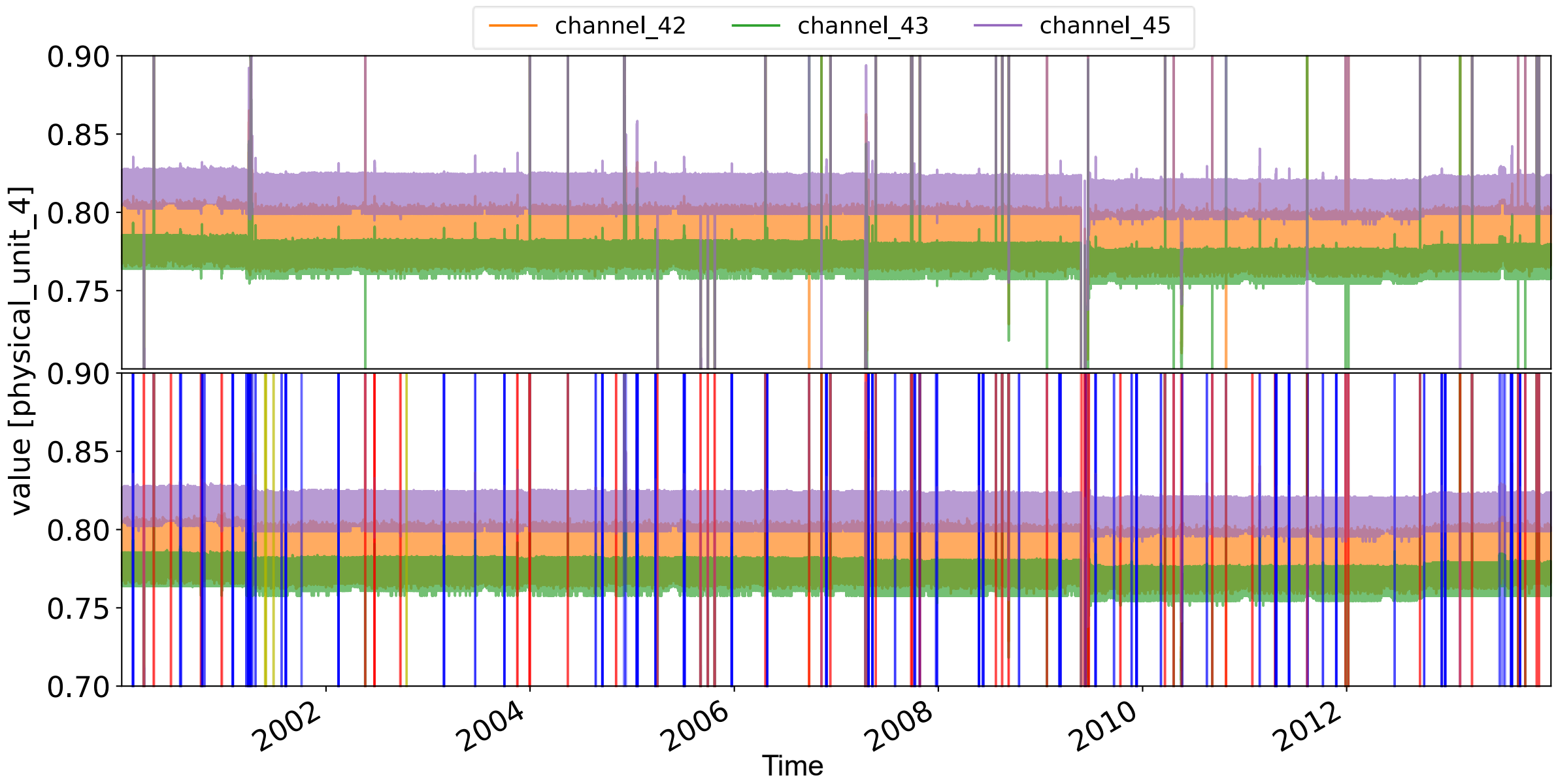

Supplementary Fig. 4. **Overview of 3 channels from group 8 in Mission1 without annotations (top panel) and annotated (bottom panel).**. Blue, yellow, and red vertical bars are rare nominal events, communication gaps, and anomalies, respectively. The close-up of these channels is presented in Supplementary Fig. 9.

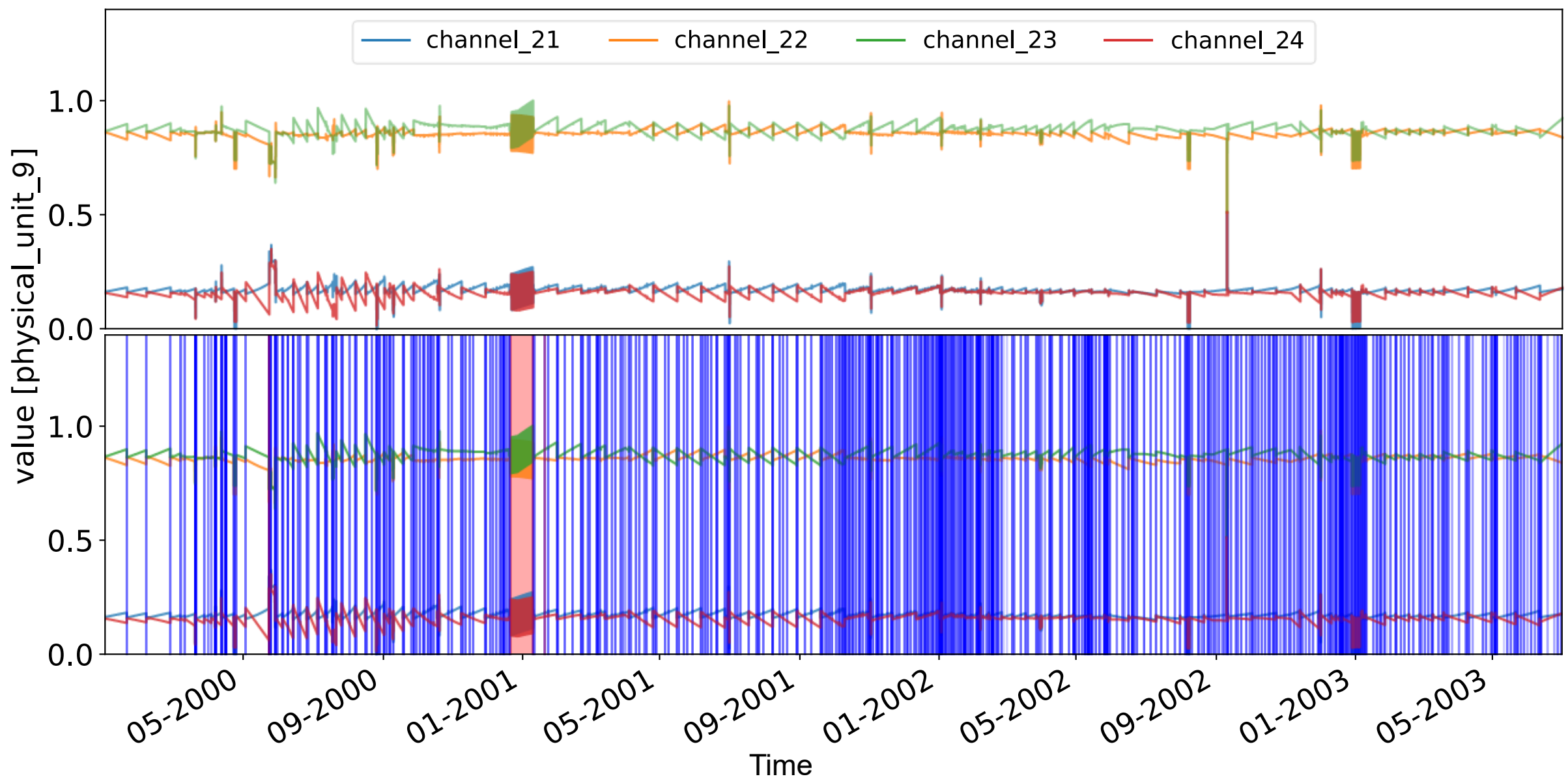

Supplementary Fig. 5. **Overview of 4 channels from group 2 in Mission2 without annotations (top panel) and annotated (bottom panel).** Blue and red vertical bars are rare nominal events and anomalies, respectively.

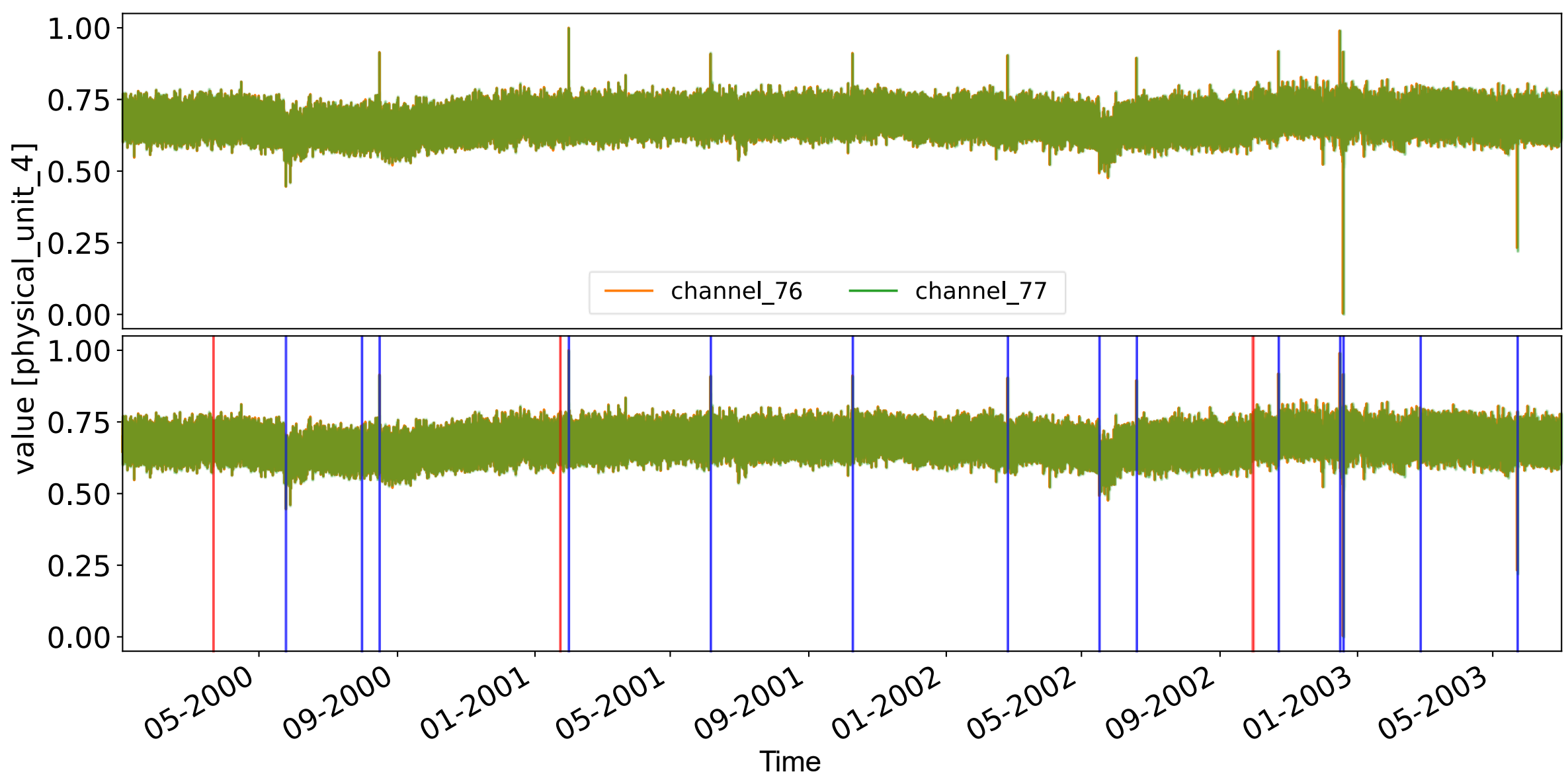

Supplementary Fig. 6. **Overview of 2 channels from group 31 in Mission2 without annotations (top panel) and annotated (bottom panel).** Blue and red vertical bars are rare nominal events and anomalies, respectively. Note that channels have very similar values, so it is hard to distinguish them.

Supplementary Table 2. **Main challenges posed for algorithms by missions in ESA-ADB**.

| Mission | Main challenges for algorithms |
|---|---|
| 1 | • Several anomalies are hard to spot (see Supplementary Table 4).<br>• Several huge outliers (usually related to rare nominal events)<br>• Low signal-to-noise ratio in channels from group 8<br>• Monotonically non-decreasing signals in channels from group 2<br>• Last 18 months include a severe concept drift in channels from groups 4, 7, and 13<br>• There is a visible seasonality with a very long period length<br>• Overabundance of telecommands |
| 2 | • Several anomalies are hard to spot when looking at individual channels only (see Supplementary Table 4).<br>• Overabundance of rare nominal events and a very small number of anomalies<br>• No obvious periodicity of the signal<br>• Monotonically non-decreasing signals in channels from group 20<br>• Many categorical and non-target channels |

### 2.1. Mission3

Mission3 is a part of ESA-AD but is omitted in ESA-ADB. Its statistics are presented in Supplementary Table 3. It was omitted mainly because of many communication gaps (see Supplementary Fig. 7), invalid segments (corrupted data), long periods of constant signals, lack of telecommands, and a small number of anomalies that are trivial to detect according to Definition 1 of Wu & Keogh[18]. However, it may still be an interesting resource for practitioners in the domain as it is fully annotated and contains a unique set of challenges related to satellite telemetry.

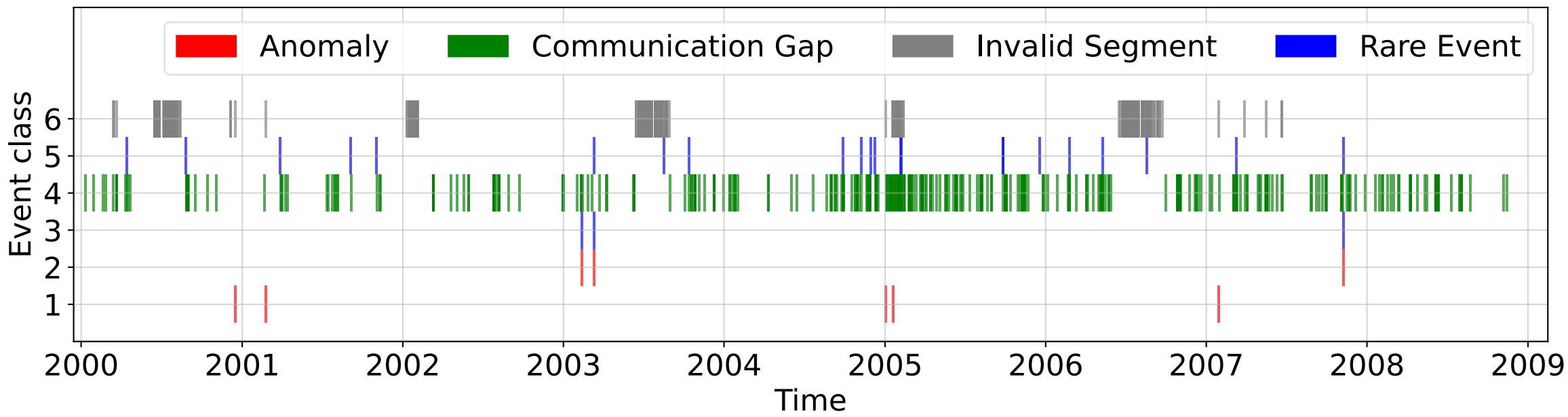

Supplementary Fig. 7. **Distributions of classes of annotated events across the timeline of Mission3**.

Supplementary Table 3. **Statistics of Mission3 data**.

| | **Mission3** |
|---|---:|
| **Channels** | **48** |
| Target / Non-target | 24 / 24 |
| Channel groups | 12 |
| Subsystems | 3 |
| **Telecommands** | **0** |
| **Data points** | **744,530,898** |
| Duration (anonymised) | 8 years |
| Compressed size [GB] | 3.47 |
| Annotated points [%] | 1.03 |
| **Annotated events** | **586** |
| Anomalies | 8 |
| Rare nominal events | 25 |
| Communication gaps | 397 |
| Invalid segments | 156 |
| Univariate / Multivariate | 8 / 25 |
| Global / Local | 28 / 5 |
| Point / Subsequence | 3 / 30 |
| **Distinct event classes** | **6** |

## 2.2. Examples of challenging events to detect

As mentioned in the Main text, the initial selection of missions was based on the presence of challenging anomalies according to SOEs. To support the analysis of results, a list of selected events of this type in test sets of ESA-ADB is provided in Supplementary Table 4. It is not a complete list. It is limited to test sets and includes only subjectively selected examples among many others. Example detections by semi-supervised algorithms trained on full (suffix "*-Full*") and lightweight (suffix "*-Light*") subsets for selected events are presented in Supplementary Material 4 as a series of figures referenced in Supplementary Table 4.

Other interesting examples include events from classes 2, 14, 15, and 22 in Mission1 where similar changes in the same channel are sometimes categorised as anomalies and sometimes as rare nominal events, depending on the presence of TCs. A similar case for Mission2 is visualised in Supplementary Fig. 8 for the non-commanded anomaly id_618 and the commanded rare event id_609. One of the important future works is to design algorithms that would be able to distinguish between such cases.

There are also some interesting nominal fragments related to atypical changes in sampling frequency in Mission1. There are 3 main examples of such behaviour in the training set on days 2001-05-28, 2001-05-31, and 2001-06-27 where rapidly changing sampling rate causes small atypical "gaps" in data for channels 41-46. In the refinement process, it was observed that those gaps are detected as anomalies by many algorithms. However, we decided that they should not be annotated, because varying sampling rates are expected in satellite telemetry and these false detections are mainly since the selected algorithms are not aware of frequency changes.

Supplementary Table 4. **List of selected challenging events annotated in test sets of ESA-ADB**.

| Mission | Event category | Event ID | Start time (YYYY-MM-DD hh:mm:ss) | Duration | Reason for selection |
|---|---|---|---|---|---|
| **Mission1** | Rare Event | id_24 | 2012-12-18 06:32:09 | 24h 15m | Hard to spot and not commanded. Not found by SOEs initially and added during the refinement process. It is related to a temporary change of nominal operational conditions. |
| | Rare Event | id_49 | 2011-10-08 07:08:39 | 10h 25m | Hard to spot, especially when looking at too narrow context. Caused by a rare TC. Supplementary Fig. 16. |
| | Rare Event | id_51 | 2011-08-14 19:12:39 | 1h 19m | Hard to spot, especially when looking at too narrow context. Caused by a rare TC. Supplementary Fig. 17. |
| | Rare Event | id_55 | 2011-04-23 08:19:39 | 0s (point) | Hard to spot in both lightweight and full sets. Caused by a unique execution of TC of priority 1. Overlaps with the rare event id_155. |
| | Anomaly | id_138 | 2009-10-13 06:39:17 | 1d 20h | Hard to spot using the lightweight subset of channels 41-46 only. Much easier to spot in channels 58-60. Supplementary Fig. 18 |
| | Anomaly | id_153 | 2011-01-28 22:29:18 | 15h 14m | Hard to spot using the lightweight subset of channels 41-46 only. Much easier to spot in channels 64-66. Supplementary Fig. 19 |
| | Rare Event | id_155 | 2011-04-21 22:15:52 | 11d | Hard to spot using the lightweight subset of channels 41-46 only. Easier to spot in multiple other channels, but hard to accurately identify the start time due to very slow changes. Supplementary Fig. 9 and Supplementary Fig. 20 |
| | Anomaly | id_157 | 2011-04-19 07:09:39 | 14h 35m | Hard to spot using the lightweight subset of channels 41-46 only. Much easier to spot in channels 64-66. |
| | Rare Event | id_159 | 2011-06-09 02:57:09 | 10d 21h | Hard to spot using the lightweight subset of channels 41-46 only. Very long annotations in other affected channels. Supplementary Fig. 21 |
| **Mission2** | Rare Event | id_466 | 2003-02-08 16:25:19 | 1h 10m | Small disturbance in 7 channels which may be easily overlooked, especially when using only the lightweight subset. |
| | Rare Event | id_591 | 2002-04-16 16:30:53 | 35m | Small disturbance in 7 channels which may be easily overlooked. |
| | Anomaly | id_631 | 2001-12-14 19:16:29 | 1h 18m | Small disturbance of unknown source in 7 channels which may be easily overlooked by operators. Supplementary Fig. 22 |
| | Anomaly | id_644 | 2002-02-18 05:42:45 | 9h 48m | Divergence of channel 81 from channel 73 which can only be detected when looking at both channels in the proper context window. |

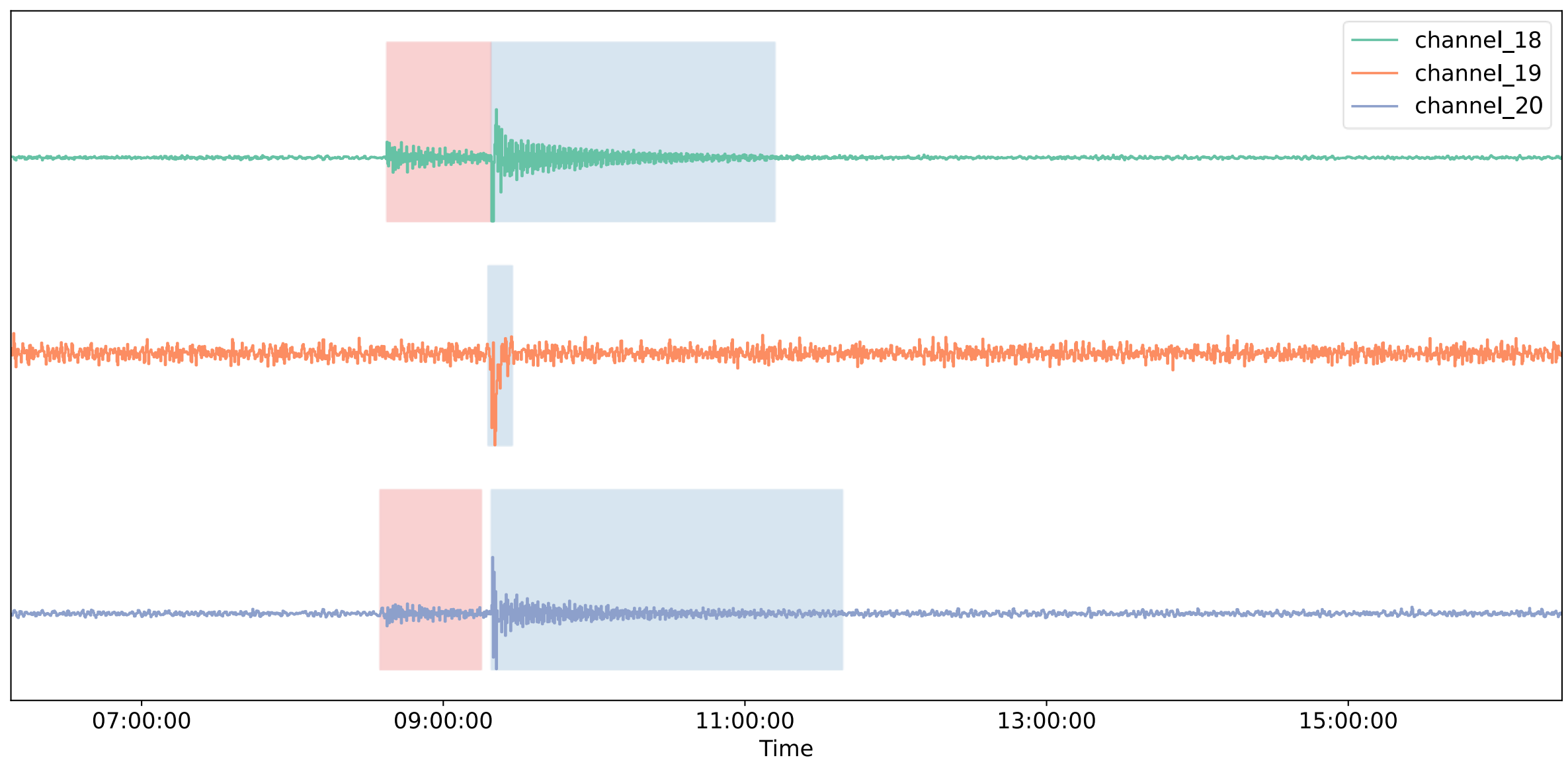

Supplementary Fig. 8. **Annotated anomaly id_618 (marked in red) directly preceding the commanded rare nominal event id_609 (marked in blue) in Mission2.** The Y axis is omitted because channels are normalised and shifted vertically for better visualisation.

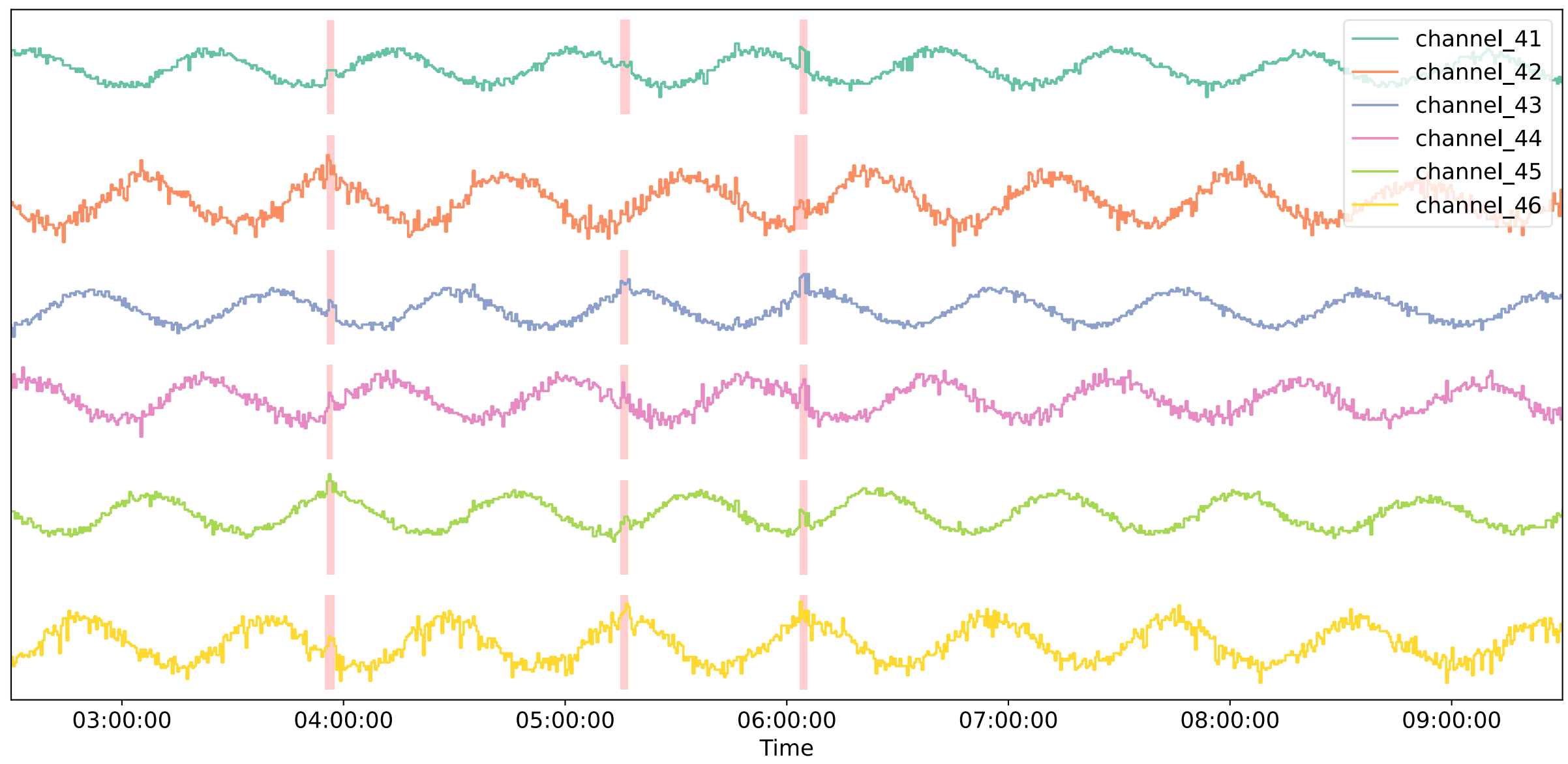

Supplementary Fig. 9. **Fragment of the annotated rare nominal event id_155 (marked in red) from Mission1**. The Y axis is omitted because channels are normalised and shifted vertically for better visualisation.

## 2.3. Anonymisation details

The anonymisation had to be applied to conform with the ESA privacy policy and to avoid any accidental disclosure of sensitive mission-specific information or metadata. The anonymisation process was carefully designed to maintain data integrity, so the results are independent of the anonymisation. The following modifications were applied as a part of the anonymisation process for each mission:

- Renaming of missions, subsystems, channels, telecommands, physical units, anomaly classes, and event types. They were consistently numbered according to their order of occurrence in files. Subsystems and physical units have consistent naming across missions, so it is possible to train cross-mission models.
- Time scaling and shifting of each mission. The timeline of every mission was scaled by a non-disclosed factor larger than 1 and shifted to start on 1st January 2000.
- Normalizing values within channel groups to <0, 1> range. Normalisation per group was applied to preserve the same dependencies between similar channels before and after anonymisation.

It was verified that the anonymisation is fully reversible and there are no numerical errors related to the limited floating point resolution of values or timestamps. Additionally, it was verified that all deterministic algorithms in the benchmark produce the same results before anonymisation.

## 2.4. Dataset structure

ESA-AD consists of three folders, one per each mission. Each folder has the same structure presented in Supplementary Table 5. There is a subfolder named *channels* and an optional subfolder named *telecommands*. Both subfolders include serialised and compressed *Pickle* files (docs.python.org/3/library/pickle.html, protocol version 4.0, *zip* compression), one for each channel and telecommand. Each file contains a single *pandas DataFrame* (pandas.pydata.org/pandas-docs/stable/reference/api/pandas.DataFrame.html, *pandas* version 1.5.3) including an index with consecutive timestamps and a single column with the corresponding raw telemetry values. Annotations of all events are in a separate file called *labels.csv* placed directly in the mission folder. It contains rows that describe anomalous fragments using 4 columns: the anomaly identifier (ID), the name of the channel affected by the anomaly, the start time, and the end time of the anomalous segment. Start and end times are defined as closed ranges and they usually represent timestamps of actual points in the dataset. There may be multiple segments with the same ID and channel name, but their time ranges cannot overlap. Additional information on anomalies can be found in the *anomaly_types.csv* file. It describes each anomaly ID with its class, subclass, category, and type. The channels are described in the *channels.csv* file using the channel name, the associated subsystem, and the physical unit. The channel description also includes group numbers that indicate similar channels and the information if the channel is a target channel. If telecommands are included in the dataset their priority is described in the *telecommands.csv* file.

Supplementary Table 5. **Folder structure of ESA-AD**

- `ESA-Mission/`
  - `channels/` folder including all channels of the mission
    - `*.zip` compressed *Pickle* files for each channel
  - `telecommands/` *(optional)* folder including all telecommands of the mission
    - `*.zip` compressed *Pickle* files for each telecommand
  - `labels.csv` annotations
  - `anomaly_types.csv` description of anomalies and rare nominal events
  - `channels.csv` description of channels
  - `events.csv` *(optional)* list of special operations and mission events
  - `telecommands.csv` *(optional)* description of telecommands

Some files are marked as optional, these files are not mandatory for the dataset or there might be missions not including these files. It should be possible to apply anomaly detection algorithms to the datasets not using the optional data, but it is expected that the optional data enhances the performance of algorithms when used. Mission2 includes an optional file *events.csv* which lists special operations and events with their start and end times according to the mission plan provided by SOEs. It was used to identify rare nominal events annotated in *labels.csv*, usually with slightly different start and end times due to different propagation times between channels.

## 2.5.Comparison to related public datasets

A quantitative comparison of the missions included in ESA-ADB and other public spacecraft-related telemetry datasets from the literature is presented in Supplementary Table 6. There are only 4 other public datasets of real-life spacecraft telemetry and a single simulated one. The most popular ones are Soil Moisture Active Passive (SMAP) and Mars Science Laboratory (MSL) released by NASA[2]. According to the search for "SMAP" and "MSL" terms in Google Scholar since 2018, there are more than 200 documents that mention these datasets in the set of more than 500 citations of the source paper[2]. Besides a lot of criticism of these datasets in the recent literature[17,18,69], there is a common misconception about the number of channels included in these datasets. The data may come from 82 different physical channels in total, but there is a separate fragment for each channel without any synchronization with other channels, so they cannot be used effectively as a multivariate dataset. This is made clear in the description of the dataset in Supplementary Table 6. NASA LASP WebTCAD[82] has tens of millions of points, but there are only 5 partially overlapping channels and no annotations of anomalies. Mars Express Power Challenge[83] is popular in satellite telemetry forecasting, but does not contain anomalies annotations.

There are also several related real-life datasets from outside the domain of satellite telemetry that are frequently used to benchmark multivariate TSAD algorithms. Notable examples include the Secure Water Treatment (SWaT)[84] and Water Distribution (WADI)[85] datasets which contain recordings from tens of channels from a real-world water treatment plant within several days. Server Machine Dataset (SMD)[26] including 5-week-long data from 38 parameters of 28 machines from 3 servers at a large internet company. The recent TELCO dataset[28] is worth noting due to related ideas of separate annotations for each channel, anomalies in training sets, and gradually increasing training set sizes. It contains 12 channels corresponding to real measurements collected over 7 months at an operation mobile internet service provider. To the best of our knowledge, the Exathlon benchmark[48], including real data traces from tens of repeated executions of streaming jobs with 2283 parameters (channels) on a Spark cluster over 2.5 months, is the only related dataset of volume comparable to ESA-AD, with more than 5 billion samples and 25 GB of data. However, it does not contain per-channel annotations and has been criticised for unrealistic anomaly density and positional bias[17].

Supplementary Table 6. **Quantitative comparison of ESA-AD and other related public datasets from the literature**.

| Dataset name | Number of channels | Total volume | Number of annotated events |
|---|---|---|---|
| ESA-AD | **Mission1**: 1 fragment with 76 channels and 698 commands<br>**Mission2**: 1 fragment with 100 channels and 123 commands | 1,551,591,259 samples<br>3,512,724 commands executions | 842 (1.17% of all samples) |
| NASA SMAP and MSL[2] | **SMAP**: 55 fragments with 1 channel and 24 commands<br>**MSL**: 27 fragments with 1 channel and 55 commands | 706,971 samples<br>410,030 commands executions | 105 (8.98% of all samples) |
| NASA LASP WebTCAD[82] | 1 fragment with 5 partially overlapping channels | 55,258,122 samples | not annotated |
| NASA Shuttle Valve Data (cs.fit.edu/~pkc/nasa/data) | **TEK**: 12 fragments with 1 channel<br>**VT1**: 27 fragments with 1 channel | 552’000 samples | 8 whole fragments |
| CATS[27] (simulated) | 1 fragment with 17 channels | 85,000,000 samples | 200 (2.15% of all samples) |
| Mars Express Power Challenge[83] | **Train**: 3 fragments with 38 channels (including 5 metadata-related)<br>**Test**: 1 testing fragment with 5 metadata channels | 198,045,083 samples | not annotated for anomalies |

# 3. Methods

## 3.1. Anomaly types

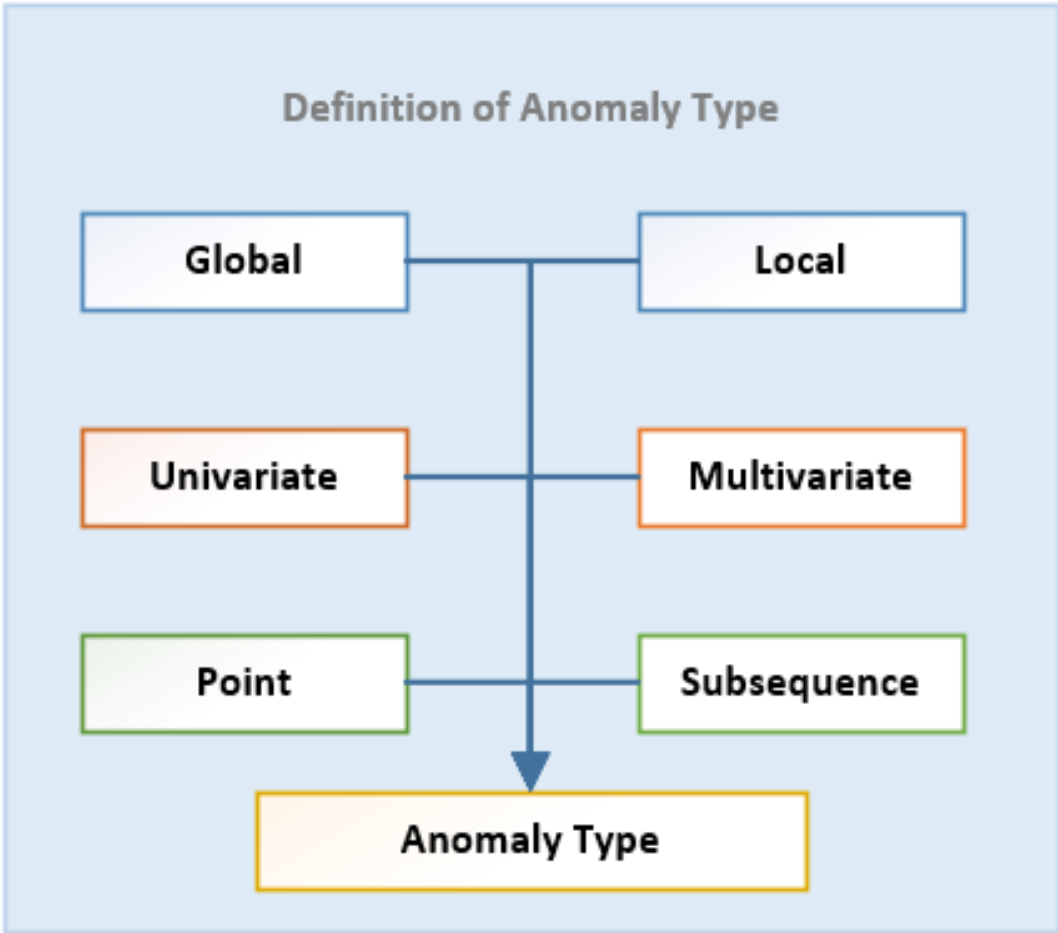

Supplementary Fig. 10. **Anomaly types considered in ESA-ADB**

Supplementary Table 7. **Distribution of 8 combinations of anomaly types across missions.**

| Length | Locality | Dimensionality | Mission1 | Mission2 | Mission3 |
|---|---|---|---|---|---|
| Point | Global | Univariate | 0.00% | 0.00% | 9.09% |
| | | Multivariate | 5.10% | 0.00% | 0.00% |
| | Local | Univariate | 0.51% | 0.00% | 0.00% |
| | | Multivariate | 0.51% | 0.00% | 0.00% |
| Subsequence | Global | Univariate | 12.24% | 0.00% | 60.61% |
| | | Multivariate | 40.31% | 90.84% | 15.15% |
| | Local | Univariate | 3.57% | 1.40% | 6.06% |
| | | Multivariate | 37.76% | 7.76% | 9.09% |

## 3.2. Metrics

### 3.2.1. Visualisations

Visualisations in this section are provided to build a better understanding of the proposed metrics. These examples are also included in unit tests of metrics in the published code.

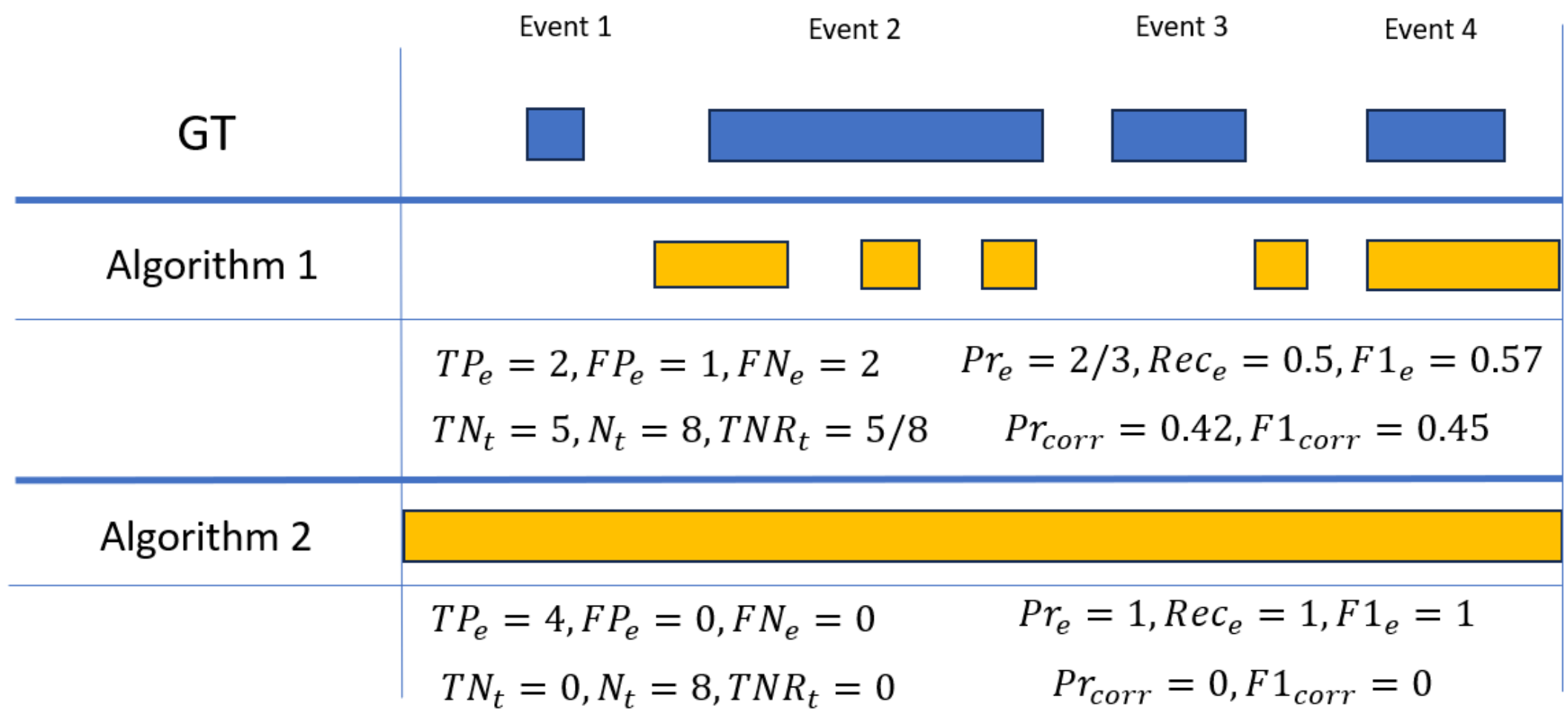

Supplementary Fig. 11. **Visualisation of differences between the original and corrected event-wise F-scores.**

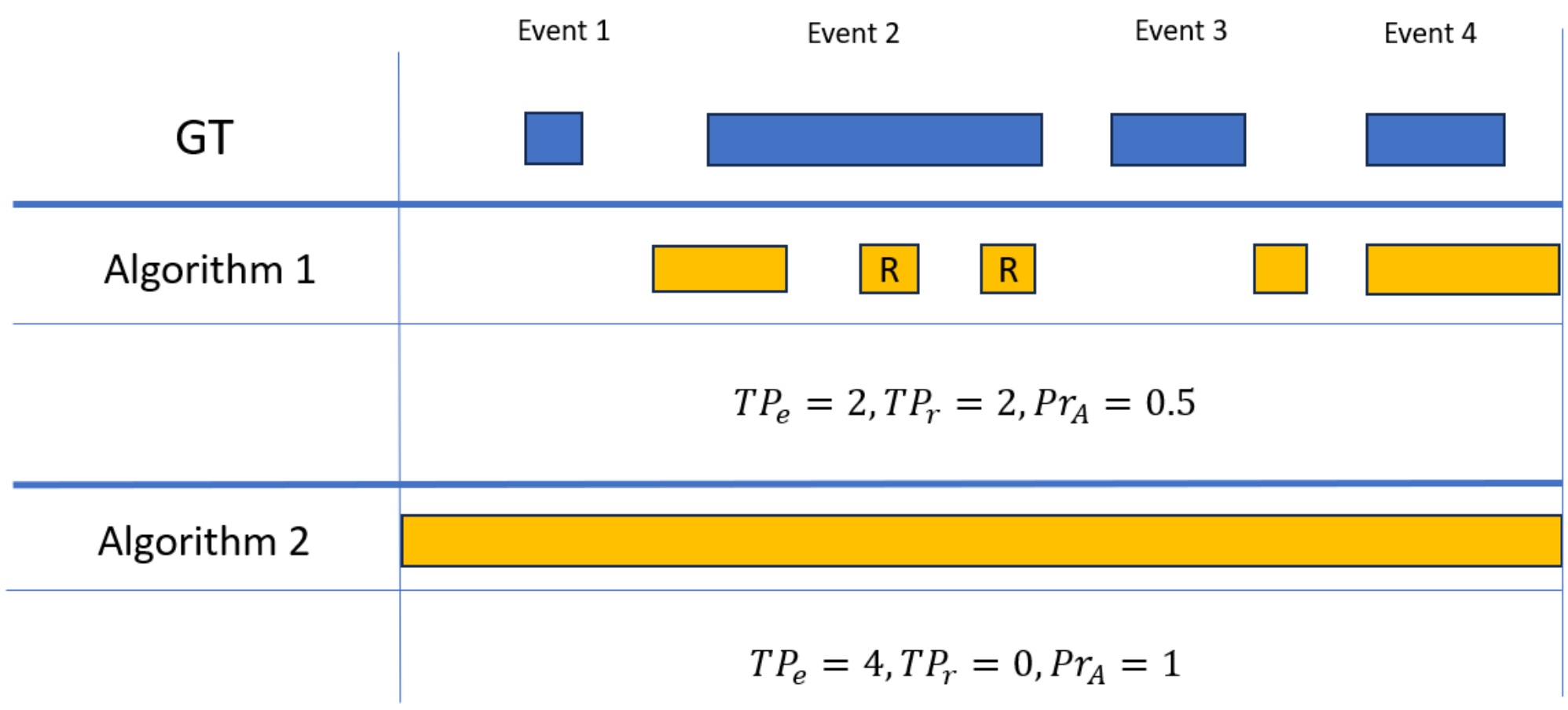

Supplementary Fig. 12. **Visualisation of the event-wise alarming precision calculation.**

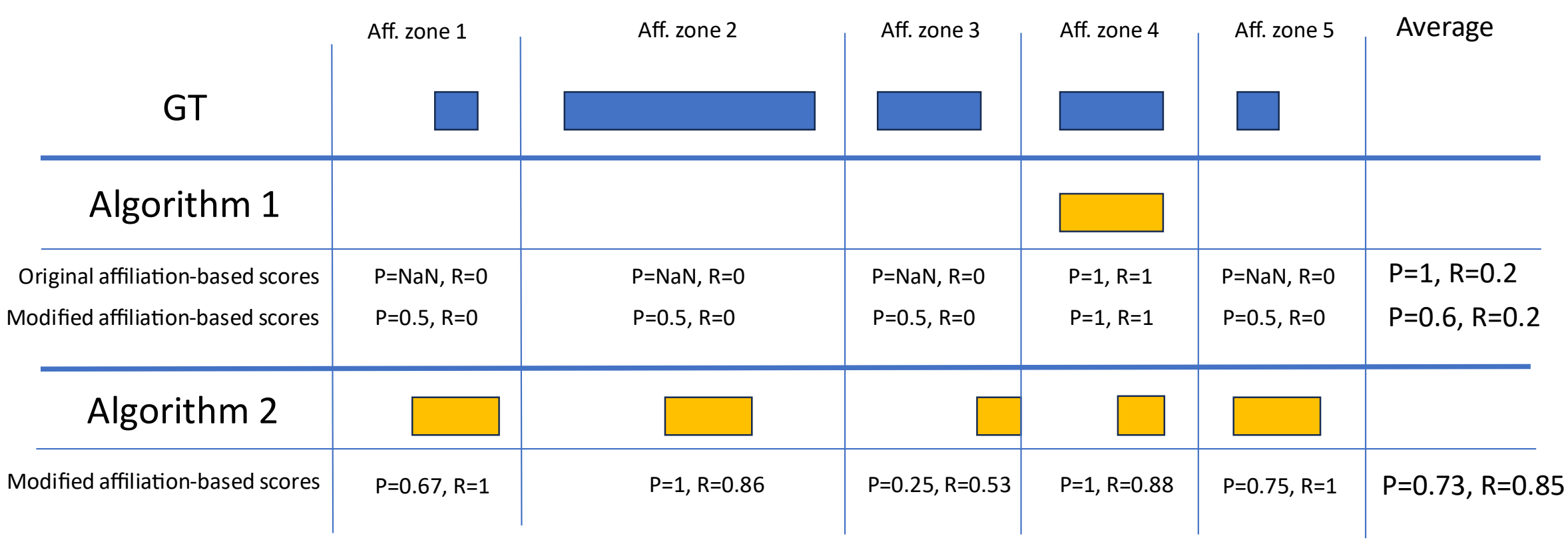

Supplementary Fig. 13. **Visualisation of differences between the original and modified affiliation-based scores on a simulated example**. (P – precision, R – Recall)

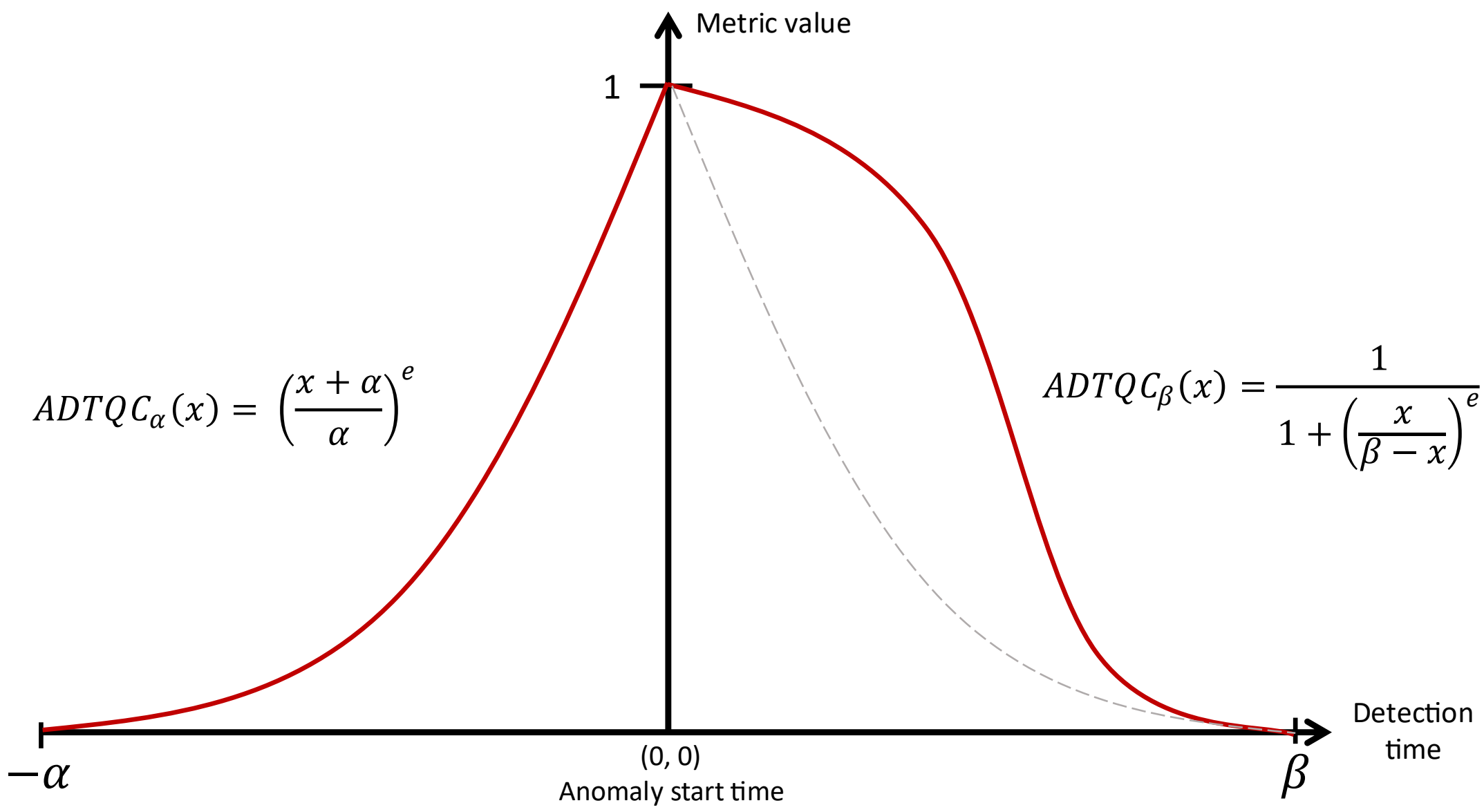

Supplementary Fig. 14. **Anomaly detection timing quality curve (ADTQC).**

### 3.2.2. Implementation details

ESA-AD has varying sampling rates and we keep them on purpose to maintain the true characteristics of satellite telemetry data. Our evaluation pipeline should handle this issue to consistently evaluate the results of algorithms using different sets of timestamps on the output. The only way to achieve this is to use metrics operating in the time domain instead of the samples domain, so that the ground truth and the detections can use completely different sets of timestamps (of different lengths and varying sampling rates). Original implementations of most metrics do not support timestamped arrays. They assume that the ground truth and the detections have the same uniformly sampled timeline. Our metrics operate on arbitrarily timestamped ground truth and detection arrays (possibly of different lengths and sampling frequencies). Hence, no matter the sampling frequency used in the algorithm, the metrics are always calculated relative to the original non-uniformly sampled ground truth. For operations on time ranges, we use the *portion* library (github.com/AlexandreDecan/portion).

Our modified version of the affiliation-based metric[22] operates on timestamped arrays, but the timestamps are transformed into the number of nanoseconds since the beginning of the dataset, so the internal implementation of the affiliation-based score is unchanged (it can operate on real numbers only). Additionally, point events (with the same start and end times) are adjusted, so that the end time is 1 nanosecond later than the start time. Such modification had to be applied because the affiliation-based score cannot be calculated for point events of zero length. The same point anomalies adjustment was applied to the channel-aware F-scores.

It is a common situation in our dataset that multiple non-overlapping fragments are annotated with the same anomaly ID. This is usually because the source of the anomaly is the same for all fragments. In such cases, we should treat all fragments as a single anomaly (i.e. when selecting affiliation zones and calculating distances) as suggested in recent literature[18,22]. To implement such correction in the affiliation-based score without changing its internal assumptions and implementation, a macro-averaging across anomaly IDs is proposed, i.e. to first aggregate zones affiliated with the same anomaly IDs by averaging their precision and recall scores and then calculate an average across all anomaly IDs.

There are several cases in our dataset where annotations for different events for the same channel are overlapping in time, i.e. when an anomaly occurred during a longer rare event. The affiliation-based metric is unable to separate such events because it is impossible to create non-overlapping affiliation zones for them, so there are no corrections for this situation to not interfere with the main principles and assumptions of the metric.

All metrics can be calculated excluding some specific event categories, classes, or types. For the corrected event-wise F-score, detections for excluded events are ignored when counting true and false positives, and a lack of detection is not counted as a false negative for them. For other metrics, excluded events are simply not considered when calculating the mean across events.

### 3.2.3. Metrics for rare nominal events

Most algorithms in the TimeEval framework (and in the literature) do not support learning rare nominal events explicitly (i.e. by one-shot learning or keeping rare events in memory). For such standard algorithms, rare events will always be detected as anomalies, so for simplicity, rare nominal events are treated as anomalies in the current benchmark. However, we strongly encourage to use ESA-AD to design models that learn nominal rare events and avoid detecting them in the future, which would be of high practical importance for mission control. For this purpose, we propose a framework to assess them:

1) The first detection of a novel rare nominal event (not seen during training) should not be penalised. However, the algorithm should be able to actively learn from the operators' feedback (i.e. "*this is not an anomaly*") and should not detect similar events in the future (one-shot learning).

2) For known rare events (seen during training or actively learned during inference), every subsequent detection should be penalised, i.e. we should minimise per-event false positive rate (FP / (FP + TN)) where FP is falsely detected rare event and TN is a correctly undetected rare event.

### 3.3. Preprocessing

Supplementary Fig. 15 presents an example of the proposed zero-order hold resampling scheme. The rightmost sample in the resampled Channel_1 is the effect of our correction for missing anomalies.

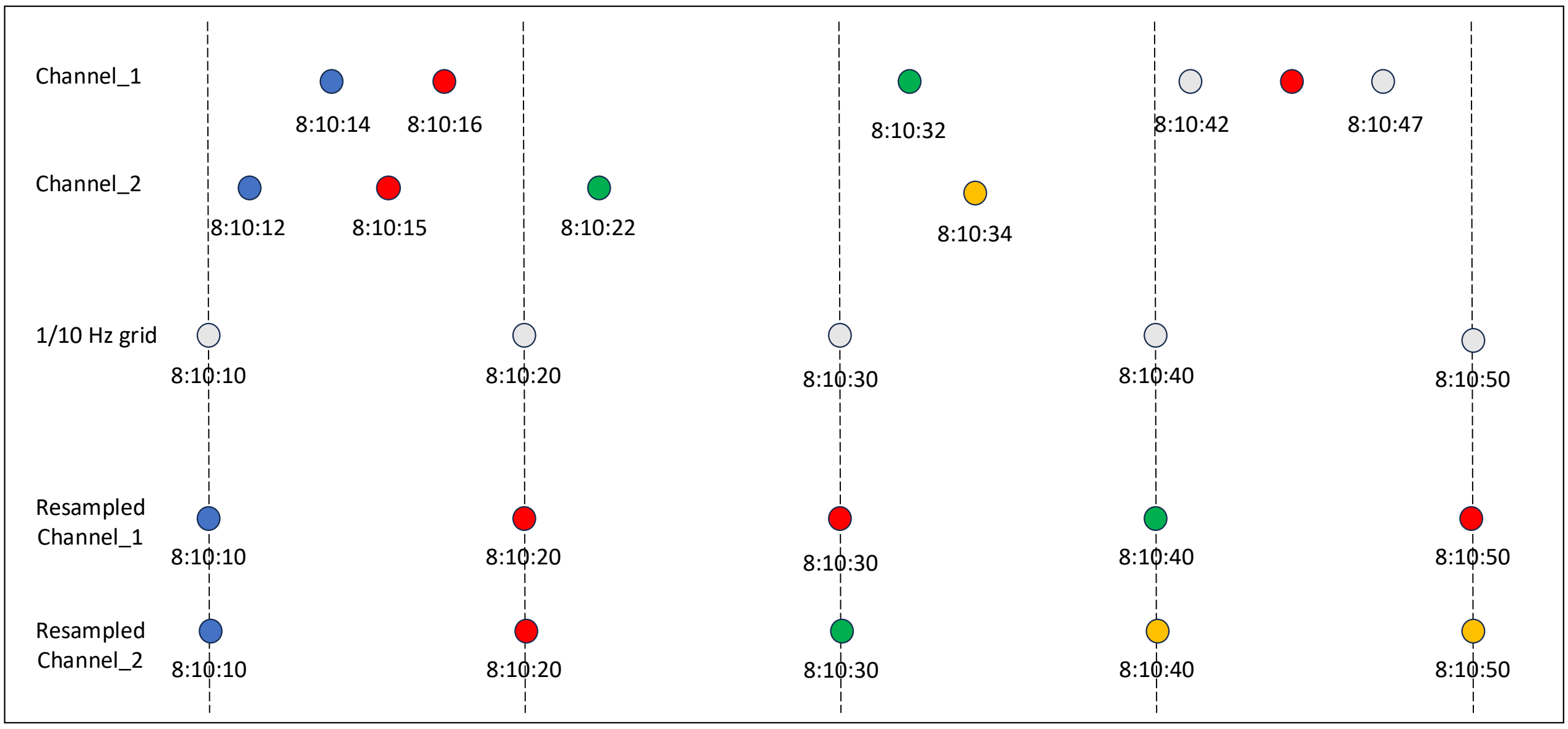

Supplementary Fig. 15. **Visualisation of our resampling procedure for two non-uniformly sampled channels**. Colours represent different values of the signal for each channel.

### 3.4. Algorithms' parametrisation

To support the full reproducibility of our results, Supplementary Table 8 lists all the algorithms' parameters and their values used in our experiments. The parameters' names directly correspond to the published code based on the TimeEval framework[29]. They use default values or settings recommended by algorithms' authors, sometimes adjusted to the specific features of our datasets (boldfaced in the table).

The number of 50 bins in HBOS was arbitrarily selected based on the analysis of the histograms of channels because the default value of 10 seemed to be much too small for our dataset. The default window size in Windowed iForest was decreased from 100 to 17 to avoid out-of-memory errors for our datasets. Many parameters of DC-VAE were adjusted to our dataset. The scaling is not used because it is already present in our preprocessing. Outliers are not rejected (*wo_outliers* is False) because our preprocessing code removes known anomalies. The window size is increased to 256 to be similar to the default Telemanom's window size (250). Also, the value of 256 showed good results on similar data in the original DC-VAE paper[28]. The number of CNN units is decreased from the default 64 to 32 because a significant overfitting was noticed in the validation scores for 64 units. The latent space dimensionality depends on the number of input channels in the same way as suggested for the TELCO dataset in the original DC-VAE code. The two main changes to Telemanom are 1) the increased number of units for full set training sets depending on the total number of input and output channels, and 2) the new min_error_value parameter to avoid magic numbers in the Telemanom code. The default value of the min_error_value is set to 0 (no magic numbers), but for Telemanom-ESA-Pruned it is arbitrarily selected to be 0.007 based on a manual analysis of reconstruction errors for the validation set, since the default value of 0.05 was much too high for some channels.

Importantly, the number of batches per epoch was limited to 1000 to avoid extremely long epoch training times for our datasets and to provide frequent validation score updates. Thus, the number of (sub)epochs was increased tenfold to 1000, and the early stopping patience was doubled to 20 for both DC-VAE and Telemanom to compensate for this.

Supplementary Table 8. **Parametrisation of algorithms used in ESA-ADB**. Boldfaced parameters and values are different from the default ones.

| Algorithm | Parameter name | Value(s) |
|---|---|---|
| PCC | max_iter | None |
| | n_components | None |
| | n_selected_components | None |
| | random_state | 42 |
| | svd_solver | auto |
| | tol | 0.0 |
| | whiten | False |
| HBOS | **n_bins** | **50** |
| | alpha | 0.1 |
| | bin_tol | 0.5 |
| | random_state | 42 |
| iForest | n_trees | 100 |
| | bootstrap | False |
| | max_features | 1.0 |
| | max_samples | None |
| | random_state | 42 |
| Windowed iForest | **n_trees** | **200** |
| | **window_size** | **17** |
| | bootstrap | False |
| | max_features | 1.0 |
| | max_samples | None |
| | random_state | 42 |
| KNN | distance_metric_order | 2 |
| | leaf_size | 30 |
| | method | Largest |
| | n_neighbors | 5 |
| GlobalSTD | tol | **3 (STD3) and 5 (STD5)** |
| | random_state | 42 |
| DC-VAE-ESA | **alpha** | **3 (STD3) and 5 (STD5)** |
| | **T (window size)** | **256** |
| | **cnn_units** | **32 (16 for Phase 1)** |
| | dil_rate | [1,2,4,8,16,32,64] |
| | kernel | 2 |
| | strs (stride length of CNN layers) | 1 |
| | batch_size | 64 |
| | **J (latent space dimensionality)** | **1/3 × total number of input channels and telecommands** |
| | **epochs** | **1000** |
| | lr (learning rate) | $10^{-3}$ |
| | seed | 123 |
| | early_stopping_delta | 0.001 |
| | **early_stopping_patience** | **20** |
| Telemanom-ESA | batch_size | 70 |
| | dropout | 0.3 |
| | early_stopping_delta | 0.0003 |
| | **early_stopping_patience** | **20** |
| | **epochs** | **1000** |
| | error_buffer | 100 |
| | layers | 2 |
| | **number of units per layer** | 80 for lightweight subsets. **Total number of input and output channels for full sets** |
| | lstm_batch_size | 64 |
| | **min_error_value (newly introduced to avoid magic numbers)** | **0 (0.007 for Telemanom-ESA-Pruned)** |
| | prediction_window_size | 10 |
| | random_state | 42 |
| | smoothing_perc | 0.05 |
| | smoothing_window_size | 30 |
| | window_size | 250 |

## 3.5. Useful scripts

There are a few additional useful scripts in the *scripts* folder in the published code:

- extract_fragments_for_OXI_annotator.py – extracts selected fragments of telemetry in the format compatible with the OXI annotation tool[45] (oxi.kplabs.pl).
- infer_anomaly_types.py –automatically assigns anomaly types based on our taxonomy.

- Mission1/2/3_timelines.ipynb – these scripts generate timelines of events as presented in Fig. 1 and Supplementary Fig. 7

- reevaluate.py – evaluates a trained model with different thresholding and different metrics, without the need for training the model again

# 4. Benchmarking results

## 4.1. Example detections

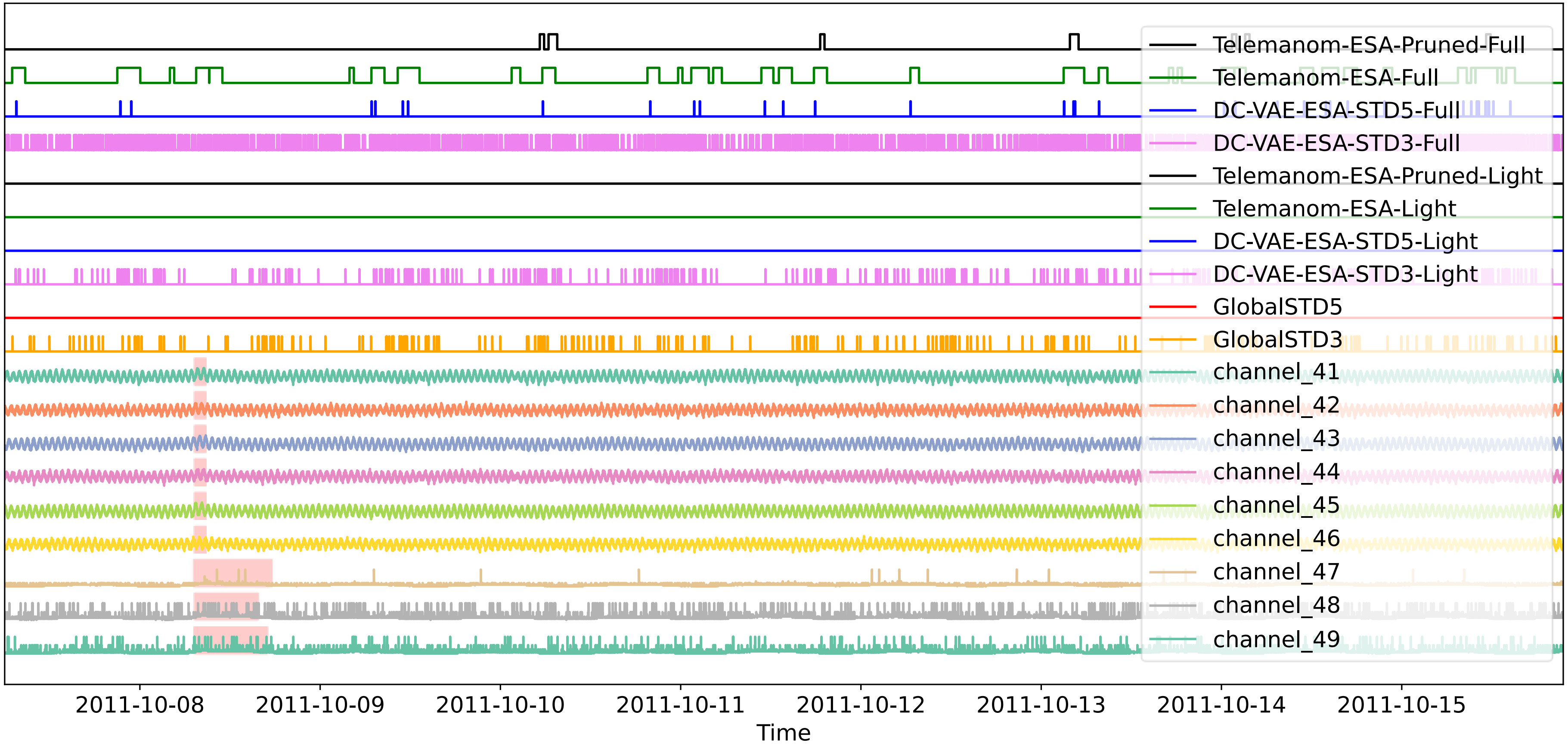

Supplementary Fig. 16. **Detections of rare nominal event id_49 (marked in red) for Mission1**. It is not detected when using only the lightweight subset of channels 41-46. For the full set, only Telemanom-ESA shows a reasonable detection, but it is surrounded by many false detections.

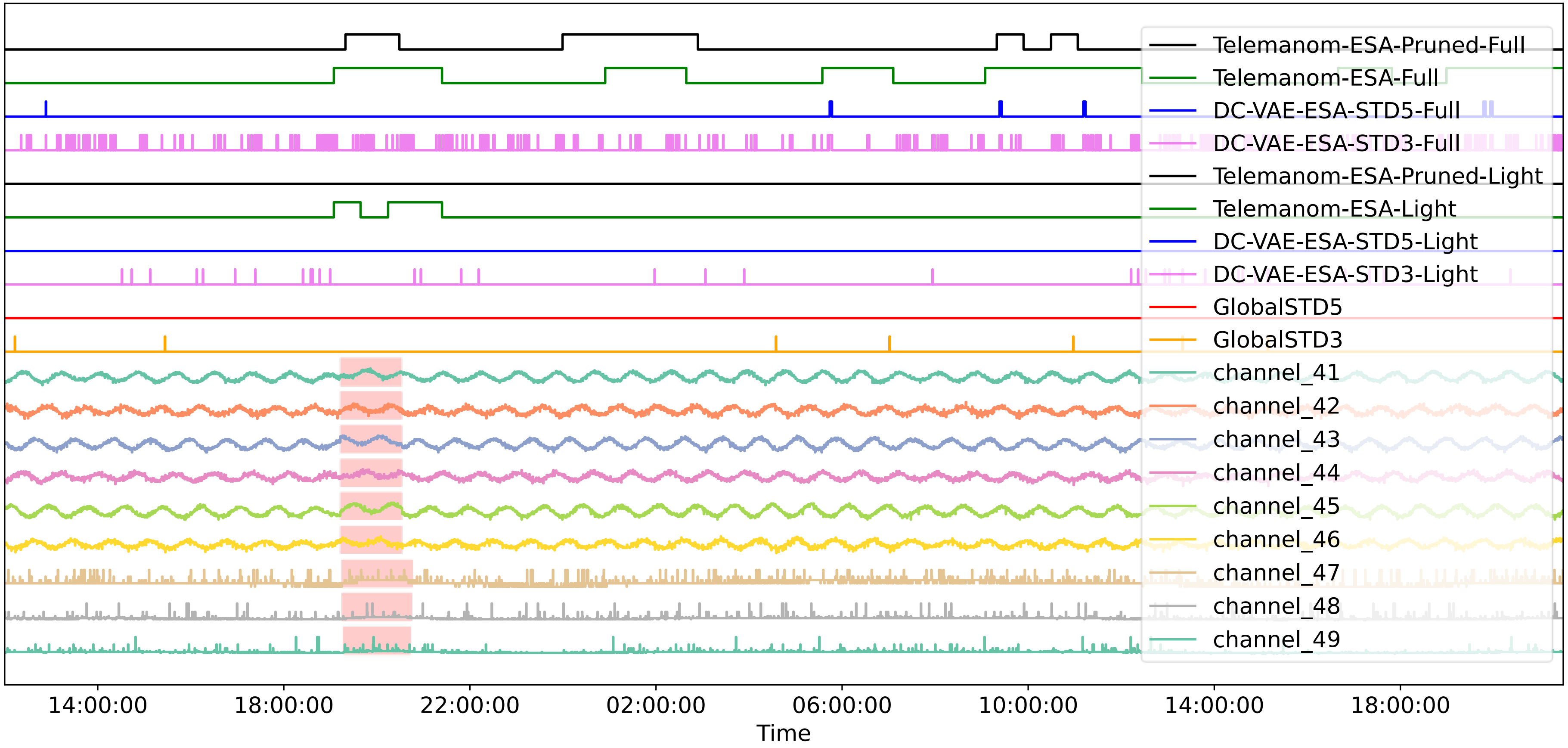

Supplementary Fig. 17. **Detections of rare nominal event id_51 (marked in red) for Mission1**. It is reasonably detected only by Telemanom-ESA. Surprisingly, Telemanom-ESA trained on the lightweight subset was also able to detect this.

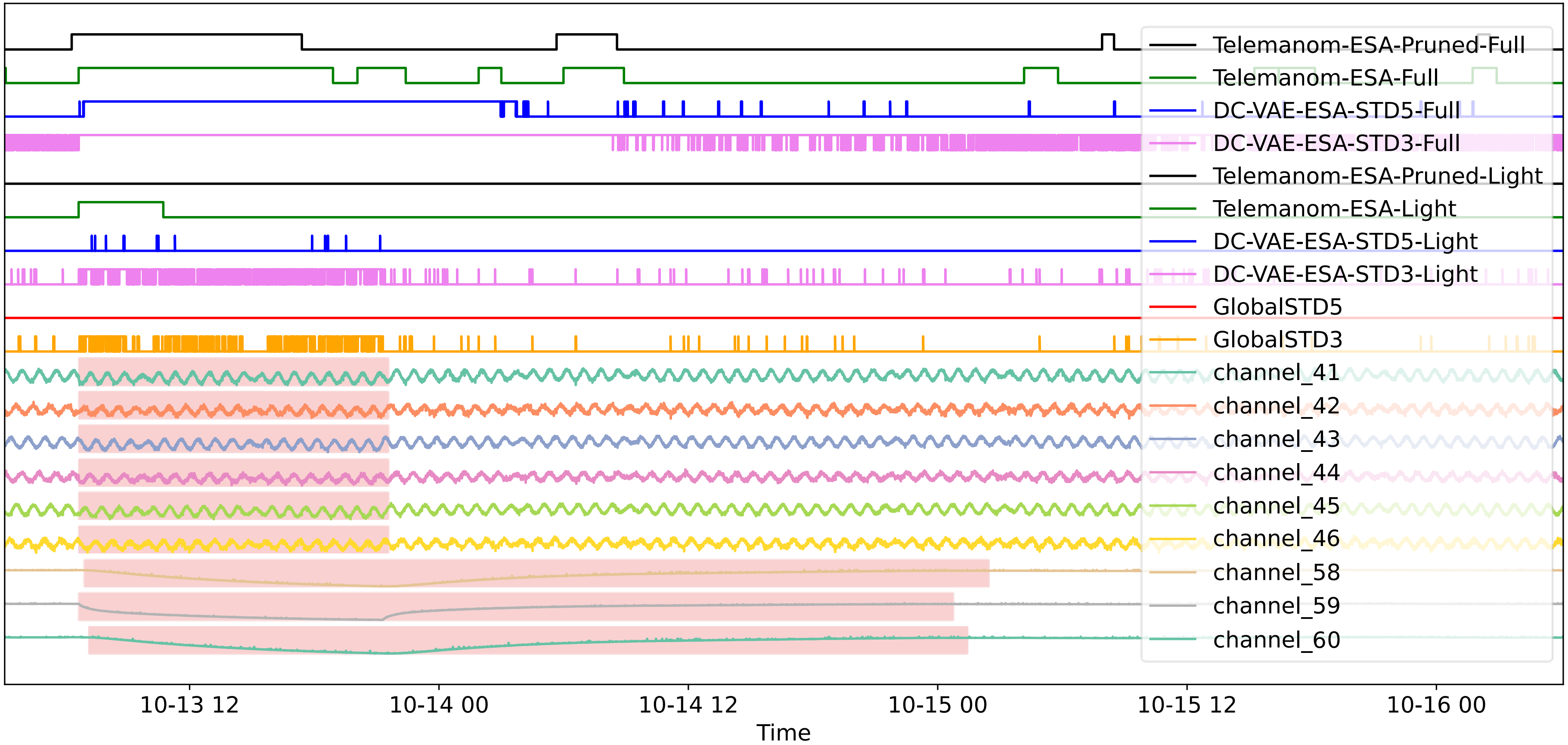

Supplementary Fig. 18. **Detections of anomaly id_138 (marked in red) for Mission1**. It is clearly visible in channels 58-60, so it is detected well by models trained on full sets of channels. However, it is not so easy using only the lightweight subset, i.e. Telemanom-ESA-Pruned-Light shows no response.

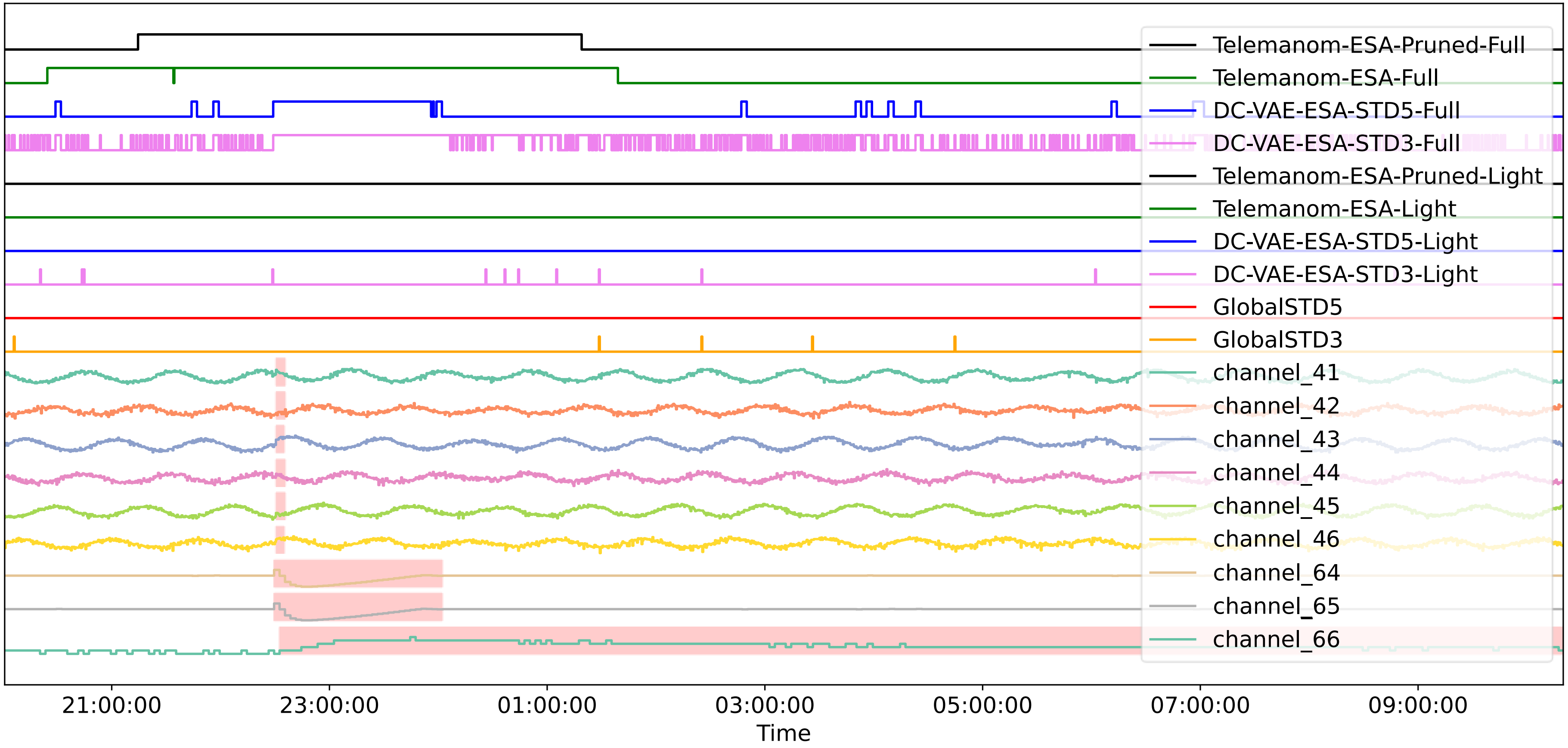

Supplementary Fig. 19. **Detections of anomaly id_153 (marked in red) for Mission1**. It is not detected when using only the lightweight subset of channels 41-46. For the full set, it is detected by all algorithms. Telemanom-ESA-Full detects it too early.

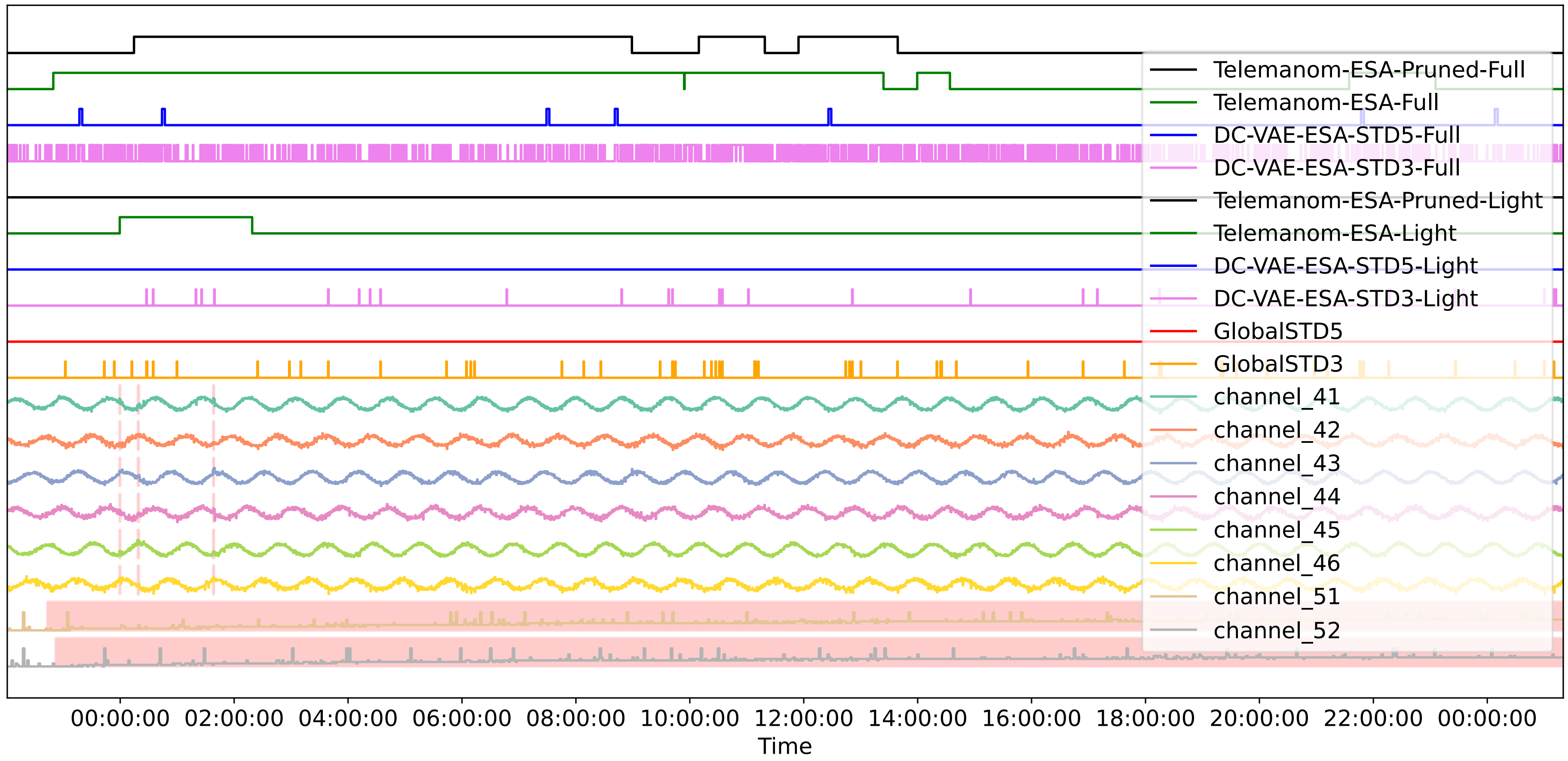

Supplementary Fig. 20. **Detections of rare nominal event id_155 (marked in red) for Mission1**. Only Telemanom-ESA was able to correctly detect this event in both lightweight and full sets of channels.

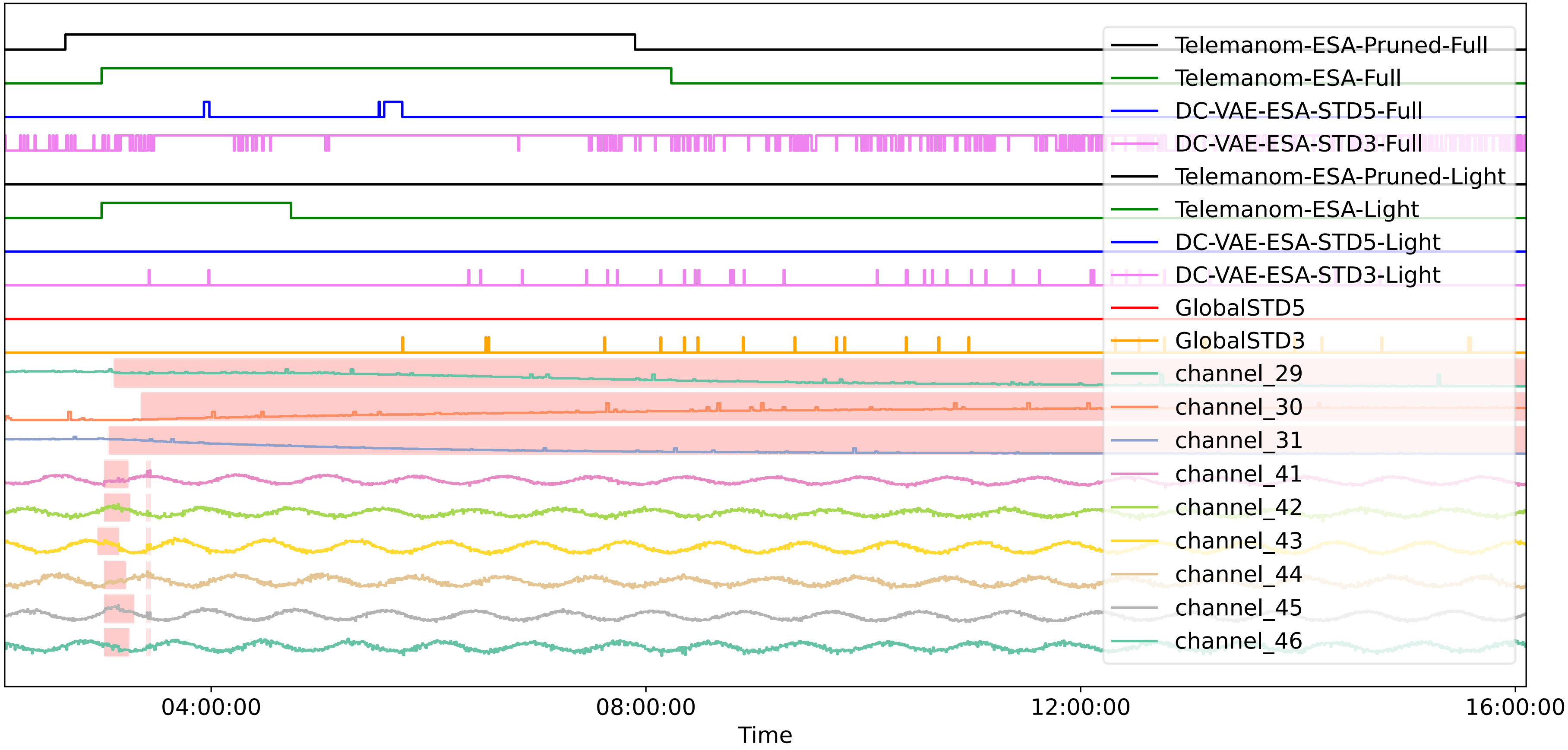

Supplementary Fig. 21. **Detection of rare nominal event id_159 (marked in red) for Mission1**. Only Telemanom-ESA was able to correctly detect this event in both lightweight and full sets of channels, with a good timing. DC-VAE-ESA-STD3-Full also seems to detected it relatively well.

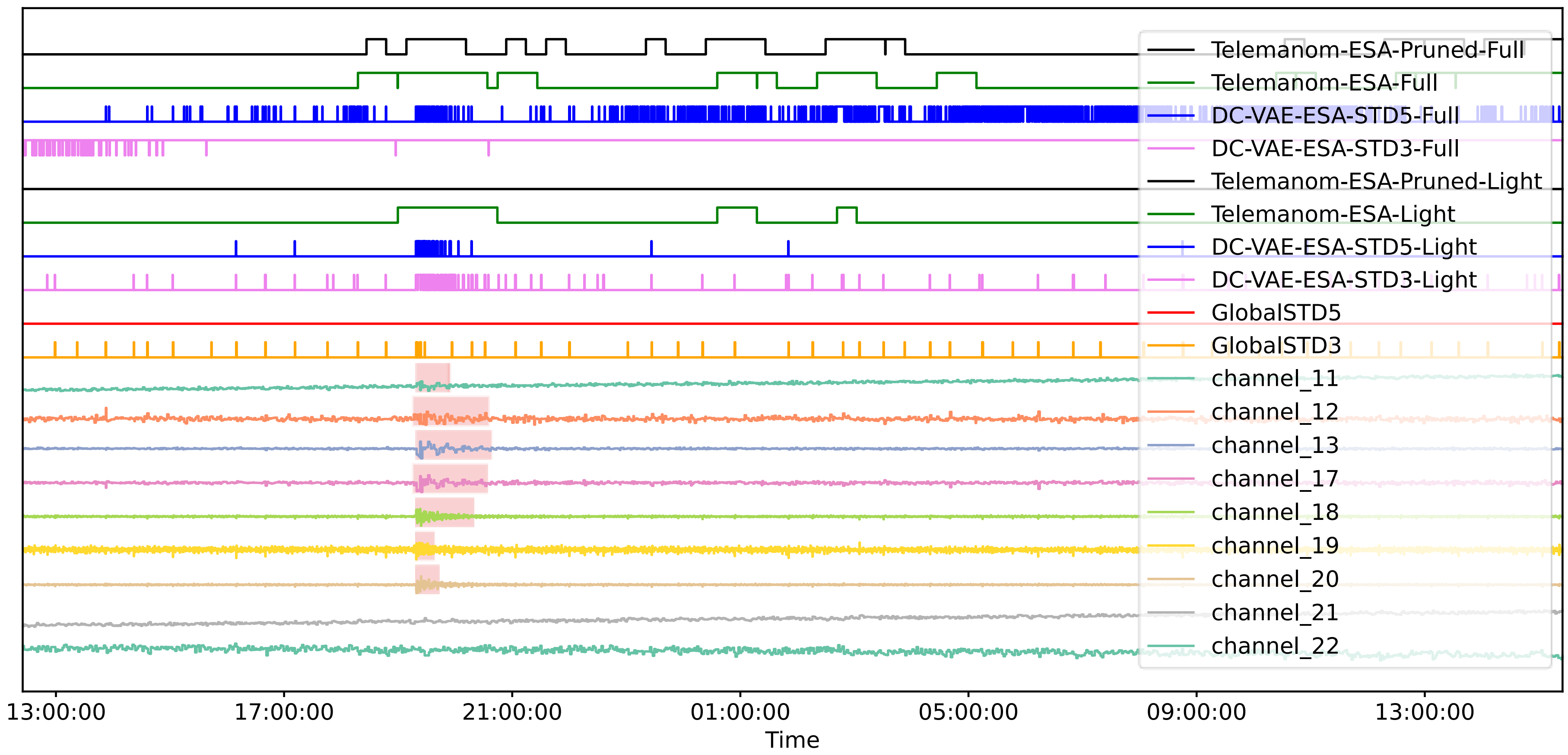

Supplementary Fig. 22. **Detection of anomaly id_631 (marked in red) for Mission2**. This anomaly is not so easy to spot manually but was detected by most algorithms, surprisingly, not by Telemanom-ESA-Pruned-Light.

## 4.2. Results for anomalies only

The analysis of the results for anomalies alone (excluding rare nominal events and communication gaps) in Supplementary Table 9 is important for understanding the performance of the algorithms in detecting the actual anomalies desired by SOEs. In this analysis, any true positives, false positives, or false negatives related to events different than anomalies are ignored (see implementation details in Supplementary Material 3.2.2). For Mission2, there are only 9 anomalies in the full test set and only 4 anomalies in the lightweight test set (see Supplementary Table 12), so the results should be interpreted with caution. A more reliable analysis can be conducted for Mission1 with 55 and 29 anomalies, respectively (see Supplementary Table 11).

Supplementary Table 9. **Benchmarking results for detection of anomalies alone in lightweight subsets of channels and all channels for missions in ESA-ADB**. Boldfaced results indicate the best values among all algorithms (excluding *After ratio* of ADTQC which is just a helper value).

**Mission1 – trained and tested on the lightweight subset of channels 41-46 – only anomalies**

| Metric | | PCC[32] | HBOS[33] | iForest[30] | Window iForest[30] | KNN[34] | Global STD3 | Global STD5 | DC-VAE-ESA STD3 | DC-VAE-ESA STD5 | Teleman-ESA | Teleman-ESA-Pruned |
|---|---|---|---|---|---|---|---|---|---|---|---|---|
| Event-wise | Precision | < 0.001 | < 0.001 | < 0.001 | < 0.001 | < 0.001 | < 0.001 | 0.205 | 0.001 | 0.021 | 0.074 | **0.999** |
| | Recall | 0.310 | 0.379 | 0.414 | 0.552 | 0.448 | 0.310 | 0.241 | 0.310 | 0.241 | **0.931** | 0.862 |
| | F0.5 | < 0.001 | < 0.001 | < 0.001 | < 0.001 | < 0.001 | < 0.001 | 0.211 | 0.001 | 0.026 | 0.090 | **0.968** |
| Channel-aware | Precision | Not available | | | | | 0.310 | 0.241 | 0.302 | 0.241 | **0.931** | 0.529 |
| | Recall | | | | | | 0.282 | 0.241 | 0.285 | 0.241 | **0.882** | 0.862 |
| | F0.5 | | | | | | 0.293 | 0.241 | 0.297 | 0.241 | **0.914** | 0.722 |
| Alarming precision | | 0.102 | 0.054 | 0.444 | 0.889 | 0.120 | 0.034 | 0.024 | 0.048 | 0.010 | 0.818 | **0.862** |
| ADTQC | After ratio | 0.889 | 0.636 | 0.500 | 0.063 | 0.385 | 0.889 | 1.000 | 1.000 | 1.000 | 0.037 | 0.040 |
| | Score | 0.826 | 0.676 | 0.730 | 0.308 | 0.670 | 0.826 | 0.919 | 0.911 | **0.921** | 0.220 | 0.159 |
| Affiliation-based | Precision | 0.536 | 0.543 | 0.532 | 0.562 | 0.521 | 0.561 | 0.919 | 0.559 | 0.906 | 0.774 | **0.927** |
| | Recall | 0.276 | 0.352 | 0.294 | 0.366 | 0.271 | 0.335 | 0.854 | 0.279 | 0.850 | 0.673 | **0.859** |
| | F0.5 | 0.451 | 0.490 | 0.458 | 0.508 | 0.440 | 0.494 | 0.906 | 0.466 | 0.894 | 0.752 | **0.912** |

**Mission1 – trained and tested on the full set of channels – only anomalies**

| Metric | | PCC[32] | HBOS[33] | iForest[30] | Window iForest[30] | KNN[34] | Global STD3 | Global STD5 | DC-VAE-ESA STD3 | DC-VAE-ESA STD5 | Teleman-ESA | Teleman-ESA-Pruned |
|---|---|---|---|---|---|---|---|---|---|---|---|---|
| Event-wise | Precision | < 0.001 | < 0.001 | < 0.001 | Out-of-memory | Out-of-memory | < 0.001 | 0.001 | < 0.001 | 0.003 | 0.004 | **0.032** |
| | Recall | 0.891 | 0.964 | 0.945 | | | 0.873 | 0.818 | 0.891 | 0.818 | **0.945** | 0.909 |
| | F0.5 | < 0.001 | < 0.001 | < 0.001 | | | < 0.001 | 0.002 | < 0.001 | 0.004 | 0.005 | **0.039** |
| Subsystem-aware | Precision | Not available | | | | | 0.491 | **0.782** | 0.424 | 0.648 | 0.712 | 0.355 |
| | Recall | | | | | | 0.721 | 0.676 | 0.739 | 0.721 | 0.855 | **0.909** |
| | F0.5 | | | | | | 0.507 | **0.748** | 0.448 | 0.644 | 0.717 | 0.397 |
| Channel-aware | Precision | Not available | | | | | 0.327 | 0.355 | 0.272 | 0.311 | **0.497** | 0.195 |
| | Recall | | | | | | 0.332 | 0.298 | 0.398 | 0.324 | **0.561** | 0.705 |
| | F0.5 | | | | | | 0.309 | 0.315 | 0.272 | 0.291 | **0.472** | 0.217 |
| Alarming precision | | 0.005 | 0.008 | 0.005 | | | 0.009 | 0.088 | 0.003 | **0.020** | 0.132 | 0.278 |
| ADTQC | After ratio | 0.633 | 0.415 | 0.423 | | | 0.708 | 0.756 | 0.673 | 0.733 | 0.327 | 0.380 |
| | Score | **0.611** | 0.553 | 0.633 | | | 0.728 | 0.654 | 0.730 | 0.654 | 0.561 | 0.536 |
| Affiliation-based | Precision | 0.527 | 0.512 | 0.501 | | | 0.521 | 0.531 | 0.512 | **0.531** | 0.512 | 0.611 |
| | Recall | 0.486 | 0.563 | 0.445 | | | 0.462 | 0.434 | 0.452 | 0.473 | **0.344** | 0.436 |
| | F0.5 | 0.519 | 0.521 | 0.489 | | | 0.508 | 0.508 | 0.499 | 0.518 | **0.467** | 0.566 |

**Mission2 – trained and tested on the lightweight subset of channels 18-28 – only anomalies**

| Metric | | PCC[32] | HBOS[33] | iForest[30] | Window iForest[30] | KNN[34] | Global STD3 | Global STD5 | DC-VAE-ESA STD3 | DC-VAE-ESA STD5 | Teleman-ESA | Teleman-ESA-Pruned |
|---|---|---|---|---|---|---|---|---|---|---|---|---|
| Event-wise | Precision | < 0.001 | 0.000 | **0.004** | 0.000 | < 0.001 | < 0.001 | < 0.001 | < 0.001 | < 0.001 | 0.001 | 0.000 |
| | Recall | **1.000** | 0.000 | **1.000** | 0.000 | **1.000** | **1.000** | **1.000** | **1.000** | **1.000** | **1.000** | 0.000 |
| | F0.5 | < 0.001 | 0.000 | **0.005** | 0.000 | < 0.001 | < 0.001 | < 0.001 | < 0.001 | < 0.001 | 0.001 | 0.000 |
| Channel-aware | Precision | Not available | | | | | **1.000** | **1.000** | **1.000** | **1.000** | 0.600 | 0.000 |
| | Recall | | | | | | 0.667 | 0.667 | **1.000** | 0.667 | **1.000** | 0.000 |
| | F0.5 | | | | | | 0.909 | 0.909 | **1.000** | 0.909 | 0.652 | 0.000 |
| Alarming precision | | 0.032 | 0.000 | 0.143 | 0.000 | 0.029 | 0.026 | 0.036 | 0.027 | 0.037 | **1.000** | 0.000 |
| ADTQC | After ratio | 1.000 | - | 1.000 | - | 1.000 | 1.000 | 1.000 | 1.000 | 1.000 | 0.000 | - |
| | Score | **1.000** | - | **1.000** | - | **1.000** | **1.000** | **1.000** | **1.000** | **1.000** | 0.358 | - |
| Affiliation-based | Precision | 0.845 | 0.500 | **1.000** | 0.500 | 0.705 | 0.826 | 0.894 | 0.816 | 0.950 | 0.781 | 0.500 |
| | Recall | 0.925 | 0.000 | 0.971 | 0.000 | 0.517 | 0.862 | 0.994 | 0.888 | 0.981 | **1.000** | 0.000 |
| | F0.5 | 0.860 | 0.000 | **0.994** | 0.000 | 0.657 | 0.833 | 0.912 | 0.830 | 0.956 | 0.817 | 0.000 |

**Mission2 – trained and tested on the full set of channels – only anomalies**

| Metric | | PCC[32] | HBOS[33] | iForest[30] | Window iForest[30] | KNN[34] | Global STD3 | Global STD5 | DC-VAE-ESA STD3 | DC-VAE-ESA STD5 | Teleman-ESA | Teleman-ESA-Pruned |
|---|---|---|---|---|---|---|---|---|---|---|---|---|
| Event-wise | Precision | **0.001** | < 0.001 | < 0.001 | < 0.001 | Out-of-memory | < 0.001 | **0.001** | < 0.001 | < 0.001 | **0.001** | **0.001** |
| | Recall | 0.667 | 0.667 | 0.667 | 0.500 | | 0.667 | 0.167 | 0.833 | 0.667 | **1.000** | **1.000** |
| | F0.5 | **0.001** | < 0.001 | < 0.001 | < 0.001 | | < 0.001 | **0.001** | < 0.001 | < 0.001 | **0.001** | **0.001** |
| Subsystem-aware | Precision | Not available | | | | | 0.167 | 0.000 | 0.333 | 0.333 | **0.417** | 0.278 |
| | Recall | | | | | | 0.167 | 0.000 | 0.500 | 0.333 | 0.833 | **1.000** |
| | F0.5 | | | | | | 0.167 | 0.000 | 0.352 | 0.333 | **0.452** | 0.324 |
| Channel-aware | Precision | Not available | | | | | 0.083 | 0.000 | 0.095 | 0.111 | **0.296** | 0.082 |
| | Recall | | | | | | 0.021 | 0.000 | 0.229 | 0.042 | 0.573 | **0.833** |
| | F0.5 | | | | | | 0.052 | 0.000 | 0.098 | 0.083 | **0.325** | 0.096 |
| Alarming precision | | 0.364 | 0.308 | 0.143 | 0.158 | | 0.031 | **1.000** | 0.026 | 0.040 | 0.375 | 0.462 |
| ADTQC | After ratio | 0.750 | 0.500 | 0.500 | 0.333 | | **1.000** | **1.000** | 0.600 | 0.750 | 0.500 | 0.500 |
| | Score | 0.542 | 0.489 | 0.493 | 0.437 | | **0.992** | 0.612 | 0.698 | 0.709 | 0.660 | 0.766 |
| Affiliation-based | Precision | 0.660 | 0.608 | 0.618 | 0.616 | | 0.523 | 0.500 | 0.539 | 0.418 | **0.671** | 0.620 |
| | Recall | 0.380 | 0.333 | 0.358 | 0.355 | | 0.318 | 0.000 | 0.522 | 0.398 | **0.709** | 0.604 |
| | F0.5 | 0.575 | 0.522 | 0.539 | 0.537 | | 0.466 | 0.000 | 0.536 | 0.414 | **0.678** | 0.617 |

### 4.3. Results for lightweight test sets using algorithms trained on full sets

For algorithms that provide separate anomaly scores for each channel, it is possible to limit the analysis of the global scores to an arbitrary subset of the channels used in training. It is especially useful to directly compare the results between models trained on full sets of channels and models trained only on lightweight subsets. Such a comparison is presented in Supplementary Table 10 for the DC-VAE-ESA and Telemanom-ESA algorithms. GlobalSTD is omitted because its results do not depend on the number of channels in training.

Supplementary Table 10. **Benchmarking results for detection of all events in lightweight test sets in ESA-ADB by algorithms trained on lightweight and full sets of channels**. Boldfaced results indicate the better value for each pair of training sets for each algorithm (excluding *After ratio* of ADTQC which is just a helper value).

| **Mission1 – tested on the lightweight test set** | | | | | | | | | |
|---|---|---|---|---|---|---|---|---|---|
| Algorithm → | | DC-VAE-ESA STD3 | | DC-VAE-ESA STD5 | | Teleman-ESA | | Teleman-ESA-Pruned | |
| Trained on → | | Light | Full | Light | Full | Light | Full | Light | Full |
| Event-wise | Precision | 0.001 | **0.008** | 0.014 | **0.216** | **0.148** | 0.027 | **0.999** | 0.043 |
| | Recall | **0.576** | 0.167 | **0.318** | 0.076 | **0.894** | 0.439 | 0.424 | 0.848 |
| | F0.5 | 0.001 | **0.009** | 0.017 | **0.158** | **0.178** | 0.033 | **0.786** | 0.054 |
| Channel-aware | Precision | **0.568** | 0.167 | **0.318** | 0.076 | **0.894** | 0.439 | 0.424 | **0.833** |
| | Recall | **0.442** | 0.101 | **0.207** | 0.066 | **0.738** | 0.328 | 0.275 | **0.848** |
| | F0.5 | **0.506** | 0.134 | **0.262** | 0.071 | **0.837** | 0.377 | 0.362 | **0.834** |
| Alarming precision | | 0.052 | **0.072** | 0.034 | **0.119** | **0.868** | 0.659 | **0.875** | 0.505 |
| ADTQC | After ratio | 0.921 | 0.909 | **0.952** | 0.800 | 0.136 | 0.586 | 0.143 | 0.286 |
| | Score | **0.805** | 0.607 | **0.799** | 0.728 | 0.428 | **0.625** | 0.197 | **0.431** |
| Affiliation-based | Precision | **0.577** | 0.562 | **0.741** | 0.524 | **0.727** | 0.616 | **0.711** | 0.621 |
| | Recall | **0.373** | 0.238 | **0.555** | 0.071 | **0.662** | 0.400 | 0.423 | **0.512** |
| | F0.5 | **0.520** | 0.441 | **0.694** | 0.231 | **0.713** | 0.556 | **0.626** | 0.596 |
| **Mission2 – tested on the lightweight test set** | | | | | | | | | |
| Algorithm → | | DC-VAE-ESA STD3 | | DC-VAE-ESA STD5 | | Teleman-ESA | | Teleman-ESA-Pruned | |
| Trained on → | | Light | Full | Light | Full | Light | Full | Light | Full |
| Event-wise | Precision | **0.003** | **0.003** | **0.064** | 0.017 | **0.188** | 0.152 | **0.978** | 0.268 |
| | Recall | **1.000** | **1.000** | **1.000** | **1.000** | 0.986 | **0.989** | 0.540 | **0.911** |
| | F0.5 | 0.003 | **0.004** | **0.079** | 0.021 | **0.224** | 0.183 | **0.842** | 0.312 |
| Channel-aware | Precision | 0.904 | **0.912** | **0.995** | 0.985 | 0.831 | **0.875** | 0.465 | **0.690** |
| | Recall | **0.554** | 0.543 | **0.451** | 0.445 | **0.870** | 0.739 | 0.384 | **0.848** |
| | F0.5 | 0.787 | **0.788** | **0.783** | 0.772 | 0.822 | **0.823** | 0.442 | **0.708** |
| Alarming precision | | 0.052 | **0.034** | **0.068** | 0.046 | **0.912** | 0.907 | **0.862** | 0.861 |
| ADTQC | After ratio | 0.908 | 0.848 | 0.991 | 0.966 | 0.087 | 0.105 | 0.351 | 0.350 |
| | Score | **0.996** | 0.934 | **0.997** | 0.989 | 0.507 | **0.508** | **0.757** | 0.676 |
| Affiliation-based | Precision | **0.680** | 0.675 | **0.939** | 0.914 | **0.688** | 0.681 | **0.759** | 0.738 |
| | Recall | 0.293 | **0.345** | **0.788** | 0.782 | **0.544** | 0.503 | 0.530 | **0.623** |
| | F0.5 | 0.538 | **0.566** | **0.904** | 0.884 | **0.654** | 0.636 | 0.699 | **0.712** |

## 4.4. Results for different mission phases

It is a common practice to periodically retrain or adapt algorithms when new telemetry becomes available from satellites, especially in the presence of significant changes in operational conditions. The experiments in this section simulate such an approach in ESA-ADB to assess the robustness of algorithms to changing conditions and to identify the earliest mission phase in which reliable detectors can be trained. These aspects are crucial for the selection of algorithms in different mission phases. Some classic algorithms may perform much better than others in early mission phases when very limited data is available, but they may be overcome by deep learning techniques in late mission phases. The goal of this section is to provide a basic analysis of these aspects in ESA-ADB. For this purpose, the effect of training set size (representing different mission phases) on the corrected event-wise F0.5-score for the test set is analysed for the lightweight subsets of each mission in Supplementary Table 13. The analysis for full sets is not conducted as the scores are very low even for the longest training set. There are 5 training set lengths (phases) proposed for Mission1 and 4 for Mission2 following the idea presented in Supplementary Fig. 23. Starting from just a few percent of the mission timeline (initial phases) to 50% of the mission (the default setting in ESA-ADB). The statistics of the phases are listed in Supplementary Table 11 (Mission1) and Supplementary Table 12 (Mission2).

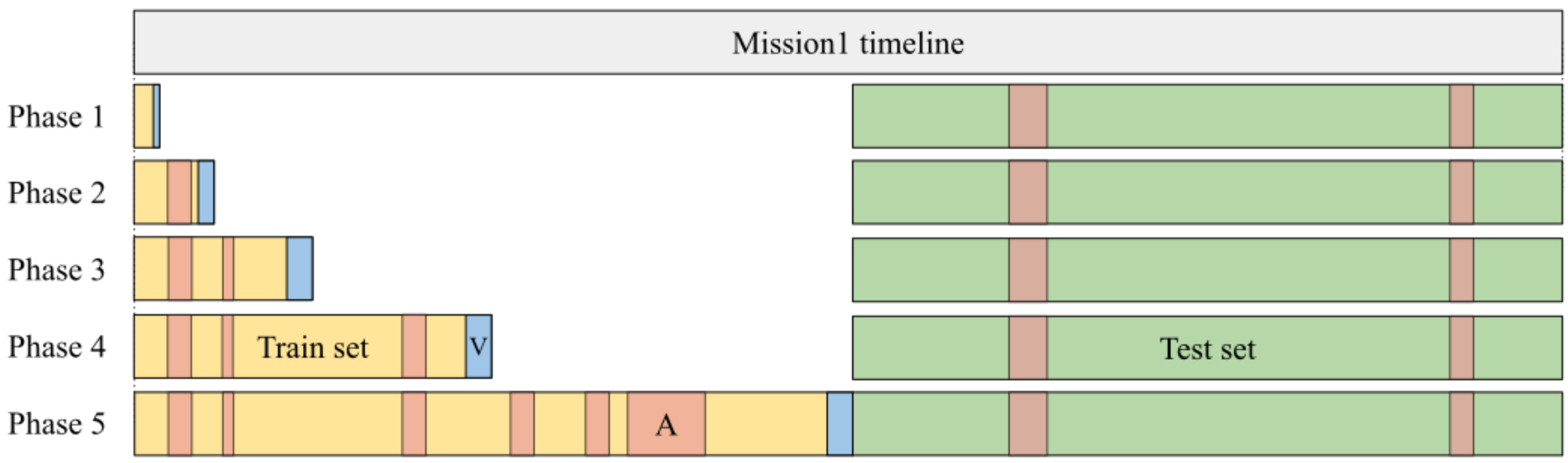

Supplementary Fig. 23. **Illustration of the idea of mission phases for Mission1**. "A" marks light red anomalous fragments and "V" marks blue validation fragments.

Supplementary Table 11. **Statistics of training, validation, and test sets for different phases of Mission1 considering the full set (top panel) and the lightweight subset of channels (bottom panel)**.

| **Mission1 – the lightweight subset** | Phase 1 | | Phase 2 | | Phase 3 | | Phase 4 | | Phase 5 | | **Test** |
|---|---|---|---|---|---|---|---|---|---|---|---|
| | Train | Val | Train | Val | Train | Val | Train | Val | Train | Val | |
| **Data points** | **1,125,600** | **314,399** | **3,900,977** | **997,530** | **8,900,105** | **1,479,360** | **19,171,279** | **1,463,274** | **39,774,080** | **1,479,370** | **40,925,288** |
| Telecommands' executions | 7,769 | 15,918 | 94,426 | 45,194 | 271,882 | 13,295 | 414,927 | 9,001 | 764,648 | 60,157 | 769,917 |
| Duration (anonymised) | 9 weeks | 3 weeks | 8 months | 2 months | 18 months | 3 months | 39 months | 3 months | 81 months | 3 months | 84 months |
| Annotated points [%] | 1.41 | 17.29 | 2.76 | 1.49 | 3.24 | 0.02 | 1.84 | 0.11 | 1.74 | 1.23 | 1.81 |
| **Annotated events** | **1** | **1** | **6** | **3** | **17** | **2** | **28** | **1** | **52** | **3** | **65** |
| Anomalies | 0 | 1 | 3 | 1 | 5 | 0 | 10 | 0 | 22 | 2 | 29 |
| Rare nominal events | 1 | 0 | 3 | 2 | 9 | 2 | 14 | 1 | 26 | 1 | 36 |
| Communication gaps | 0 | 0 | 0 | 0 | 3 | 0 | 4 | 0 | 4 | 0 | 0 |
| Univariate / Multivariate | 0 / 1 | 0 / 1 | 0 / 6 | 0 / 3 | 0 / 14 | 0 / 2 | 0 / 24 | 0 / 1 | 0 / 48 | 0 / 3 | 1 / 64 |
| Global / Local | 1 / 0 | 1 / 0 | 4 / 2 | 2 / 1 | 11 / 3 | 1 / 1 | 18 / 6 | 1 / 0 | 39 / 9 | 3 / 0 | 40 / 25 |
| Point / Subsequence | 0 / 1 | 0 / 1 | 0 / 6 | 0 / 3 | 0 / 14 | 0 / 2 | 0 / 24 | 0 / 1 | 1 / 47 | 2 / 1 | 9 / 56 |
| **Distinct event classes** | **1** | **1** | **5** | **2** | **10** | **2** | **15** | **1** | **17** | **2** | **13** |

| **Mission1 - the full set** | Phase 1 | | Phase 2 | | Phase 3 | | Phase 4 | | Phase 5 | | **Test** |
|---|---|---|---|---|---|---|---|---|---|---|---|
| | Train | Val | Train | Val | Train | Val | Train | Val | Train | Val | |
| **Data points** | **8,954,221** | **2,176,171** | **29,416,435** | **7,890,008** | **68,888,013** | **10,761,293** | **144,775,815** | **10,273,971** | **305,515,601** | **10,741,556** | **428,599,738** |
| Annotated points [%] | 2.11 | 10.62 | 1.87 | 1.52 | 1.96 | 0.93 | 1.32 | 0.03 | 1.33 | 1.62 | 2.25 |
| **Annotated events** | **5** | **4** | **20** | **8** | **54** | **4** | **73** | **1** | **104** | **5** | **91** |
| Anomalies | 4 | 1 | 13 | 2 | 27 | 2 | 40 | 0 | 59 | 4 | 55 |
| Rare nominal events | 1 | 3 | 7 | 6 | 24 | 2 | 29 | 1 | 41 | 1 | 36 |
| Communication gaps | 0 | 0 | 0 | 0 | 3 | 0 | 4 | 0 | 4 | 0 | 0 |
| Univariate / Multivariate | 3 / 2 | 3 / 1 | 11 / 9 | 5 / 3 | 31 / 20 | 0 / 4 | 31 / 38 | 0 / 1 | 31 / 69 | 0 / 5 | 1 / 90 |
| Global / Local | 3 / 2 | 4 / 0 | 11 / 9 | 5 / 3 | 36 / 15 | 1 / 3 | 44 / 25 | 1 / 0 | 67 / 33 | 3 / 2 | 43 / 48 |
| Point / Subsequence | 0 / 5 | 0 / 4 | 0 / 20 | 0 / 8 | 0 / 51 | 0 / 4 | 1 / 68 | 0 / 1 | 2 / 98 | 2 / 3 | 9 / 82 |
| **Distinct event classes** | **3** | **2** | **6** | **3** | **11** | **3** | **16** | **1** | **18** | **3** | **17** |

Supplementary Table 12. **Statistics of training, validation, and test sets for different phases of Mission2 considering the full set (top panel) and the lightweight subset of channels (bottom panel)**. There are no communication gaps and all events are of subsequence type, so these statistics are omitted.

| **Mission2 – the lightweight subset** | Phase 1 | | Phase 2 | | Phase 3 | | Phase 4 | | **Test** |
|---|---|---|---|---|---|---|---|---|---|
| | Train | Val | Train | Val | Train | Val | Train | Val | |
| **Data points** | **1,457,269** | **506,869** | **7,741,250** | **2,032,657** | **15,714,523** | **3,867,017** | **34,998,975** | **5,830,297** | **46,153,954** |
| Telecommands' executions | 34,185 | 11,694 | 179,930 | 48,313 | 372,643 | 93,496 | 815,370 | 130,968 | 1,077,677 |
| Duration (anonymised) | 3 weeks | 1 week | 4 months | 1 month | 8 months | 2 months | 18 months | 3 months | 21 months |
| Annotated points [%] | 0.83 | 0.49 | 2.62 | 2.02 | 1.94 | 3.10 | 3.74 | 1.02 | 2.02 |
| **Annotated events** | **14** | **4** | **83** | **19** | **140** | **27** | **246** | **27** | **349** |
| Anomalies | 0 | 0 | 2 | 2 | 11 | 2 | 18 | 0 | 4 |
| Rare nominal events | 14 | 4 | 81 | 17 | 129 | 25 | 228 | 27 | 345 |
| Univariate / Multivariate | 0 / 14 | 0 / 4 | 0 / 83 | 0 / 19 | 0 / 140 | 0 / 27 | 1 / 245 | 0 / 27 | 1 / 348 |
| Global / Local | 12 / 2 | 3 / 1 | 67 / 16 | 17 / 2 | 119 / 21 | 24 / 3 | 214 / 32 | 26 / 1 | 333 / 16 |
| **Distinct event classes** | **3** | **3** | **12** | **6** | **15** | **9** | **21** | **5** | **22** |

| **Mission2 – the full set** | Phase 1 | | Phase 2 | | Phase 3 | | Phase 4 | | **Test** |
|---|---|---|---|---|---|---|---|---|---|
| | Train | Val | Train | Val | Train | Val | Train | Val | |
| **Data points** | **13,914,918** | **4,841,396** | **74,356,579** | **19,067,743** | **151,093,710** | **37,624,768** | **338,658,318** | **56,746,734** | **444,603,954** |
| Annotated points [%] | 0.20 | 0.12 | 0.64 | 0.51 | 0.50 | 0.59 | 0.66 | 0.21 | 0.54 |
| **Annotated events** | **14** | **4** | **85** | **22** | **146** | **28** | **256** | **27** | **361** |
| Anomalies | 0 | 0 | 4 | 5 | 16 | 3 | 25 | 0 | 9 |
| Rare nominal events | 14 | 4 | 81 | 17 | 130 | 25 | 231 | 27 | 352 |
| Univariate / Multivariate | 0 / 14 | 0 / 4 | 1 / 84 | 2 / 20 | 3 / 143 | 1 / 27 | 5 / 251 | 0 / 27 | 4 / 357 |
| Global / Local | 12 / 2 | 3 / 1 | 67 / 18 | 17 / 5 | 120 / 26 | 24 / 4 | 217 / 39 | 26 / 1 | 340 / 21 |
| **Distinct event classes** | **3** | **3** | **14** | **8** | **18** | **10** | **24** | **5** | **26** |

Supplementary Table 13. **The effect of mission phase on the corrected event-wise F0.5-score for selected algorithms trained and tested on the lightweight subsets of channels from missions in ESA-ADB**. Boldfaced results indicate the best values among all phases.

<table>
<tr><th colspan="12">Mission1 – trained and tested on lightweight subset of channels 41-46</th></tr>
<tr><th>Phase</th><th>PCC[32]</th><th>HBOS[33]</th><th>iForest[30]</th><th>Window iForest[30]</th><th>KNN[34]</th><th>Global STD3</th><th>Global STD5</th><th>DC-VAE-ESA STD3</th><th>DC-VAE-ESA STD5</th><th>Teleman-ESA</th><th>Teleman-ESA-Pruned</th></tr>
<tr><td>1</td><td colspan="5" rowspan="5">< 0.001</td><td rowspan="4">< 0.001</td><td>0.041</td><td rowspan="2">< 0.001</td><td>0.007</td><td>0.059</td><td>0.227</td></tr>
<tr><td>2</td><td>0.037</td><td>0.012</td><td>0.058</td><td>0.311</td></tr>
<tr><td>3</td><td>0.104</td><td>0.007</td><td>0.085</td><td>0.122</td><td>0.776</td></tr>
<tr><td>4</td><td>0.217</td><td>0.009</td><td>0.030</td><td>0.309</td><td>0.776</td></tr>
<tr><td>5</td><td>0.001</td><td>0.253</td><td>0.003</td><td>0.075</td><td>0.178</td><td>0.786</td></tr>
<tr><th colspan="12">Mission2 – trained and tested on lightweight subset of channels 18-28</th></tr>
<tr><th>Phase</th><th>PCC[32]</th><th>HBOS[33]</th><th>iForest[30]</th><th>Window iForest[30]</th><th>KNN[34]</th><th>Global STD3</th><th>Global STD5</th><th>DC-VAE-ESA STD3</th><th>DC-VAE-ESA STD5</th><th>Teleman-ESA</th><th>Teleman-ESA-Pruned</th></tr>
<tr><td>1</td><td>< 0.001</td><td>< 0.001</td><td>0.006</td><td>0.020</td><td rowspan="3">< 0.001</td><td>< 0.001</td><td>< 0.001</td><td rowspan="2">< 0.001</td><td>< 0.001</td><td>0.234</td><td>0.622</td></tr>
<tr><td>2</td><td>0.062</td><td>0.007</td><td>0.456</td><td>0.901</td><td>0.001</td><td>0.011</td><td>0.001</td><td>0.259</td><td>0.757</td></tr>
<tr><td>3</td><td>0.013</td><td>0.040</td><td>0.585</td><td>0.947</td><td>0.006</td><td>0.014</td><td>0.001</td><td>0.009</td><td>0.253</td><td>0.731</td></tr>
<tr><td>4</td><td>0.036</td><td>0.068</td><td>0.609</td><td>0.949</td><td>0.001</td><td>0.007</td><td>0.075</td><td>0.003</td><td>0.079</td><td>0.224</td><td>0.842</td></tr>
</table>

There is a clear correlation between the training set length and the event-wise F0.5 scores for test sets for both missions. Especially significant improvements are visible between phases 2 and 3 for Mission1 and phases 1 and 2 for Mission 2. A clear example is Windowed iForest for which the event-wise F0.5-score goes from 0.020 to 0.901 for Mission2 in the phase 2. Based on this observation, the minimal reasonable training length can be estimated to be 21 months for Mission1 and 5 months for Mission2. Suprisingly, the longest training sets do not always ensure the best results. There are some exceptions for which training on the longest training set does not give optimal results, i.e. PCC, DC-VAE-ESA, and Telemanom-ESA. We can only hypothesize what is the reason behind that, but it may be related to the concept drift present in the data.

## 4.5. Computational resources and limitations

Experiments were run on three different machines:

1. Nvidia Tesla T4 GPU (16 GB VRAM), Intel Xeon Gold 5222 CPU 3.80 GHz, and 64 GB RAM, (for CPU-intensive and memory-intensive algorithms)
2. Nvidia 3060 RTX GPU (6 GB VRAM), Intel i7-10870H CPU 2.20 GHz, and 32 GB RAM
3. Nvidia 3090 RTX GPU (24 GB VRAM), Intel i7-8700H CPU 3.20 GHz, and 32 GB RAM (for GPU-intensive algorithms)

Given the limited resources, there are limits to the amount of time and memory that each algorithm can run. The algorithm is rejected with an out-of-memory error if Machine 1 goes out of RAM. Algorithms are rejected with an out-of-time error if it takes more than 5 days to train or test a CPU-intensive algorithm on Machine 1, or a GPU-intensive algorithm on Machine 3.

## 4.6. Processing times

Algorithms for satellite telemetry monitoring must not only be accurate but also fast enough to run in real-time on computational resources available to mission control and, in the extreme case, on board satellites. We measured the times of training (Supplementary Table 14) and execution (Supplementary Table 15) of algorithms on our hardware resources (listed in Supplementary Material 4.5). These numbers are not directly comparable because the algorithms were run in parallel processes on different machines. They give a rough approximation of the computational burden of each algorithm based on a single run in ESA-ADB. The training and execution times do not include resampling which was done once as an intermediate step before all experiments. The resampling of the test sets took about 1.5 hours for Mission1 and around 1 hour for Mission2, both on Machine 2.

The deep learning-based Telemanom has the longest training and execution times (excluding the execution time of KNN for channels 18-28 of Mission2), but it is still fast enough to provide real-time anomaly detection in both missions using the proposed resampling (0.033 Hz for Mission1, 0.056 Hz for Mission2). The total execution time (including resampling) for the full Mission1 test set is 3.5h which is just 0.02% of the test set duration, for Mission2 it is 4.5h and 0.08%, respectively. Thus, real-time execution should be possible even for sampling rates higher than 30 Hz. Moreover, in our previous works, we have shown that Telemanom can be run in real-time on-board the OPS-SAT satellite with a limited number of channels[14]. The important advantage of simple algorithms is that they are very fast and their training and execution times do not grow significantly with the number of channels, so it may be feasible to retrain them frequently during a mission.

Supplementary Table 14. **Training times (in seconds) of algorithms used in ESA-ADB**.

| Algorithm | Mission1 train set | | Mission2 train set | |
|---|---|---|---|---|
| | channels 41-46 | Full | channels 18-28 | Full |
| PCC | 90 | 143 | 63 | 75 |
| HBOS | 110 | 111 | 66 | 68 |
| iForest | 655 | 714 | 345 | 308 |
| Windowed iForest | 2833 (0.8h) | Out-of-memory | 1998 (0.6h) | 14585 (4h) |
| KNN | 3844 (1h) | Out-of-memory | 4754 (1.5h) | Out-of-memory |
| GlobalSTD | 101 | 108 | 60 | 90 |
| DC-VAE-ESA | 13466 (3.7h) | 18210 (5h) | 12440 (3.5h) | 4679 (1.3h) |
| Telemanom-ESA | 13115 (3.6h) | 30451 (8.5h) | 19725 (5.5h) | 12328 (3.5h) |

Supplementary Table 15. **Execution times (in seconds) of algorithms used in ESA-ADB**.

| Algorithm | Mission1 test set | | Mission2 test set | |
|---|---|---|---|---|
| | channels 41-46 | Full | channels 18-28 | Full |
| PCC | 124 | 141 | 73 | 76 |
| HBOS | 135 | 137 | 74 | 76 |
| iForest | 393 | 369 | 199 | 174 |
| Windowed iForest | 586 | Out-of-memory | 381 | 939 |
| KNN | 1233 | Out-of-memory | 21673 (6h) | Out-of-memory |
| GlobalSTD | 178 | 182 | 95 | 289 |
| DC-VAE-ESA | 5251 (1.5h) | 6010 (1.7h) | 3068 (0.9h) | 7900 (2.2h) |
| Telemanom-ESA | 6931 (1.9h) | 7271 (2h) | 4666 (1.3h) | 11078 (3.1h) |